\pdfoutput=1
\documentclass[11pt]{article}

\usepackage[]{acl}

\usepackage{times}
\usepackage{latexsym}

\usepackage[T1]{fontenc}
\usepackage[utf8]{inputenc}

\usepackage{microtype}

\usepackage{inconsolata}

\usepackage{graphicx}
\usepackage{url}
\usepackage{colortbl}
\usepackage{multirow}

\title{MultiSocial: Multilingual Benchmark of Machine-Generated Text Detection of Social-Media Texts}

\author{Dominik Macko, Jakub Kopal, Robert Moro, Ivan Srba \\
  Kempelen Institute of Intelligent Technologies, Slovakia\\
  \texttt{\{dominik.macko, jakub.kopal, robert.moro, ivan.srba\}}@kinit.sk \\
  }

\begin{document}
\maketitle
\begin{abstract}
Recent LLMs are able to generate high-quality multilingual texts, indistinguishable for humans from authentic human-written ones. Research in machine-generated text detection is however mostly focused on the English language and longer texts, such as news articles, scientific papers or student essays. Social-media texts are usually much shorter and often feature informal language, grammatical errors, or distinct linguistic items (e.g., emoticons, hashtags). There is a gap in studying the ability of existing methods in detection of such texts, reflected also in the lack of existing multilingual benchmark datasets. To fill this gap we propose the first multilingual (22 languages) and multi-platform (5 social media platforms) dataset for benchmarking machine-generated text detection in the social-media domain, called MultiSocial\footnote{Code: \url{https://github.com/kinit-sk/multisocial} Dataset: \url{https://doi.org/10.5281/zenodo.13846152}}. It contains 472,097 texts, of which about 58k are human-written and approximately the same amount is generated by each of 7 multilingual LLMs. We use this benchmark to compare existing detection methods in zero-shot as well as fine-tuned form. Our results indicate that the fine-tuned detectors have no problem to be trained on social-media texts and that the platform selection for training matters.
\end{abstract}

\section{Introduction}
\label{sec:into}

The most advanced text-generation AI models, called large language models (LLMs), are able to generate high-quality texts in various languages \citep{qin2024multilingual}. Although this presents an opportunity to make a human life and work more efficient, it also presents a threat of being misused, as such generated texts are not easily recognisable by humans \citep{zellers2019defending}. This is especially crucial in regard to social-media networks (SMN, such as Facebook, X/Twitter, etc.), where anyone can be a source of a harmful content (without editorial consent) with a potentially wide reach (depending on a network) \citep{aimeur2023fake}. To prevent the LLM misuse (e.g., social engineering, disinformation spreading), a reliable mechanism to detect machine-generated text (MGT) is needed.

\begin{figure}[t]
%\vspace{-5mm}
\centering
\includegraphics[width=\linewidth]{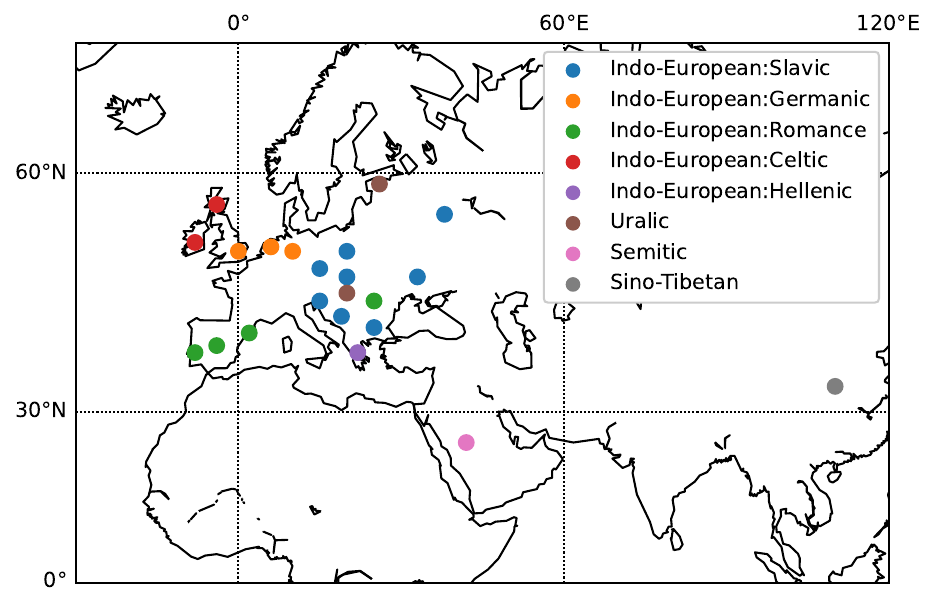}
\vspace{-3mm}
\caption{MultiSocial coverage of languages.}
\label{fig:languages}
%\vspace{-5mm}
\end{figure}

Unfortunately, the existing research in MGT detection (MGTD) focuses either solely on English (as in most NLP fields, e.g., \citealp{dugan2024raid}) or leaves SMN texts out of scope due to challenges they bring \citep{kumarage2024survey, lin2024detecting}. These challenges include very informal writing style often used in SMN, such as using slang, ignoring grammar rules, using distinct linguistic items in the texts (e.g., emoticons or hashtags), or usually including very short lengths of the texts.

Our work fills a gap in the state-of-the-art (SOTA) by introducing a new heavily multilingual dataset for MGTD research in the social-media domain. We use this dataset to further benchmark the existing SOTA detectors in various aspects. Specifically, our key contributions include:

\textbf{(1)} The first \textbf{multilingual evaluation of SOTA MGTD methods (statistical, pre-trained, fine-tuned) on social-media texts}, focusing on multilingual as well as cross-lingual capability of existing detectors and comparison of different categories.
The best detectors perform similarly across all tested languages, although there are still differences between English and non-English in zero-shot evaluation.

\textbf{(2)} The first \textbf{multi-platform and cross-platform evaluation of SOTA MGTD methods} in social-media domain, evaluating differences in MGTD performance based on text types and sources in multiple languages. We found that Telegram offers the best cross-lingual capability.

\textbf{(3)} The unique \textbf{multilingual, multi-platform, and multi-generator benchmark dataset} of human-written and machine-generated social-media texts, called MultiSocial, covering 22 languages, 5 SMN platforms, and 7 LLM generators.

\section{Related Work}
\label{sec:related}
%\vspace{-1mm}

Multilingual machine-generated text detection has been gaining attention recently. There have been multiple non-English language MGT detection shared tasks in the last years, such as Russian at RuATD~2022 \citep{Shamardina_2022}, Spanish at AuTexTification~2023 \citep{sarvazyan2023overview}, Dutch at CLIN33 \citep{clin33}, or multilingual at SemEval-2024 Task~8 \citep{semeval2024task8}.

The last one covers 9 languages, and is based on M4GT-Bench \citep{wang2024m4gtbench}, multilingual, multi-domain, and multi-generator MGT benchmark. However, its coverage of the SMN domain is rather sparse (solely English-only Reddit texts are included). Moreover, language coverage of the domains is highly imbalanced (e.g., most languages are represented only in the news domain, while Arabic and German also in Wikipedia domain, and Chinese only in the QA domain) and there are only one or two generators used in the multilingual settings. Therefore, cross-lingual evaluation is somewhat limited using such data.

Another benchmark dataset called MULTITuDE \citep{macko-etal-2023-multitude} covers 11 languages; however, 8 of them are included in the test split only with a limited number of samples. Moreover, it is focused on the news domain only, in which the texts are typically long, formal, and carefully checked for grammatical correctness. Such settings are clearly different than those of social media texts. An extension of MULTITuDE dataset to evaluate various authorship obfuscation methods is proposed in \citep{macko2024authorship}. Although it enables to evaluate robustness of MGT detection methods, it is still limited to news domain.

There is also MAiDE-up dataset \citep{ignat2024maideup} of hotel reviews available, covering 10 languages; however, it is limited to GPT-4 generated data only (limiting the generalization of conclusions). Although such texts are more similar to social-media texts than news articles, they cover a single topic (accommodation) and still use a different communication style than the social-media networks.
Many other works have focused on detection of fake reviews, but only few of them reflected multilingualism \citep{duma2024fake}.

Benchmark datasets HC3 \citep{guo-etal-2023-hc3} and SAID \citep{cui2023said} contain Chinese and English texts covering forum-like question-answering domain. HC3 contains only ChatGPT-generated machine texts, SAID contains real bot-generated texts (relying on human annotations to identify them). The downside of SAID is that the specific generation model of individual texts is not known. Otherwise, social-media texts are covered only in rather old TweepFake \citep{Fagni_2021} dataset, which includes English-only tweets and cannot be used to evaluate MGT detection methods on social-media texts generated by the most modern LLMs. Fox8-23 \citep{yang2023anatomy} dataset also focuses on English, and covers presumably only ChatGPT-generated tweets. Similarly, F3 \citep{lucas-etal-2023-fighting} dataset contains English ChatGPT-generated real and fake news as well as tweets. There are other monolingual works, such as \citep{temnikova-etal-2023-looking} focused on fake tweets in Bulgarian; however, the crafted dataset is not publicly available.

\begin{table}[!t]
%\vspace{-3mm}
\centering
\resizebox{\linewidth}{!}{
\begin{tabular}{p{3.1cm}|cccp{1cm}c}
\hline
\bfseries Dataset & \bfseries H/M & \bfseries LLM & \bfseries Lang & \bfseries Domain & \bfseries SMN \\
\hline
\textbf{TweepFake} \citep{Fagni_2021} & 12k/12k & 6 & 1 & SMN & 1 \\
\textbf{Fox8-23} \citep{yang2023anatomy} & 228k/170k & 1 & 1 & SMN & 1 \\
\textbf{F3} \citep{lucas-etal-2023-fighting} & 28k/28k & 1 & 1 & news, SMN & 1 \\
\textbf{HC3} \citep{guo-etal-2023-hc3} & 81k/44k & 1 & 2 & QA, Wiki & 0 \\
\textbf{SAID} \citep{cui2023said} & 87k/131k & N/A & 2 & QA & 0 \\
\textbf{MULTITuDE} \citep{macko-etal-2023-multitude} & 8k/66k & 8 & 11 & news & 0 \\
\textbf{M4GT-Bench} \citep{wang2024m4gtbench} & 93k/124k & 8 & 9 & 6 & 0 \\
\hline
\textbf{MultiSocial} (ours) & 58k/414k & 7 & 22 & SMN & 5 \\
\hline
\end{tabular}
}
%\vspace{-2mm}
\caption{Comparison of existing publicly available MGT detection datasets either multilingual or focused on social-media network (SMN) domain. H/M refers to the no. of human-written and machine-generated samples.}
\label{tab:datasets}
%\vspace{-5mm}
\end{table}

\begin{table*}[!t]
\centering
%\vspace{-1.5mm}
\resizebox{\textwidth}{!}{
\begin{tabular}{l|ccccc|c|ccccc|c}
\hline
& \multicolumn{6}{c|}{\bfseries Train} & \multicolumn{6}{c}{\bfseries Test} \\
\bfseries Language & \bfseries Discord & \bfseries Gab & \bfseries Telegram & \bfseries Twitter & \bfseries WhatsApp & \bfseries all & \bfseries Discord & \bfseries Gab & \bfseries Telegram & \bfseries Twitter & \bfseries WhatsApp & \bfseries all \\
\hline
\bfseries Arabic (ar) & 0 & 0 & 7724 & 7872 & 0 & 15596 & 0 & 1556 & 2319 & 2364 & 1750 & 7989 \\
\bfseries Bulgarian (bg) & 0 & 0 & 7555 & 7930 & 0 & 15485 & 0 & 192 & 2334 & 2367 & 0 & 4893 \\
\bfseries Catalan (ca) & 6984 & 0 & 7157 & 0 & 0 & 14141 & 2105 & 1264 & 2134 & 1824 & 144 & 7471 \\
\bfseries Chinese (zh) & 0 & 0 & 7924 & 0 & 0 & 7924 & 0 & 830 & 2380 & 238 & 32 & 3480 \\
\bfseries Croatian (hr) & 7502 & 0 & 7065 & 0 & 0 & 14567 & 2255 & 1340 & 2315 & 91 & 38 & 6039 \\
\bfseries Czech (cs) & 3450 & 0 & 7690 & 0 & 0 & 11140 & 2175 & 492 & 2309 & 1017 & 134 & 6127 \\
\bfseries Dutch (nl) & 7391 & 7900 & 7750 & 7933 & 0 & 30974 & 2191 & 2356 & 2318 & 2369 & 397 & 9631 \\
\bfseries English (en) & 7789 & 7760 & 7824 & 7871 & 7782 & 39026 & 2341 & 2340 & 2350 & 2365 & 2334 & 11730 \\
\bfseries Estonian (et) & 6974 & 0 & 7520 & 0 & 0 & 14494 & 2071 & 805 & 2259 & 164 & 120 & 5419 \\
\bfseries German (de) & 3407 & 7863 & 7880 & 2095 & 0 & 21245 & 2232 & 2366 & 2356 & 2345 & 304 & 9603 \\
\bfseries Greek (el) & 0 & 0 & 3814 & 0 & 0 & 3814 & 0 & 1195 & 2274 & 146 & 35 & 3650 \\
\bfseries Hungarian (hu) & 7079 & 0 & 7461 & 0 & 0 & 14540 & 2094 & 1211 & 2228 & 413 & 22 & 5968 \\
\bfseries Irish (ga) & 0 & 0 & 0 & 0 & 0 & 0 & 1319 & 968 & 821 & 45 & 0 & 3153 \\
\bfseries Polish (pl) & 7158 & 1829 & 7733 & 0 & 0 & 16720 & 2136 & 2311 & 2310 & 172 & 62 & 6991 \\
\bfseries Portuguese (pt) & 6860 & 7916 & 7842 & 6481 & 4354 & 33453 & 2284 & 2371 & 2347 & 2360 & 2363 & 11725 \\
\bfseries Romanian (ro) & 7436 & 851 & 6792 & 64 & 0 & 15143 & 2236 & 2349 & 2298 & 2378 & 132 & 9393 \\
\bfseries Russian (ru) & 0 & 7875 & 7827 & 362 & 0 & 16064 & 0 & 2355 & 2340 & 2361 & 960 & 8016 \\
\bfseries Scottish Gaelic (gd) & 0 & 0 & 0 & 0 & 0 & 0 & 150 & 35 & 34 & 0 & 0 & 219 \\
\bfseries Slovak (sk) & 0 & 0 & 0 & 0 & 0 & 0 & 107 & 308 & 1508 & 110 & 0 & 2033 \\
\bfseries Slovenian (sl) & 0 & 0 & 0 & 0 & 0 & 0 & 203 & 1912 & 917 & 40 & 0 & 3072 \\
\bfseries Spanish (es) & 7588 & 7883 & 7884 & 7922 & 7804 & 39081 & 2268 & 2361 & 2354 & 2376 & 2341 & 11700 \\
\bfseries Ukrainian (uk) & 0 & 0 & 7802 & 0 & 0 & 7802 & 0 & 174 & 2342 & 70 & 0 & 2586 \\
\hline
\bfseries \textbf{Total} & 79618 & 49877 & 133244 & 48530 & 19940 & \textbf{331209} & 28167 & 31091 & 44847 & 25615 & 11168 & \textbf{140888} \\
\hline
\end{tabular}
}
%\vspace{-2mm}
\caption{MultiSocial text sample counts across languages and platforms for train and test split.}
\label{tab:multisocial_sample_counts}
%\vspace{-5mm}
\end{table*}

%The existing datasets are not usable for benchmarking of multilingual MGT detection methods in social-media domain. Therefore, a new benchmark dataset is required. 
Table~\ref{tab:datasets} includes comparison of the new dataset proposed in this work (described in the next section) with the selected existing publicly available datasets for MGT detection.

%\vspace{-1mm}
\section{Dataset}
\label{sec:dataset}
%\vspace{-1mm}

Since there is no dataset of multilingual SMN texts containing human-written texts along with the texts generated by SOTA text-generation LLMs, we have crafted a new MultiSocial benchmark dataset. It contains human-written data from five different social-media platforms (reused from the existing multilingual SMN datasets), namely Telegram, Twitter (X), Gab, Discord, and WhatsApp, including post-like as well as chat-like texts. For each authentic human-written text, the texts generated by 7 SOTA LLMs (representatives of private and open multilingual LLMs of various sizes and architectures) are included by using \textbf{three iterations of paraphrasing}. Other text-generation approaches were also considered, outputs of which were evaluated and compared by humans, automated similarity metrics, and meta-evaluation utilizing LLM judges. The final approach was selected based on sufficient output quality and similarity to the human texts (to avoid detection biases). Details regarding dataset construction (including selection, evaluation, pre-processing, generation, and post-processing of texts) are provided in Appendix~\ref{sec:datacreation}.

Dataset consists of 472,097 texts (58k are human-written) split into train and test subsets, of which sample counts are summarized across languages and platforms in Table~\ref{tab:multisocial_sample_counts}. We have conducted linguistic analysis of the machine-generated texts along with the similarity comparison to human texts, with a manual human check of a balanced subset, and identified small portion (about 1\%) of noise in the generated data (e.g., ``As an AI model...''), indicating model failure during generation. We have intentionally left such samples in the dataset (clearly marked in the data) for further analysis purpose (as indicated in Appendix~\ref{sec:datacreation}). We however filter-out the identified noise for the purpose of the experiments in this study. Although such a post-processing cannot guarantee 100\% removal of noisy data, the obvious failures are cleaned. Furthermore, we have used meta-evaluation utilizing LLM judges to compare the quality of the generated texts of individual generators to the quality of original human texts, indicating that \textbf{machine-generated texts are of similar or higher quality} (see Appendix~\ref{sec:datacreation})).

\begin{table*}[!t]
\centering
\resizebox{\linewidth}{!}{
\begin{tabular}{lccccccc}
\hline
\bfseries Generator & \bfseries METEOR $\uparrow$ & \bfseries BERTScore $\uparrow$ & \bfseries ngram $\uparrow$ & \bfseries LD $\downarrow$ & \bfseries MAUVE $\downarrow$ & \bfseries LangCheck $\downarrow$ \\
\hline
\bfseries Aya-101 & \bfseries 0.195 (±0.23) & \bfseries 0.675 (±0.14) & \bfseries 0.156 (±0.18) & 1.104 (±0.91) & 0.063 & 10.59\% \\
\bfseries Gemini & 0.152 (±0.20) & 0.621 (±0.10) & 0.087 (±0.17) & 16.296 (±33.39) & \bfseries 0.025 & 6.16\% \\
\bfseries GPT-3.5-Turbo-0125 & 0.143 (±0.18) & 0.664 (±0.10) & 0.080 (±0.10) & 2.359 (±3.82) & 0.076 & 22.80\% \\
\bfseries Mistral-7B-Instruct-v0.2 & 0.152 (±0.16) & 0.652 (±0.08) & 0.088 (±0.09) & 2.383 (±1.93) & 0.047 & 10.61\% \\
\bfseries OPT-IML-Max-30b & 0.127 (±0.19) & 0.659 (±0.12) & 0.105 (±0.14) & \bfseries 0.998 (±0.57) & 0.108 & 14.88\% \\
\bfseries v5-Eagle-7B-HF & 0.108 (±0.15) & 0.628 (±0.07) & 0.071 (±0.10) & 2.568 (±2.10) & 0.027 & \bfseries 6.01\% \\
\bfseries Vicuna-13b & 0.133 (±0.17) & 0.650 (±0.09) & 0.089 (±0.11) & 1.811 (±1.40) & 0.042 & 6.26\% \\
\hline
\end{tabular}
}
\caption{Similarity analysis between machine-generated (3-iteration paraphrased) and human-written (original) social-media texts for individual generation models [mean ($\pm$ std)]. Arrows refer to values representing more similar texts, boldfaced values represent the most similar texts for each metric.}
\label{tab:similarityanalysis}
%\vspace{-3mm}
\end{table*}

\paragraph{Language Selection.}
The MultiSocial benchmark dataset covers 22 (high- and low-resource) languages (some only in the testing split), 20 of which (18 of Indo-European and 2 of Uralic language families) have been selected based on our research-projects needs. However, we have intentionally included 2 more (Arabic of Semitic and Chinese of Sino-Tibetan family), which are completely linguistically and geographically unrelated, to study cross-lingual characteristics (\figurename~\ref{fig:languages}).
The dataset is strongly focused on the Indo-European language family, but contains 4 language families in total. Test split includes all of the train languages, but also 2 Celtic languages, 1 more South Slavic and 1 more West Slavic language. There are 5 writing scripts in both train and test splits of the dataset, where majority of languages uses Latin, but there is also Cyrillic (Russian, Ukrainian, Bulgarian), Arabic, Hanzi (Chinese), and Greek. Nice feature is that the train split includes at least 3 representatives of Germanic, Romance, Slavic-Latin, and Slavic-Cyrillic, which enables various combinations of studies regarding multilingual and cross-lingual characteristics of machine-generated text detection. Furthermore, in Slavic and Romance languages, there are included at least 2 representatives of languages that can be considered high-resource and low-resource.

Although not all languages have enough samples from each of five platforms (due to unavailability of human samples in the selected source datasets), specific subsets can be used to study specific characteristics (e.g., cross-platform transferability using English and Spanish languages, cross-lingual transferability using Telegram platform, surprise language and/or surprise platform evaluation not using specific portions of the train data, etc.).

\paragraph{Human-Machine Similarity Analysis.}
As mentioned, we have conducted a similarity analysis of the final generated texts by the selected generation models (reported in Table~\ref{tab:similarityanalysis}). Definitions of the used metrics are available in Appendix~\ref{sec:datacreation}. We can observe only small differences between the generators. Aya-101 generated the most similar texts in general, while OPT-IML-Max-30B provided the worst results (based on higher MAUVE and LangCheck scores). ChatGPT (GPT-3.5-Turbo) also generated the texts resulting in a little higher MAUVE score and the highest language mismatch (LangCheck). However, the language mismatch of ChatGPT (using all the selected languages) was not confirmed by longer news articles generation (FastText language detection in such texts is definitely more accurate), where it actually achieved the lowest (under 1\%) LangCheck score among the generators. We assume that in social-media text generation it can better follow the original style of the text than the other generators (e.g., grammatically incorrect, slang), which is more difficult for accurate language detection.

\section{Detection Methods}
\label{sec:methods}

For the benchmark purpose, we have covered 3 specific categories of detection methods: \textit{statistical zero-shot} (methods using statistical differences to differentiate human-written and machine-generated texts applicable without training), \textit{pre-trained} (directly applicable models that are fine-tuned for the MGT detection task using different data -- i.e., out-of-distribution), and \textit{fine-tuned} (foundation models fine-tuned for the MGT detection task using MultiSocial dataset -- i.e., in-distribution).

For statistical category, we have selected the following 5 most-promising methods (excluding perturbation-based and multi-generation methods due to high computing costs): \textbf{Binoculars} \citep{hans2024spotting} with Falcon-7B \citep{falcon40b} as an observer model and Falcon-7B-Instruct as a performer model, 
\textbf{Fast-DetectGPT} \citep{bao2023fast} with GPT-J-6B \citep{gpt-j} as both the reference and sampling models, 
\textbf{LLM-Deviation} \citep{wu2023mfd}, 
\textbf{DetectLLM-LRR} \citep{su2023detectllm}, 
\textbf{S5} \citep{spiegel-macko-2024-kinit} (multiplying 5 statistical metrics of Likelihood, Entropy, Rank, Log-Rank, and LLM-Deviation), 
all three of them using GPT-J-6B as a base model (the same as in Fast-DetectGPT).

For pre-trained category, we have selected the following 5 detectors with a multilingual potential: 
\textbf{ChatGPT-detector-RoBERTa-Chinese} \citep{guo-etal-2023-hc3} (with a Chinese fine-tuned model), 
\textbf{Longformer Detector} \citep{li2023deepfake} (showing decent multilingual potential in \citealp{macko2024authorship}), 
\textbf{ruRoBERTa-ruatd-binary}\footnote{\scriptsize\url{https://huggingface.co/orzhan/ruroberta-ruatd-binary}} (as a best single-model system in RuATD~2022 \citealp{Shamardina_2022}), 
\textbf{BLOOMZ-3B-mixed-detector} \citep{sivesind_2023} (with a heavily multilingual fine-tuned LLM), and 
\textbf{RoBERTa-Large OpenAI Detectors} \citep{solaiman2019release} (as a widely used representative, although English-only).

For fine-tuned category, we have selected 7 multilingual foundational models covering multiple architectures and model sizes: 
\textbf{mDeBERTa-v3-base} \citep{he2021debertav3} (as the best detector in \citealp{macko-etal-2023-multitude}), 
\textbf{XLM-RoBERTa-large} \citep{DBLP:journals/corr/abs-1911-02116} (as the best detector in \citealp{macko-etal-2023-multitude} based on AUC ROC),
\textbf{Mistral-7B} \citep{jiang2023mistral} (as the best single-model multilingual detector in of SemEval-2024 Task~8 \citealp{semeval2024task8}), \textbf{Llama-3-8B} \citep{llama3modelcard} (as a SOTA multilingual smaller LLM), \textbf{Aya-101} \citep{ustun2024aya} (as a SOTA representative of multilingual LLMs with encoder-decoder architecture), \textbf{BLOOMZ-3B} \citep{muennighoff2022crosslingual} (as the best pre-trained detector base model), and \textbf{Falcon-rw-1B} (as a smaller version of the best performing model at ALTA~2023 \citealp{gagiano2023prompt}, since the 7B model version is already covered by better performing Mistral, \citealp{spiegel-macko-2024-kinit}).

For statistical and pre-trained categories of detection methods, we have used their versions implemented in the IMGTB framework \citep{spiegel-macko-2024-imgtb}.

\section{Experimental Results}
\label{sec:experiments}

Firstly, we provide benchmark evaluation of the selected MGTD methods on all MultiSocial test data (Table~\ref{tab:benchmark}). It provides a fair comparison of the methods, although the data samples among platforms and languages are not perfectly balanced. Therefore, for further experiments targeting specific cross-lingual and cross-platform research questions, the carefully selected parts of train and test splits are used (described in the corresponding subsections). Per-language, per-platform, and per-generator results are provided in Appendix~\ref{sec:data}.
Although worse-than-random performances of some pre-trained detectors indicate a potential problem with the detection (e.g., flipped labels), it is not the case as confirmed by the results in Table~\ref{tab:data_perllm_pretrained}.

\begin{table}[!t]
%\vspace{-3mm}
\centering
\resizebox{\linewidth}{!}{
\addtolength{\tabcolsep}{-4pt}
\begin{tabular}{cm{6cm}@{}cc}
\hline
\bfseries Rank & \bfseries Detector & \bfseries AUC ROC & \bfseries \begin{tabular}{@{}c@{}}MacroF1\\ @5\%FPR\end{tabular} \\
\hline
{\cellcolor[HTML]{B6D7A8}} 1 & {\cellcolor[HTML]{B6D7A8}} Llama-3-8b-MultiSocial & {\cellcolor[HTML]{B6D7A8}} 0.9769 & {\cellcolor[HTML]{B6D7A8}} 0.8696 \\
{\cellcolor[HTML]{B6D7A8}} 2 & {\cellcolor[HTML]{B6D7A8}} Mistral-7b-v0.1-MultiSocial & {\cellcolor[HTML]{B6D7A8}} 0.9768 & {\cellcolor[HTML]{B6D7A8}} 0.8692 \\
{\cellcolor[HTML]{B6D7A8}} 3 & {\cellcolor[HTML]{B6D7A8}} Aya-101-MultiSocial & {\cellcolor[HTML]{B6D7A8}} 0.9731 & {\cellcolor[HTML]{B6D7A8}} 0.8462 \\
{\cellcolor[HTML]{B6D7A8}} 4 & {\cellcolor[HTML]{B6D7A8}} Falcon-rw-1b-MultiSocial & {\cellcolor[HTML]{B6D7A8}} 0.9592 & {\cellcolor[HTML]{B6D7A8}} 0.7810 \\
{\cellcolor[HTML]{B6D7A8}} 5 & {\cellcolor[HTML]{B6D7A8}} BLOOMZ-3b-MultiSocial & {\cellcolor[HTML]{B6D7A8}} 0.9582 & {\cellcolor[HTML]{B6D7A8}} 0.7843 \\
{\cellcolor[HTML]{B6D7A8}} 6 & {\cellcolor[HTML]{B6D7A8}} XLM-RoBERTa-large-MultiSocial & {\cellcolor[HTML]{B6D7A8}} 0.9553 & {\cellcolor[HTML]{B6D7A8}} 0.7840 \\
{\cellcolor[HTML]{B6D7A8}} 7 & {\cellcolor[HTML]{B6D7A8}} mDeBERTa-v3-base-MultiSocial & {\cellcolor[HTML]{B6D7A8}} 0.9544 & {\cellcolor[HTML]{B6D7A8}} 0.7652 \\
{\cellcolor[HTML]{9FC5E8}} 8 & {\cellcolor[HTML]{9FC5E8}} BLOOMZ-3b-mixed-Detector & {\cellcolor[HTML]{9FC5E8}} 0.7553 & {\cellcolor[HTML]{9FC5E8}} 0.3024 \\
{\cellcolor[HTML]{F9CB9C}} 9 & {\cellcolor[HTML]{F9CB9C}} DetectLLM-LRR & {\cellcolor[HTML]{F9CB9C}} 0.7464 & {\cellcolor[HTML]{F9CB9C}} 0.2523 \\
{\cellcolor[HTML]{F9CB9C}} 10 & {\cellcolor[HTML]{F9CB9C}} LLM-Deviation & {\cellcolor[HTML]{F9CB9C}} 0.7454 & {\cellcolor[HTML]{F9CB9C}} 0.2497 \\
{\cellcolor[HTML]{F9CB9C}} 11 & {\cellcolor[HTML]{F9CB9C}} S5 & {\cellcolor[HTML]{F9CB9C}} 0.7418 & {\cellcolor[HTML]{F9CB9C}} 0.2465 \\
{\cellcolor[HTML]{F9CB9C}} 12 & {\cellcolor[HTML]{F9CB9C}} Fast-Detect-GPT & {\cellcolor[HTML]{F9CB9C}} 0.7418 & {\cellcolor[HTML]{F9CB9C}} 0.3605 \\
{\cellcolor[HTML]{F9CB9C}} 13 & {\cellcolor[HTML]{F9CB9C}} Binoculars & {\cellcolor[HTML]{F9CB9C}} 0.7248 & {\cellcolor[HTML]{F9CB9C}} 0.2815 \\
{\cellcolor[HTML]{9FC5E8}} 14 & {\cellcolor[HTML]{9FC5E8}} ChatGPT-Detector-RoBERTa-Chinese & {\cellcolor[HTML]{9FC5E8}} 0.7180 & {\cellcolor[HTML]{9FC5E8}} 0.3416 \\
{\cellcolor[HTML]{9FC5E8}} 15 & {\cellcolor[HTML]{9FC5E8}} ruRoBERTa-ruatd-binary & {\cellcolor[HTML]{9FC5E8}} 0.4817 & {\cellcolor[HTML]{9FC5E8}} 0.1711 \\
{\cellcolor[HTML]{9FC5E8}} 16 & {\cellcolor[HTML]{9FC5E8}} Longformer Detector & {\cellcolor[HTML]{9FC5E8}} 0.4615 & {\cellcolor[HTML]{9FC5E8}} 0.1516 \\
{\cellcolor[HTML]{9FC5E8}} 17 & {\cellcolor[HTML]{9FC5E8}} RoBERTa-large-OpenAI-Detector & {\cellcolor[HTML]{9FC5E8}} 0.3450 & {\cellcolor[HTML]{9FC5E8}} 0.1376 \\
\hline
\end{tabular}
}
\caption{Benchmark of the selected MGTD methods of \colorbox[HTML]{F9CB9C}{statistical}, \colorbox[HTML]{9FC5E8}{pre-trained}, and \colorbox[HTML]{B6D7A8}{fine-tuned} categories (as defined in Section~\ref{sec:methods}). The highlight color identifies the category of the method in the table.}
\label{tab:benchmark}
%\vspace{-3mm}
\end{table}

\begin{table*}[!t]
\centering
\resizebox{\textwidth}{!}{
\addtolength{\tabcolsep}{-2pt}
\begin{tabular}{c|p{3.5cm}|cccccccccccccccccccccc|c}
\hline
 &  & \multicolumn{23}{c}{\bfseries Test Language [AUC ROC]} \\
\bfseries Category & \bfseries Detector & \bfseries ar & \bfseries bg & \bfseries ca & \bfseries cs & \bfseries de & \bfseries el & \bfseries en & \bfseries es & \bfseries et & \bfseries ga & \bfseries gd & \bfseries hr & \bfseries hu & \bfseries nl & \bfseries pl & \bfseries pt & \bfseries ro & \bfseries ru & \bfseries sk & \bfseries sl & \bfseries uk & \bfseries zh & \bfseries all \\
\hline
\multirow[c]{9}{*}{\bfseries P} & \bfseries BLOOMZ-3b-mixed-Detector & {\cellcolor[HTML]{C4CBE3}} \color[HTML]{000000} 0.79 & {\cellcolor[HTML]{D2D3E7}} \color[HTML]{000000} 0.74 & {\cellcolor[HTML]{C2CBE2}} \color[HTML]{000000} 0.79 & {\cellcolor[HTML]{C1CAE2}} \color[HTML]{000000} 0.80 & {\cellcolor[HTML]{CACEE5}} \color[HTML]{000000} 0.77 & {\cellcolor[HTML]{CCCFE5}} \color[HTML]{000000} 0.76 & {\cellcolor[HTML]{BFC9E1}} \color[HTML]{000000} 0.80 & {\cellcolor[HTML]{C4CBE3}} \color[HTML]{000000} 0.79 & {\cellcolor[HTML]{B4C4DF}} \color[HTML]{000000} 0.83 & {\cellcolor[HTML]{C8CDE4}} \color[HTML]{000000} 0.78 & {\cellcolor[HTML]{E4E1EF}} \color[HTML]{000000} 0.66 & {\cellcolor[HTML]{C9CEE4}} \color[HTML]{000000} 0.77 & {\cellcolor[HTML]{B4C4DF}} \color[HTML]{000000} 0.84 & {\cellcolor[HTML]{D1D2E6}} \color[HTML]{000000} 0.75 & {\cellcolor[HTML]{C5CCE3}} \color[HTML]{000000} 0.78 & {\cellcolor[HTML]{C2CBE2}} \color[HTML]{000000} 0.79 & {\cellcolor[HTML]{E4E1EF}} \color[HTML]{000000} 0.66 & {\cellcolor[HTML]{E0DEED}} \color[HTML]{000000} 0.68 & {\cellcolor[HTML]{CED0E6}} \color[HTML]{000000} 0.76 & {\cellcolor[HTML]{E8E4F0}} \color[HTML]{000000} 0.64 & {\cellcolor[HTML]{DCDAEB}} \color[HTML]{000000} 0.70 & {\cellcolor[HTML]{DEDCEC}} \color[HTML]{000000} 0.69 & {\cellcolor[HTML]{CED0E6}} \color[HTML]{000000} 0.76 \\
\bfseries  & \bfseries ChatGPT-Detector-RoBERTa-Chinese & {\cellcolor[HTML]{D6D6E9}} \color[HTML]{000000} 0.72 & {\cellcolor[HTML]{C0C9E2}} \color[HTML]{000000} 0.80 & {\cellcolor[HTML]{CED0E6}} \color[HTML]{000000} 0.76 & {\cellcolor[HTML]{E3E0EE}} \color[HTML]{000000} 0.66 & {\cellcolor[HTML]{C1CAE2}} \color[HTML]{000000} 0.80 & {\cellcolor[HTML]{D1D2E6}} \color[HTML]{000000} 0.75 & {\cellcolor[HTML]{9EBAD9}} \color[HTML]{000000} 0.90 & {\cellcolor[HTML]{BCC7E1}} \color[HTML]{000000} 0.81 & {\cellcolor[HTML]{BBC7E0}} \color[HTML]{000000} 0.82 & {\cellcolor[HTML]{D4D4E8}} \color[HTML]{000000} 0.73 & {\cellcolor[HTML]{EAE6F1}} \color[HTML]{000000} 0.63 & {\cellcolor[HTML]{ECE7F2}} \color[HTML]{000000} 0.63 & {\cellcolor[HTML]{C5CCE3}} \color[HTML]{000000} 0.78 & {\cellcolor[HTML]{DCDAEB}} \color[HTML]{000000} 0.70 & {\cellcolor[HTML]{EDE7F2}} \color[HTML]{000000} 0.62 & {\cellcolor[HTML]{D4D4E8}} \color[HTML]{000000} 0.73 & {\cellcolor[HTML]{D1D2E6}} \color[HTML]{000000} 0.75 & {\cellcolor[HTML]{CED0E6}} \color[HTML]{000000} 0.76 & {\cellcolor[HTML]{EAE6F1}} \color[HTML]{000000} 0.63 & {\cellcolor[HTML]{E5E1EF}} \color[HTML]{000000} 0.66 & {\cellcolor[HTML]{D2D3E7}} \color[HTML]{000000} 0.74 & {\cellcolor[HTML]{BDC8E1}} \color[HTML]{000000} 0.81 & {\cellcolor[HTML]{D8D7E9}} \color[HTML]{000000} 0.72 \\
\bfseries  & \bfseries Longformer Detector & {\cellcolor[HTML]{FFF7FB}} \color[HTML]{000000} 0.34 & {\cellcolor[HTML]{FFF7FB}} \color[HTML]{000000} 0.48 & {\cellcolor[HTML]{FFF7FB}} \color[HTML]{000000} 0.32 & {\cellcolor[HTML]{FFF7FB}} \color[HTML]{000000} 0.47 & {\cellcolor[HTML]{FFF7FB}} \color[HTML]{000000} 0.43 & {\cellcolor[HTML]{F9F2F8}} \color[HTML]{000000} 0.54 & {\cellcolor[HTML]{E7E3F0}} \color[HTML]{000000} 0.65 & {\cellcolor[HTML]{FFF7FB}} \color[HTML]{000000} 0.43 & {\cellcolor[HTML]{FFF7FB}} \color[HTML]{000000} 0.48 & {\cellcolor[HTML]{FBF3F9}} \color[HTML]{000000} 0.53 & {\cellcolor[HTML]{FFF7FB}} \color[HTML]{000000} 0.49 & {\cellcolor[HTML]{FDF5FA}} \color[HTML]{000000} 0.51 & {\cellcolor[HTML]{FFF7FB}} \color[HTML]{000000} 0.50 & {\cellcolor[HTML]{FFF7FB}} \color[HTML]{000000} 0.41 & {\cellcolor[HTML]{FFF7FB}} \color[HTML]{000000} 0.46 & {\cellcolor[HTML]{FFF7FB}} \color[HTML]{000000} 0.42 & {\cellcolor[HTML]{FFF7FB}} \color[HTML]{000000} 0.41 & {\cellcolor[HTML]{FFF7FB}} \color[HTML]{000000} 0.46 & {\cellcolor[HTML]{FFF7FB}} \color[HTML]{000000} 0.45 & {\cellcolor[HTML]{FFF7FB}} \color[HTML]{000000} 0.46 & {\cellcolor[HTML]{EFE9F3}} \color[HTML]{000000} 0.61 & {\cellcolor[HTML]{FFF7FB}} \color[HTML]{000000} 0.47 & {\cellcolor[HTML]{FFF7FB}} \color[HTML]{000000} 0.46 \\
\bfseries  & \bfseries RoBERTa-large-OpenAI-Detector & {\cellcolor[HTML]{D3D4E7}} \color[HTML]{000000} 0.73 & {\cellcolor[HTML]{FFF7FB}} \color[HTML]{000000} 0.43 & {\cellcolor[HTML]{FFF7FB}} \color[HTML]{000000} 0.43 & {\cellcolor[HTML]{FFF7FB}} \color[HTML]{000000} 0.14 & {\cellcolor[HTML]{FFF7FB}} \color[HTML]{000000} 0.32 & {\cellcolor[HTML]{D2D2E7}} \color[HTML]{000000} 0.74 & {\cellcolor[HTML]{FDF5FA}} \color[HTML]{000000} 0.52 & {\cellcolor[HTML]{FFF7FB}} \color[HTML]{000000} 0.30 & {\cellcolor[HTML]{FFF7FB}} \color[HTML]{000000} 0.20 & {\cellcolor[HTML]{FFF7FB}} \color[HTML]{000000} 0.30 & {\cellcolor[HTML]{FFF7FB}} \color[HTML]{000000} 0.48 & {\cellcolor[HTML]{FFF7FB}} \color[HTML]{000000} 0.19 & {\cellcolor[HTML]{FFF7FB}} \color[HTML]{000000} 0.13 & {\cellcolor[HTML]{FFF7FB}} \color[HTML]{000000} 0.30 & {\cellcolor[HTML]{FFF7FB}} \color[HTML]{000000} 0.21 & {\cellcolor[HTML]{FFF7FB}} \color[HTML]{000000} 0.23 & {\cellcolor[HTML]{FFF7FB}} \color[HTML]{000000} 0.24 & {\cellcolor[HTML]{FAF2F8}} \color[HTML]{000000} 0.54 & {\cellcolor[HTML]{FFF7FB}} \color[HTML]{000000} 0.26 & {\cellcolor[HTML]{FFF7FB}} \color[HTML]{000000} 0.33 & {\cellcolor[HTML]{FFF7FB}} \color[HTML]{000000} 0.36 & {\cellcolor[HTML]{F0EAF4}} \color[HTML]{000000} 0.60 & {\cellcolor[HTML]{FFF7FB}} \color[HTML]{000000} 0.35 \\
\bfseries  & \bfseries ruRoBERTa-ruatd-binary & {\cellcolor[HTML]{FFF7FB}} \color[HTML]{000000} 0.40 & {\cellcolor[HTML]{EAE6F1}} \color[HTML]{000000} 0.63 & {\cellcolor[HTML]{F6EFF7}} \color[HTML]{000000} 0.56 & {\cellcolor[HTML]{FFF7FB}} \color[HTML]{000000} 0.43 & {\cellcolor[HTML]{FFF7FB}} \color[HTML]{000000} 0.43 & {\cellcolor[HTML]{FFF7FB}} \color[HTML]{000000} 0.35 & {\cellcolor[HTML]{F7F0F7}} \color[HTML]{000000} 0.56 & {\cellcolor[HTML]{FFF7FB}} \color[HTML]{000000} 0.47 & {\cellcolor[HTML]{FFF7FB}} \color[HTML]{000000} 0.43 & {\cellcolor[HTML]{FFF7FB}} \color[HTML]{000000} 0.49 & {\cellcolor[HTML]{FFF7FB}} \color[HTML]{000000} 0.47 & {\cellcolor[HTML]{FFF7FB}} \color[HTML]{000000} 0.48 & {\cellcolor[HTML]{FFF7FB}} \color[HTML]{000000} 0.46 & {\cellcolor[HTML]{FFF7FB}} \color[HTML]{000000} 0.50 & {\cellcolor[HTML]{FFF7FB}} \color[HTML]{000000} 0.44 & {\cellcolor[HTML]{FFF7FB}} \color[HTML]{000000} 0.43 & {\cellcolor[HTML]{FFF7FB}} \color[HTML]{000000} 0.34 & {\cellcolor[HTML]{DAD9EA}} \color[HTML]{000000} 0.70 & {\cellcolor[HTML]{FFF7FB}} \color[HTML]{000000} 0.45 & {\cellcolor[HTML]{FFF7FB}} \color[HTML]{000000} 0.47 & {\cellcolor[HTML]{F2ECF5}} \color[HTML]{000000} 0.59 & {\cellcolor[HTML]{FFF7FB}} \color[HTML]{000000} 0.44 & {\cellcolor[HTML]{FFF7FB}} \color[HTML]{000000} 0.48 \\
\hline
\multirow[c]{5}{*}{\bfseries S} & \bfseries Binoculars & {\cellcolor[HTML]{DAD9EA}} \color[HTML]{000000} 0.70 & {\cellcolor[HTML]{E1DFED}} \color[HTML]{000000} 0.68 & {\cellcolor[HTML]{EEE8F3}} \color[HTML]{000000} 0.62 & {\cellcolor[HTML]{D2D3E7}} \color[HTML]{000000} 0.74 & {\cellcolor[HTML]{D0D1E6}} \color[HTML]{000000} 0.75 & {\cellcolor[HTML]{C2CBE2}} \color[HTML]{000000} 0.79 & {\cellcolor[HTML]{BFC9E1}} \color[HTML]{000000} 0.80 & {\cellcolor[HTML]{CED0E6}} \color[HTML]{000000} 0.76 & {\cellcolor[HTML]{D2D3E7}} \color[HTML]{000000} 0.74 & {\cellcolor[HTML]{DAD9EA}} \color[HTML]{000000} 0.71 & {\cellcolor[HTML]{D9D8EA}} \color[HTML]{000000} 0.71 & {\cellcolor[HTML]{C4CBE3}} \color[HTML]{000000} 0.79 & {\cellcolor[HTML]{C5CCE3}} \color[HTML]{000000} 0.78 & {\cellcolor[HTML]{D6D6E9}} \color[HTML]{000000} 0.72 & {\cellcolor[HTML]{D1D2E6}} \color[HTML]{000000} 0.75 & {\cellcolor[HTML]{D0D1E6}} \color[HTML]{000000} 0.75 & {\cellcolor[HTML]{D2D2E7}} \color[HTML]{000000} 0.74 & {\cellcolor[HTML]{D8D7E9}} \color[HTML]{000000} 0.72 & {\cellcolor[HTML]{D9D8EA}} \color[HTML]{000000} 0.71 & {\cellcolor[HTML]{D6D6E9}} \color[HTML]{000000} 0.73 & {\cellcolor[HTML]{E9E5F1}} \color[HTML]{000000} 0.64 & {\cellcolor[HTML]{D3D4E7}} \color[HTML]{000000} 0.74 & {\cellcolor[HTML]{D6D6E9}} \color[HTML]{000000} 0.72 \\
\bfseries  & \bfseries DetectLLM-LRR & {\cellcolor[HTML]{C2CBE2}} \color[HTML]{000000} 0.79 & {\cellcolor[HTML]{ACC0DD}} \color[HTML]{000000} 0.86 & {\cellcolor[HTML]{DFDDEC}} \color[HTML]{000000} 0.69 & {\cellcolor[HTML]{93B5D6}} \color[HTML]{000000} 0.93 & {\cellcolor[HTML]{C6CCE3}} \color[HTML]{000000} 0.78 & {\cellcolor[HTML]{A4BCDA}} \color[HTML]{000000} 0.88 & {\cellcolor[HTML]{BFC9E1}} \color[HTML]{000000} 0.80 & {\cellcolor[HTML]{B8C6E0}} \color[HTML]{000000} 0.82 & {\cellcolor[HTML]{A2BCDA}} \color[HTML]{000000} 0.88 & {\cellcolor[HTML]{C4CBE3}} \color[HTML]{000000} 0.79 & {\cellcolor[HTML]{D2D3E7}} \color[HTML]{000000} 0.74 & {\cellcolor[HTML]{A2BCDA}} \color[HTML]{000000} 0.88 & {\cellcolor[HTML]{8CB3D5}} \color[HTML]{000000} 0.94 & {\cellcolor[HTML]{C4CBE3}} \color[HTML]{000000} 0.79 & {\cellcolor[HTML]{A4BCDA}} \color[HTML]{000000} 0.88 & {\cellcolor[HTML]{B3C3DE}} \color[HTML]{000000} 0.84 & {\cellcolor[HTML]{A7BDDB}} \color[HTML]{000000} 0.87 & {\cellcolor[HTML]{C5CCE3}} \color[HTML]{000000} 0.78 & {\cellcolor[HTML]{ADC1DD}} \color[HTML]{000000} 0.85 & {\cellcolor[HTML]{C2CBE2}} \color[HTML]{000000} 0.79 & {\cellcolor[HTML]{D2D2E7}} \color[HTML]{000000} 0.75 & {\cellcolor[HTML]{C8CDE4}} \color[HTML]{000000} 0.78 & {\cellcolor[HTML]{D1D2E6}} \color[HTML]{000000} 0.75 \\
\bfseries  & \bfseries Fast-Detect-GPT & {\cellcolor[HTML]{D0D1E6}} \color[HTML]{000000} 0.75 & {\cellcolor[HTML]{E6E2EF}} \color[HTML]{000000} 0.65 & {\cellcolor[HTML]{EEE8F3}} \color[HTML]{000000} 0.61 & {\cellcolor[HTML]{BDC8E1}} \color[HTML]{000000} 0.81 & {\cellcolor[HTML]{CACEE5}} \color[HTML]{000000} 0.77 & {\cellcolor[HTML]{E4E1EF}} \color[HTML]{000000} 0.66 & {\cellcolor[HTML]{BFC9E1}} \color[HTML]{000000} 0.80 & {\cellcolor[HTML]{D2D2E7}} \color[HTML]{000000} 0.74 & {\cellcolor[HTML]{DEDCEC}} \color[HTML]{000000} 0.69 & {\cellcolor[HTML]{DBDAEB}} \color[HTML]{000000} 0.70 & {\cellcolor[HTML]{D2D3E7}} \color[HTML]{000000} 0.74 & {\cellcolor[HTML]{C0C9E2}} \color[HTML]{000000} 0.80 & {\cellcolor[HTML]{CACEE5}} \color[HTML]{000000} 0.77 & {\cellcolor[HTML]{D2D3E7}} \color[HTML]{000000} 0.74 & {\cellcolor[HTML]{C8CDE4}} \color[HTML]{000000} 0.77 & {\cellcolor[HTML]{C9CEE4}} \color[HTML]{000000} 0.77 & {\cellcolor[HTML]{C9CEE4}} \color[HTML]{000000} 0.77 & {\cellcolor[HTML]{D4D4E8}} \color[HTML]{000000} 0.73 & {\cellcolor[HTML]{D9D8EA}} \color[HTML]{000000} 0.71 & {\cellcolor[HTML]{D2D3E7}} \color[HTML]{000000} 0.74 & {\cellcolor[HTML]{DDDBEC}} \color[HTML]{000000} 0.70 & {\cellcolor[HTML]{D2D3E7}} \color[HTML]{000000} 0.74 & {\cellcolor[HTML]{D2D3E7}} \color[HTML]{000000} 0.74 \\
\bfseries  & \bfseries LLM-Deviation & {\cellcolor[HTML]{B9C6E0}} \color[HTML]{000000} 0.82 & {\cellcolor[HTML]{A9BFDC}} \color[HTML]{000000} 0.86 & {\cellcolor[HTML]{E0DEED}} \color[HTML]{000000} 0.68 & {\cellcolor[HTML]{8FB4D6}} \color[HTML]{000000} 0.93 & {\cellcolor[HTML]{C4CBE3}} \color[HTML]{000000} 0.79 & {\cellcolor[HTML]{A1BBDA}} \color[HTML]{000000} 0.89 & {\cellcolor[HTML]{C1CAE2}} \color[HTML]{000000} 0.80 & {\cellcolor[HTML]{B8C6E0}} \color[HTML]{000000} 0.82 & {\cellcolor[HTML]{9EBAD9}} \color[HTML]{000000} 0.90 & {\cellcolor[HTML]{BBC7E0}} \color[HTML]{000000} 0.81 & {\cellcolor[HTML]{C4CBE3}} \color[HTML]{000000} 0.79 & {\cellcolor[HTML]{9FBAD9}} \color[HTML]{000000} 0.89 & {\cellcolor[HTML]{8BB2D4}} \color[HTML]{000000} 0.94 & {\cellcolor[HTML]{C4CBE3}} \color[HTML]{000000} 0.79 & {\cellcolor[HTML]{A2BCDA}} \color[HTML]{000000} 0.89 & {\cellcolor[HTML]{B0C2DE}} \color[HTML]{000000} 0.84 & {\cellcolor[HTML]{A4BCDA}} \color[HTML]{000000} 0.88 & {\cellcolor[HTML]{C5CCE3}} \color[HTML]{000000} 0.78 & {\cellcolor[HTML]{ABBFDC}} \color[HTML]{000000} 0.86 & {\cellcolor[HTML]{BDC8E1}} \color[HTML]{000000} 0.81 & {\cellcolor[HTML]{D0D1E6}} \color[HTML]{000000} 0.75 & {\cellcolor[HTML]{C4CBE3}} \color[HTML]{000000} 0.79 & {\cellcolor[HTML]{D2D2E7}} \color[HTML]{000000} 0.75 \\
\bfseries  & \bfseries S5 & {\cellcolor[HTML]{BFC9E1}} \color[HTML]{000000} 0.80 & {\cellcolor[HTML]{AFC1DD}} \color[HTML]{000000} 0.85 & {\cellcolor[HTML]{E0DEED}} \color[HTML]{000000} 0.68 & {\cellcolor[HTML]{93B5D6}} \color[HTML]{000000} 0.92 & {\cellcolor[HTML]{C6CCE3}} \color[HTML]{000000} 0.78 & {\cellcolor[HTML]{A2BCDA}} \color[HTML]{000000} 0.88 & {\cellcolor[HTML]{C8CDE4}} \color[HTML]{000000} 0.77 & {\cellcolor[HTML]{BCC7E1}} \color[HTML]{000000} 0.81 & {\cellcolor[HTML]{A2BCDA}} \color[HTML]{000000} 0.89 & {\cellcolor[HTML]{BFC9E1}} \color[HTML]{000000} 0.80 & {\cellcolor[HTML]{C6CCE3}} \color[HTML]{000000} 0.78 & {\cellcolor[HTML]{A2BCDA}} \color[HTML]{000000} 0.88 & {\cellcolor[HTML]{8EB3D5}} \color[HTML]{000000} 0.94 & {\cellcolor[HTML]{C8CDE4}} \color[HTML]{000000} 0.77 & {\cellcolor[HTML]{A5BDDB}} \color[HTML]{000000} 0.88 & {\cellcolor[HTML]{B4C4DF}} \color[HTML]{000000} 0.83 & {\cellcolor[HTML]{A5BDDB}} \color[HTML]{000000} 0.88 & {\cellcolor[HTML]{C6CCE3}} \color[HTML]{000000} 0.78 & {\cellcolor[HTML]{AFC1DD}} \color[HTML]{000000} 0.85 & {\cellcolor[HTML]{C1CAE2}} \color[HTML]{000000} 0.80 & {\cellcolor[HTML]{D2D2E7}} \color[HTML]{000000} 0.74 & {\cellcolor[HTML]{C6CCE3}} \color[HTML]{000000} 0.78 & {\cellcolor[HTML]{D2D3E7}} \color[HTML]{000000} 0.74 \\
\hline
\end{tabular}
}
%\vspace{-1mm}
\caption{Per-language AUC ROC performance of zero-shot statistical (S) and pre-trained (P) MGT detectors. The data are too difficult for the three under-performing pre-trained models.}
\label{tab:data_zeroshot}
%\vspace{-3mm}
\end{table*}

\begin{table*}[!t]
\centering
\resizebox{\textwidth}{!}{
\addtolength{\tabcolsep}{-2pt}
\begin{tabular}{c|p{3.8cm}|cccccccccccccccccccccc|c}
\hline
 &  & \multicolumn{23}{c}{\bfseries Test Language [AUC ROC mean]} \\
\bfseries Category & \bfseries Platform & \bfseries ar & \bfseries bg & \bfseries ca & \bfseries cs & \bfseries de & \bfseries el & \bfseries en & \bfseries es & \bfseries et & \bfseries ga & \bfseries gd & \bfseries hr & \bfseries hu & \bfseries nl & \bfseries pl & \bfseries pt & \bfseries ro & \bfseries ru & \bfseries sk & \bfseries sl & \bfseries uk & \bfseries zh & \bfseries all \\
\hline
\multirow[c]{6}{*}{\bfseries P} & \bfseries Discord & {\cellcolor[HTML]{000000}} \color[HTML]{000000} {\cellcolor{white}} N/A & {\cellcolor[HTML]{000000}} \color[HTML]{000000} {\cellcolor{white}} N/A & {\cellcolor[HTML]{94B6D7}} \color[HTML]{000000} 0.92 & {\cellcolor[HTML]{BCC7E1}} \color[HTML]{000000} 0.81 & {\cellcolor[HTML]{ABBFDC}} \color[HTML]{000000} 0.86 & {\cellcolor[HTML]{000000}} \color[HTML]{000000} {\cellcolor{white}} N/A & {\cellcolor[HTML]{99B8D8}} \color[HTML]{000000} 0.91 & {\cellcolor[HTML]{9FBAD9}} \color[HTML]{000000} 0.89 & {\cellcolor[HTML]{ACC0DD}} \color[HTML]{000000} 0.86 & {\cellcolor[HTML]{000000}} \color[HTML]{000000} {\cellcolor{white}} N/A & {\cellcolor[HTML]{000000}} \color[HTML]{000000} {\cellcolor{white}} N/A & {\cellcolor[HTML]{D1D2E6}} \color[HTML]{000000} 0.75 & {\cellcolor[HTML]{B1C2DE}} \color[HTML]{000000} 0.84 & {\cellcolor[HTML]{BCC7E1}} \color[HTML]{000000} 0.81 & {\cellcolor[HTML]{BDC8E1}} \color[HTML]{000000} 0.81 & {\cellcolor[HTML]{B9C6E0}} \color[HTML]{000000} 0.82 & {\cellcolor[HTML]{BDC8E1}} \color[HTML]{000000} 0.81 & {\cellcolor[HTML]{000000}} \color[HTML]{000000} {\cellcolor{white}} N/A & {\cellcolor[HTML]{000000}} \color[HTML]{000000} {\cellcolor{white}} N/A & {\cellcolor[HTML]{000000}} \color[HTML]{000000} {\cellcolor{white}} N/A & {\cellcolor[HTML]{000000}} \color[HTML]{000000} {\cellcolor{white}} N/A & {\cellcolor[HTML]{000000}} \color[HTML]{000000} {\cellcolor{white}} N/A & {\cellcolor[HTML]{B9C6E0}} \color[HTML]{000000} 0.82 \\
\bfseries  & \bfseries Gab & {\cellcolor[HTML]{000000}} \color[HTML]{000000} {\cellcolor{white}} N/A & {\cellcolor[HTML]{000000}} \color[HTML]{000000} {\cellcolor{white}} N/A & {\cellcolor[HTML]{000000}} \color[HTML]{000000} {\cellcolor{white}} N/A & {\cellcolor[HTML]{000000}} \color[HTML]{000000} {\cellcolor{white}} N/A & {\cellcolor[HTML]{DAD9EA}} \color[HTML]{000000} 0.71 & {\cellcolor[HTML]{000000}} \color[HTML]{000000} {\cellcolor{white}} N/A & {\cellcolor[HTML]{B8C6E0}} \color[HTML]{000000} 0.82 & {\cellcolor[HTML]{D9D8EA}} \color[HTML]{000000} 0.71 & {\cellcolor[HTML]{000000}} \color[HTML]{000000} {\cellcolor{white}} N/A & {\cellcolor[HTML]{000000}} \color[HTML]{000000} {\cellcolor{white}} N/A & {\cellcolor[HTML]{000000}} \color[HTML]{000000} {\cellcolor{white}} N/A & {\cellcolor[HTML]{000000}} \color[HTML]{000000} {\cellcolor{white}} N/A & {\cellcolor[HTML]{000000}} \color[HTML]{000000} {\cellcolor{white}} N/A & {\cellcolor[HTML]{D9D8EA}} \color[HTML]{000000} 0.71 & {\cellcolor[HTML]{E9E5F1}} \color[HTML]{000000} 0.64 & {\cellcolor[HTML]{D8D7E9}} \color[HTML]{000000} 0.72 & {\cellcolor[HTML]{EFE9F3}} \color[HTML]{000000} 0.61 & {\cellcolor[HTML]{EDE8F3}} \color[HTML]{000000} 0.62 & {\cellcolor[HTML]{000000}} \color[HTML]{000000} {\cellcolor{white}} N/A & {\cellcolor[HTML]{000000}} \color[HTML]{000000} {\cellcolor{white}} N/A & {\cellcolor[HTML]{000000}} \color[HTML]{000000} {\cellcolor{white}} N/A & {\cellcolor[HTML]{000000}} \color[HTML]{000000} {\cellcolor{white}} N/A & {\cellcolor[HTML]{E4E1EF}} \color[HTML]{000000} 0.66 \\
\bfseries  & \bfseries Telegram & {\cellcolor[HTML]{D4D4E8}} \color[HTML]{000000} 0.73 & {\cellcolor[HTML]{C8CDE4}} \color[HTML]{000000} 0.77 & {\cellcolor[HTML]{D4D4E8}} \color[HTML]{000000} 0.73 & {\cellcolor[HTML]{D4D4E8}} \color[HTML]{000000} 0.73 & {\cellcolor[HTML]{C2CBE2}} \color[HTML]{000000} 0.79 & {\cellcolor[HTML]{BBC7E0}} \color[HTML]{000000} 0.81 & {\cellcolor[HTML]{A9BFDC}} \color[HTML]{000000} 0.86 & {\cellcolor[HTML]{BFC9E1}} \color[HTML]{000000} 0.80 & {\cellcolor[HTML]{B4C4DF}} \color[HTML]{000000} 0.84 & {\cellcolor[HTML]{000000}} \color[HTML]{000000} {\cellcolor{white}} N/A & {\cellcolor[HTML]{000000}} \color[HTML]{000000} {\cellcolor{white}} N/A & {\cellcolor[HTML]{D9D8EA}} \color[HTML]{000000} 0.71 & {\cellcolor[HTML]{ACC0DD}} \color[HTML]{000000} 0.86 & {\cellcolor[HTML]{E0DDED}} \color[HTML]{000000} 0.68 & {\cellcolor[HTML]{DCDAEB}} \color[HTML]{000000} 0.70 & {\cellcolor[HTML]{C9CEE4}} \color[HTML]{000000} 0.77 & {\cellcolor[HTML]{C8CDE4}} \color[HTML]{000000} 0.77 & {\cellcolor[HTML]{CCCFE5}} \color[HTML]{000000} 0.76 & {\cellcolor[HTML]{000000}} \color[HTML]{000000} {\cellcolor{white}} N/A & {\cellcolor[HTML]{000000}} \color[HTML]{000000} {\cellcolor{white}} N/A & {\cellcolor[HTML]{D6D6E9}} \color[HTML]{000000} 0.73 & {\cellcolor[HTML]{CDD0E5}} \color[HTML]{000000} 0.76 & {\cellcolor[HTML]{D2D3E7}} \color[HTML]{000000} 0.74 \\
\bfseries  & \bfseries Twitter & {\cellcolor[HTML]{B3C3DE}} \color[HTML]{000000} 0.84 & {\cellcolor[HTML]{C1CAE2}} \color[HTML]{000000} 0.79 & {\cellcolor[HTML]{000000}} \color[HTML]{000000} {\cellcolor{white}} N/A & {\cellcolor[HTML]{000000}} \color[HTML]{000000} {\cellcolor{white}} N/A & {\cellcolor[HTML]{C8CDE4}} \color[HTML]{000000} 0.78 & {\cellcolor[HTML]{000000}} \color[HTML]{000000} {\cellcolor{white}} N/A & {\cellcolor[HTML]{ADC1DD}} \color[HTML]{000000} 0.85 & {\cellcolor[HTML]{D3D4E7}} \color[HTML]{000000} 0.73 & {\cellcolor[HTML]{000000}} \color[HTML]{000000} {\cellcolor{white}} N/A & {\cellcolor[HTML]{000000}} \color[HTML]{000000} {\cellcolor{white}} N/A & {\cellcolor[HTML]{000000}} \color[HTML]{000000} {\cellcolor{white}} N/A & {\cellcolor[HTML]{000000}} \color[HTML]{000000} {\cellcolor{white}} N/A & {\cellcolor[HTML]{000000}} \color[HTML]{000000} {\cellcolor{white}} N/A & {\cellcolor[HTML]{D4D4E8}} \color[HTML]{000000} 0.73 & {\cellcolor[HTML]{000000}} \color[HTML]{000000} {\cellcolor{white}} N/A & {\cellcolor[HTML]{D0D1E6}} \color[HTML]{000000} 0.75 & {\cellcolor[HTML]{E7E3F0}} \color[HTML]{000000} 0.65 & {\cellcolor[HTML]{BFC9E1}} \color[HTML]{000000} 0.80 & {\cellcolor[HTML]{000000}} \color[HTML]{000000} {\cellcolor{white}} N/A & {\cellcolor[HTML]{000000}} \color[HTML]{000000} {\cellcolor{white}} N/A & {\cellcolor[HTML]{000000}} \color[HTML]{000000} {\cellcolor{white}} N/A & {\cellcolor[HTML]{000000}} \color[HTML]{000000} {\cellcolor{white}} N/A & {\cellcolor[HTML]{D5D5E8}} \color[HTML]{000000} 0.73 \\
\bfseries  & \bfseries WhatsApp & {\cellcolor[HTML]{000000}} \color[HTML]{000000} {\cellcolor{white}} N/A & {\cellcolor[HTML]{000000}} \color[HTML]{000000} {\cellcolor{white}} N/A & {\cellcolor[HTML]{000000}} \color[HTML]{000000} {\cellcolor{white}} N/A & {\cellcolor[HTML]{000000}} \color[HTML]{000000} {\cellcolor{white}} N/A & {\cellcolor[HTML]{000000}} \color[HTML]{000000} {\cellcolor{white}} N/A & {\cellcolor[HTML]{000000}} \color[HTML]{000000} {\cellcolor{white}} N/A & {\cellcolor[HTML]{B8C6E0}} \color[HTML]{000000} 0.82 & {\cellcolor[HTML]{ABBFDC}} \color[HTML]{000000} 0.86 & {\cellcolor[HTML]{000000}} \color[HTML]{000000} {\cellcolor{white}} N/A & {\cellcolor[HTML]{000000}} \color[HTML]{000000} {\cellcolor{white}} N/A & {\cellcolor[HTML]{000000}} \color[HTML]{000000} {\cellcolor{white}} N/A & {\cellcolor[HTML]{000000}} \color[HTML]{000000} {\cellcolor{white}} N/A & {\cellcolor[HTML]{000000}} \color[HTML]{000000} {\cellcolor{white}} N/A & {\cellcolor[HTML]{000000}} \color[HTML]{000000} {\cellcolor{white}} N/A & {\cellcolor[HTML]{000000}} \color[HTML]{000000} {\cellcolor{white}} N/A & {\cellcolor[HTML]{CDD0E5}} \color[HTML]{000000} 0.76 & {\cellcolor[HTML]{000000}} \color[HTML]{000000} {\cellcolor{white}} N/A & {\cellcolor[HTML]{000000}} \color[HTML]{000000} {\cellcolor{white}} N/A & {\cellcolor[HTML]{000000}} \color[HTML]{000000} {\cellcolor{white}} N/A & {\cellcolor[HTML]{000000}} \color[HTML]{000000} {\cellcolor{white}} N/A & {\cellcolor[HTML]{000000}} \color[HTML]{000000} {\cellcolor{white}} N/A & {\cellcolor[HTML]{000000}} \color[HTML]{000000} {\cellcolor{white}} N/A & {\cellcolor[HTML]{C2CBE2}} \color[HTML]{000000} 0.79 \\
\bfseries  & \bfseries {all} & {\cellcolor[HTML]{CED0E6}} \color[HTML]{000000} 0.76 & {\cellcolor[HTML]{CACEE5}} \color[HTML]{000000} 0.77 & {\cellcolor[HTML]{C8CDE4}} \color[HTML]{000000} 0.77 & {\cellcolor[HTML]{D4D4E8}} \color[HTML]{000000} 0.73 & {\cellcolor[HTML]{C6CCE3}} \color[HTML]{000000} 0.78 & {\cellcolor[HTML]{CED0E6}} \color[HTML]{000000} 0.75 & {\cellcolor[HTML]{AFC1DD}} \color[HTML]{000000} 0.85 & {\cellcolor[HTML]{C0C9E2}} \color[HTML]{000000} 0.80 & {\cellcolor[HTML]{B7C5DF}} \color[HTML]{000000} 0.82 & {\cellcolor[HTML]{CED0E6}} \color[HTML]{000000} 0.75 & {\cellcolor[HTML]{000000}} \color[HTML]{000000} {\cellcolor{white}} N/A & {\cellcolor[HTML]{DBDAEB}} \color[HTML]{000000} 0.70 & {\cellcolor[HTML]{BCC7E1}} \color[HTML]{000000} 0.81 & {\cellcolor[HTML]{D7D6E9}} \color[HTML]{000000} 0.72 & {\cellcolor[HTML]{DAD9EA}} \color[HTML]{000000} 0.70 & {\cellcolor[HTML]{CDD0E5}} \color[HTML]{000000} 0.76 & {\cellcolor[HTML]{DAD9EA}} \color[HTML]{000000} 0.71 & {\cellcolor[HTML]{D8D7E9}} \color[HTML]{000000} 0.72 & {\cellcolor[HTML]{DCDAEB}} \color[HTML]{000000} 0.70 & {\cellcolor[HTML]{E7E3F0}} \color[HTML]{000000} 0.65 & {\cellcolor[HTML]{D7D6E9}} \color[HTML]{000000} 0.72 & {\cellcolor[HTML]{D1D2E6}} \color[HTML]{000000} 0.75 & {\cellcolor[HTML]{D3D4E7}} \color[HTML]{000000} 0.74 \\
\hline
\multirow[c]{6}{*}{\bfseries S} & \bfseries Discord & {\cellcolor[HTML]{000000}} \color[HTML]{000000} {\cellcolor{white}} N/A & {\cellcolor[HTML]{000000}} \color[HTML]{000000} {\cellcolor{white}} N/A & {\cellcolor[HTML]{AFC1DD}} \color[HTML]{000000} 0.85 & {\cellcolor[HTML]{9CB9D9}} \color[HTML]{000000} 0.90 & {\cellcolor[HTML]{A8BEDC}} \color[HTML]{000000} 0.87 & {\cellcolor[HTML]{000000}} \color[HTML]{000000} {\cellcolor{white}} N/A & {\cellcolor[HTML]{9EBAD9}} \color[HTML]{000000} 0.90 & {\cellcolor[HTML]{A1BBDA}} \color[HTML]{000000} 0.89 & {\cellcolor[HTML]{ACC0DD}} \color[HTML]{000000} 0.86 & {\cellcolor[HTML]{000000}} \color[HTML]{000000} {\cellcolor{white}} N/A & {\cellcolor[HTML]{000000}} \color[HTML]{000000} {\cellcolor{white}} N/A & {\cellcolor[HTML]{A1BBDA}} \color[HTML]{000000} 0.89 & {\cellcolor[HTML]{97B7D7}} \color[HTML]{000000} 0.91 & {\cellcolor[HTML]{A2BCDA}} \color[HTML]{000000} 0.89 & {\cellcolor[HTML]{94B6D7}} \color[HTML]{000000} 0.92 & {\cellcolor[HTML]{9CB9D9}} \color[HTML]{000000} 0.90 & {\cellcolor[HTML]{9FBAD9}} \color[HTML]{000000} 0.89 & {\cellcolor[HTML]{000000}} \color[HTML]{000000} {\cellcolor{white}} N/A & {\cellcolor[HTML]{000000}} \color[HTML]{000000} {\cellcolor{white}} N/A & {\cellcolor[HTML]{000000}} \color[HTML]{000000} {\cellcolor{white}} N/A & {\cellcolor[HTML]{000000}} \color[HTML]{000000} {\cellcolor{white}} N/A & {\cellcolor[HTML]{000000}} \color[HTML]{000000} {\cellcolor{white}} N/A & {\cellcolor[HTML]{A4BCDA}} \color[HTML]{000000} 0.88 \\
\bfseries  & \bfseries Gab & {\cellcolor[HTML]{000000}} \color[HTML]{000000} {\cellcolor{white}} N/A & {\cellcolor[HTML]{000000}} \color[HTML]{000000} {\cellcolor{white}} N/A & {\cellcolor[HTML]{000000}} \color[HTML]{000000} {\cellcolor{white}} N/A & {\cellcolor[HTML]{000000}} \color[HTML]{000000} {\cellcolor{white}} N/A & {\cellcolor[HTML]{D4D4E8}} \color[HTML]{000000} 0.73 & {\cellcolor[HTML]{000000}} \color[HTML]{000000} {\cellcolor{white}} N/A & {\cellcolor[HTML]{CDD0E5}} \color[HTML]{000000} 0.76 & {\cellcolor[HTML]{D2D2E7}} \color[HTML]{000000} 0.74 & {\cellcolor[HTML]{000000}} \color[HTML]{000000} {\cellcolor{white}} N/A & {\cellcolor[HTML]{000000}} \color[HTML]{000000} {\cellcolor{white}} N/A & {\cellcolor[HTML]{000000}} \color[HTML]{000000} {\cellcolor{white}} N/A & {\cellcolor[HTML]{000000}} \color[HTML]{000000} {\cellcolor{white}} N/A & {\cellcolor[HTML]{000000}} \color[HTML]{000000} {\cellcolor{white}} N/A & {\cellcolor[HTML]{D7D6E9}} \color[HTML]{000000} 0.72 & {\cellcolor[HTML]{C9CEE4}} \color[HTML]{000000} 0.77 & {\cellcolor[HTML]{D4D4E8}} \color[HTML]{000000} 0.73 & {\cellcolor[HTML]{D2D3E7}} \color[HTML]{000000} 0.74 & {\cellcolor[HTML]{D9D8EA}} \color[HTML]{000000} 0.71 & {\cellcolor[HTML]{000000}} \color[HTML]{000000} {\cellcolor{white}} N/A & {\cellcolor[HTML]{000000}} \color[HTML]{000000} {\cellcolor{white}} N/A & {\cellcolor[HTML]{000000}} \color[HTML]{000000} {\cellcolor{white}} N/A & {\cellcolor[HTML]{000000}} \color[HTML]{000000} {\cellcolor{white}} N/A & {\cellcolor[HTML]{DEDCEC}} \color[HTML]{000000} 0.69 \\
\bfseries  & \bfseries Telegram & {\cellcolor[HTML]{CDD0E5}} \color[HTML]{000000} 0.76 & {\cellcolor[HTML]{C6CCE3}} \color[HTML]{000000} 0.78 & {\cellcolor[HTML]{EDE8F3}} \color[HTML]{000000} 0.62 & {\cellcolor[HTML]{A4BCDA}} \color[HTML]{000000} 0.88 & {\cellcolor[HTML]{D9D8EA}} \color[HTML]{000000} 0.71 & {\cellcolor[HTML]{ABBFDC}} \color[HTML]{000000} 0.86 & {\cellcolor[HTML]{BDC8E1}} \color[HTML]{000000} 0.81 & {\cellcolor[HTML]{C9CEE4}} \color[HTML]{000000} 0.77 & {\cellcolor[HTML]{B1C2DE}} \color[HTML]{000000} 0.84 & {\cellcolor[HTML]{000000}} \color[HTML]{000000} {\cellcolor{white}} N/A & {\cellcolor[HTML]{000000}} \color[HTML]{000000} {\cellcolor{white}} N/A & {\cellcolor[HTML]{9FBAD9}} \color[HTML]{000000} 0.89 & {\cellcolor[HTML]{97B7D7}} \color[HTML]{000000} 0.91 & {\cellcolor[HTML]{D2D3E7}} \color[HTML]{000000} 0.74 & {\cellcolor[HTML]{B1C2DE}} \color[HTML]{000000} 0.84 & {\cellcolor[HTML]{B9C6E0}} \color[HTML]{000000} 0.82 & {\cellcolor[HTML]{A7BDDB}} \color[HTML]{000000} 0.87 & {\cellcolor[HTML]{D2D3E7}} \color[HTML]{000000} 0.74 & {\cellcolor[HTML]{000000}} \color[HTML]{000000} {\cellcolor{white}} N/A & {\cellcolor[HTML]{000000}} \color[HTML]{000000} {\cellcolor{white}} N/A & {\cellcolor[HTML]{D9D8EA}} \color[HTML]{000000} 0.71 & {\cellcolor[HTML]{D1D2E6}} \color[HTML]{000000} 0.75 & {\cellcolor[HTML]{D3D4E7}} \color[HTML]{000000} 0.74 \\
\bfseries  & \bfseries Twitter & {\cellcolor[HTML]{CCCFE5}} \color[HTML]{000000} 0.76 & {\cellcolor[HTML]{C2CBE2}} \color[HTML]{000000} 0.79 & {\cellcolor[HTML]{000000}} \color[HTML]{000000} {\cellcolor{white}} N/A & {\cellcolor[HTML]{000000}} \color[HTML]{000000} {\cellcolor{white}} N/A & {\cellcolor[HTML]{A7BDDB}} \color[HTML]{000000} 0.87 & {\cellcolor[HTML]{000000}} \color[HTML]{000000} {\cellcolor{white}} N/A & {\cellcolor[HTML]{B7C5DF}} \color[HTML]{000000} 0.83 & {\cellcolor[HTML]{D8D7E9}} \color[HTML]{000000} 0.72 & {\cellcolor[HTML]{000000}} \color[HTML]{000000} {\cellcolor{white}} N/A & {\cellcolor[HTML]{000000}} \color[HTML]{000000} {\cellcolor{white}} N/A & {\cellcolor[HTML]{000000}} \color[HTML]{000000} {\cellcolor{white}} N/A & {\cellcolor[HTML]{000000}} \color[HTML]{000000} {\cellcolor{white}} N/A & {\cellcolor[HTML]{000000}} \color[HTML]{000000} {\cellcolor{white}} N/A & {\cellcolor[HTML]{C8CDE4}} \color[HTML]{000000} 0.78 & {\cellcolor[HTML]{000000}} \color[HTML]{000000} {\cellcolor{white}} N/A & {\cellcolor[HTML]{A9BFDC}} \color[HTML]{000000} 0.86 & {\cellcolor[HTML]{AFC1DD}} \color[HTML]{000000} 0.85 & {\cellcolor[HTML]{9AB8D8}} \color[HTML]{000000} 0.90 & {\cellcolor[HTML]{000000}} \color[HTML]{000000} {\cellcolor{white}} N/A & {\cellcolor[HTML]{000000}} \color[HTML]{000000} {\cellcolor{white}} N/A & {\cellcolor[HTML]{000000}} \color[HTML]{000000} {\cellcolor{white}} N/A & {\cellcolor[HTML]{000000}} \color[HTML]{000000} {\cellcolor{white}} N/A & {\cellcolor[HTML]{D2D2E7}} \color[HTML]{000000} 0.74 \\
\bfseries  & \bfseries WhatsApp & {\cellcolor[HTML]{000000}} \color[HTML]{000000} {\cellcolor{white}} N/A & {\cellcolor[HTML]{000000}} \color[HTML]{000000} {\cellcolor{white}} N/A & {\cellcolor[HTML]{000000}} \color[HTML]{000000} {\cellcolor{white}} N/A & {\cellcolor[HTML]{000000}} \color[HTML]{000000} {\cellcolor{white}} N/A & {\cellcolor[HTML]{000000}} \color[HTML]{000000} {\cellcolor{white}} N/A & {\cellcolor[HTML]{000000}} \color[HTML]{000000} {\cellcolor{white}} N/A & {\cellcolor[HTML]{DEDCEC}} \color[HTML]{000000} 0.69 & {\cellcolor[HTML]{ADC1DD}} \color[HTML]{000000} 0.85 & {\cellcolor[HTML]{000000}} \color[HTML]{000000} {\cellcolor{white}} N/A & {\cellcolor[HTML]{000000}} \color[HTML]{000000} {\cellcolor{white}} N/A & {\cellcolor[HTML]{000000}} \color[HTML]{000000} {\cellcolor{white}} N/A & {\cellcolor[HTML]{000000}} \color[HTML]{000000} {\cellcolor{white}} N/A & {\cellcolor[HTML]{000000}} \color[HTML]{000000} {\cellcolor{white}} N/A & {\cellcolor[HTML]{000000}} \color[HTML]{000000} {\cellcolor{white}} N/A & {\cellcolor[HTML]{000000}} \color[HTML]{000000} {\cellcolor{white}} N/A & {\cellcolor[HTML]{C6CCE3}} \color[HTML]{000000} 0.78 & {\cellcolor[HTML]{000000}} \color[HTML]{000000} {\cellcolor{white}} N/A & {\cellcolor[HTML]{000000}} \color[HTML]{000000} {\cellcolor{white}} N/A & {\cellcolor[HTML]{000000}} \color[HTML]{000000} {\cellcolor{white}} N/A & {\cellcolor[HTML]{000000}} \color[HTML]{000000} {\cellcolor{white}} N/A & {\cellcolor[HTML]{000000}} \color[HTML]{000000} {\cellcolor{white}} N/A & {\cellcolor[HTML]{000000}} \color[HTML]{000000} {\cellcolor{white}} N/A & {\cellcolor[HTML]{D7D6E9}} \color[HTML]{000000} 0.72 \\
\bfseries  & \bfseries {all} & {\cellcolor[HTML]{C9CEE4}} \color[HTML]{000000} 0.77 & {\cellcolor[HTML]{C6CCE3}} \color[HTML]{000000} 0.78 & {\cellcolor[HTML]{E6E2EF}} \color[HTML]{000000} 0.65 & {\cellcolor[HTML]{A9BFDC}} \color[HTML]{000000} 0.87 & {\cellcolor[HTML]{C9CEE4}} \color[HTML]{000000} 0.77 & {\cellcolor[HTML]{B8C6E0}} \color[HTML]{000000} 0.82 & {\cellcolor[HTML]{C1CAE2}} \color[HTML]{000000} 0.79 & {\cellcolor[HTML]{C2CBE2}} \color[HTML]{000000} 0.79 & {\cellcolor[HTML]{B9C6E0}} \color[HTML]{000000} 0.82 & {\cellcolor[HTML]{CCCFE5}} \color[HTML]{000000} 0.76 & {\cellcolor[HTML]{000000}} \color[HTML]{000000} {\cellcolor{white}} N/A & {\cellcolor[HTML]{AFC1DD}} \color[HTML]{000000} 0.85 & {\cellcolor[HTML]{A7BDDB}} \color[HTML]{000000} 0.87 & {\cellcolor[HTML]{CCCFE5}} \color[HTML]{000000} 0.76 & {\cellcolor[HTML]{B4C4DF}} \color[HTML]{000000} 0.83 & {\cellcolor[HTML]{BDC8E1}} \color[HTML]{000000} 0.81 & {\cellcolor[HTML]{B5C4DF}} \color[HTML]{000000} 0.83 & {\cellcolor[HTML]{CDD0E5}} \color[HTML]{000000} 0.76 & {\cellcolor[HTML]{C1CAE2}} \color[HTML]{000000} 0.80 & {\cellcolor[HTML]{C9CEE4}} \color[HTML]{000000} 0.77 & {\cellcolor[HTML]{D9D8EA}} \color[HTML]{000000} 0.71 & {\cellcolor[HTML]{CCCFE5}} \color[HTML]{000000} 0.76 & {\cellcolor[HTML]{D2D3E7}} \color[HTML]{000000} 0.74 \\
\hline
\end{tabular}
}
%\vspace{-1mm}
\caption{Per-platform mean AUC ROC performance of well-performing zero-shot MGT detectors per category. N/A refers to not enough samples (at least 2000) in MultiSocial for a combination of language and platform. Discord data are the easiest for the detection, Gab data are the most difficult.}
\label{tab:data_zeroshot_mean}
%\vspace{-3mm}
\end{table*}

For comparison of MGTD methods, we use \textbf{AUC ROC} (area under the curve of receiver operating characteristic) as a classification-threshold independent metric (not affected by a threshold calibration on in-domain data), also used by \citep{hans2024spotting}. Due to imbalanced test data (machine class contains 7x more samples), we also use \textbf{Macro avg. F1-score @ 5\% FPR} (false positive rate) as a metric balancing between a precision and a recall of the classification, while the threshold is calibrated based on the train data (to avoid data leakage) ROC curve to achieve 5\% FPR (similarly used in \citealp{dugan2024raid}).

Since Gemini-generated data have used a slightly different generation process (e.g., jailbreak prompt, see Appendix~\ref{sec:datacreation}) and achieved highly outlier word-count and unique-words scores in Table~\ref{tab:data_stats_generated}, we do not use Gemini-generated data in the training (fine-tuning or classification-threshold calibration). Nevertheless, they are still included in the evaluation (can serve for unseen generator evaluation).

%\vspace{-1.5mm}
\subsection{Multilingual Zero-shot Detection}
%\vspace{-0.5mm}

This experiment is focused on the following research question: \textit{\textbf{RQ1:} How well are social-media texts of multiple languages and platforms detectable by MGT detectors applicable in zero-shot manner (out-of-distribution, without further in-domain training)?}
Since SMN texts are usually shorter than commonly used news articles, the detection performance of existing directly usable (i.e., without in-domain training) MGT detectors is still unknown (it could differ from the reported performance on other domains). This has not been evaluated in the multilingual settings. Are there differences in multilingual MGT detection among different sources of SMN content (e.g., Twitter vs. Telegram vs. Gab)? Is there a difference between statistical (could be language independent) and pre-trained (heavily dependent on pre-training languages) zero-shot detectors?

To answer these questions, we compare the per-language performance of statistical and pre-trained MGTD methods (collectively called zero-shot methods for this purpose) based on AUC ROC (to avoid in-domain touch with the data) in Table~\ref{tab:data_zeroshot}. To compare per-language performance per platforms, we consider only cases where there are at least 2000 samples available per platform and language (approximately 250 texts per each generator). The summarized results of the comparison are provided in Table~\ref{tab:data_zeroshot_mean}. The \textit{all} row represents performance of MGT detectors of the corresponding category for all platforms data combined (excluding only results for Scottish due to not having enough samples). Due to low performance of three out of five selected pre-trained detectors (see Table~\ref{tab:benchmark}), we average results for this category only for the two well-performing detectors (BLOOMZ-3b-mixed-Detector and ChatGPT-Detector-RoBERTa-Chinese). Full results are available in Appendix~\ref{sec:data}.

To evaluate whether the differences between mean AUC ROC of statistical and the best pre-trained MGT detectors are statistically significant, we conduct paired t-tests for each test language and check whether p-value is < 0.05.
%We observe that only for Catalan, Scottish, and Slovenian, the differences are statistically significant (p-value < 0.05).
Analogously, we have verified significance of differences between per-platform means in each category.
%Most of differences in statistical category are statistically significant; on the other hand, Twitter and WhatsApp to Gab differences are statistically significant in pre-trained category.

\textbf{There are differences in performance of SOTA zero-shot MGTD methods on texts of English and non-English languages.} When considering the well-performing pre-trained and all statistical detectors, the difference between performances on English and combined non-English texts is statistically significant (higher on English). However, Longformer Detector clearly performed better on English than the other languages. Similarly, ruRoBERTa performed better on Russian, Ukrainian, and Bulgarian than the others. OpenAI Detector has clearly not been trained on SMN texts, since not performing well even in English (there are also huge differences in performance based on the generators, see Table~\ref{tab:data_perllm_pretrained}). The other two pre-trained and all statistical detectors performed similarly across languages, although the Chinese detector performed worse on Slavic-Latin languages.

\begin{table*}[!t]
\centering
\resizebox{\textwidth}{!}{
\addtolength{\tabcolsep}{-2pt}
\begin{tabular}{p{3cm}|cccccccccccccccccccccc|c}
\hline
 & \multicolumn{23}{c}{\bfseries Test Language [AUC ROC]} \\
\bfseries Detector & \bfseries ar & \bfseries bg & \bfseries ca & \bfseries cs & \bfseries de & \bfseries el & \bfseries en & \bfseries es & \bfseries et & \bfseries ga$\star$ & \bfseries gd$\star$ & \bfseries hr & \bfseries hu & \bfseries nl & \bfseries pl & \bfseries pt & \bfseries ro & \bfseries ru & \bfseries sk$\star$ & \bfseries sl$\star$ & \bfseries uk & \bfseries zh & \bfseries all \\
\hline
\bfseries Aya-101-MultiSocial & {\cellcolor[HTML]{81AED2}} \color[HTML]{000000} 0.97 & {\cellcolor[HTML]{78ABD0}} \color[HTML]{000000} 0.99 & {\cellcolor[HTML]{7EADD1}} \color[HTML]{000000} 0.97 & {\cellcolor[HTML]{7BACD1}} \color[HTML]{000000} 0.98 & {\cellcolor[HTML]{80AED2}} \color[HTML]{000000} 0.97 & {\cellcolor[HTML]{80AED2}} \color[HTML]{000000} 0.97 & {\cellcolor[HTML]{7EADD1}} \color[HTML]{000000} 0.98 & {\cellcolor[HTML]{7DACD1}} \color[HTML]{000000} 0.98 & {\cellcolor[HTML]{79ABD0}} \color[HTML]{000000} 0.98 & {\cellcolor[HTML]{89B1D4}} \color[HTML]{000000} 0.95 & {\cellcolor[HTML]{93B5D6}} \color[HTML]{000000} 0.92 & {\cellcolor[HTML]{7BACD1}} \color[HTML]{000000} 0.98 & {\cellcolor[HTML]{76AAD0}} \color[HTML]{000000} 0.99 & {\cellcolor[HTML]{80AED2}} \color[HTML]{000000} 0.97 & {\cellcolor[HTML]{7DACD1}} \color[HTML]{000000} 0.98 & {\cellcolor[HTML]{7EADD1}} \color[HTML]{000000} 0.98 & {\cellcolor[HTML]{7DACD1}} \color[HTML]{000000} 0.98 & {\cellcolor[HTML]{86B0D3}} \color[HTML]{000000} 0.96 & {\cellcolor[HTML]{7DACD1}} \color[HTML]{000000} 0.98 & {\cellcolor[HTML]{89B1D4}} \color[HTML]{000000} 0.95 & {\cellcolor[HTML]{88B1D4}} \color[HTML]{000000} 0.95 & {\cellcolor[HTML]{7EADD1}} \color[HTML]{000000} 0.97 & {\cellcolor[HTML]{7EADD1}} \color[HTML]{000000} 0.97 \\
\bfseries BLOOMZ-3b-MultiSocial & {\cellcolor[HTML]{84B0D3}} \color[HTML]{000000} 0.96 & {\cellcolor[HTML]{7EADD1}} \color[HTML]{000000} 0.98 & {\cellcolor[HTML]{83AFD3}} \color[HTML]{000000} 0.96 & {\cellcolor[HTML]{80AED2}} \color[HTML]{000000} 0.97 & {\cellcolor[HTML]{88B1D4}} \color[HTML]{000000} 0.95 & {\cellcolor[HTML]{84B0D3}} \color[HTML]{000000} 0.96 & {\cellcolor[HTML]{7DACD1}} \color[HTML]{000000} 0.98 & {\cellcolor[HTML]{7EADD1}} \color[HTML]{000000} 0.97 & {\cellcolor[HTML]{7DACD1}} \color[HTML]{000000} 0.98 & {\cellcolor[HTML]{9AB8D8}} \color[HTML]{000000} 0.90 & {\cellcolor[HTML]{BBC7E0}} \color[HTML]{000000} 0.82 & {\cellcolor[HTML]{84B0D3}} \color[HTML]{000000} 0.96 & {\cellcolor[HTML]{78ABD0}} \color[HTML]{000000} 0.99 & {\cellcolor[HTML]{8CB3D5}} \color[HTML]{000000} 0.94 & {\cellcolor[HTML]{86B0D3}} \color[HTML]{000000} 0.95 & {\cellcolor[HTML]{80AED2}} \color[HTML]{000000} 0.97 & {\cellcolor[HTML]{89B1D4}} \color[HTML]{000000} 0.95 & {\cellcolor[HTML]{8CB3D5}} \color[HTML]{000000} 0.94 & {\cellcolor[HTML]{86B0D3}} \color[HTML]{000000} 0.95 & {\cellcolor[HTML]{A5BDDB}} \color[HTML]{000000} 0.88 & {\cellcolor[HTML]{9EBAD9}} \color[HTML]{000000} 0.90 & {\cellcolor[HTML]{80AED2}} \color[HTML]{000000} 0.97 & {\cellcolor[HTML]{84B0D3}} \color[HTML]{000000} 0.96 \\
\bfseries Falcon-rw-1b-MultiSocial & {\cellcolor[HTML]{89B1D4}} \color[HTML]{000000} 0.95 & {\cellcolor[HTML]{7DACD1}} \color[HTML]{000000} 0.98 & {\cellcolor[HTML]{81AED2}} \color[HTML]{000000} 0.97 & {\cellcolor[HTML]{7EADD1}} \color[HTML]{000000} 0.97 & {\cellcolor[HTML]{86B0D3}} \color[HTML]{000000} 0.96 & {\cellcolor[HTML]{83AFD3}} \color[HTML]{000000} 0.96 & {\cellcolor[HTML]{7DACD1}} \color[HTML]{000000} 0.98 & {\cellcolor[HTML]{83AFD3}} \color[HTML]{000000} 0.96 & {\cellcolor[HTML]{7DACD1}} \color[HTML]{000000} 0.98 & {\cellcolor[HTML]{93B5D6}} \color[HTML]{000000} 0.92 & {\cellcolor[HTML]{A7BDDB}} \color[HTML]{000000} 0.87 & {\cellcolor[HTML]{83AFD3}} \color[HTML]{000000} 0.96 & {\cellcolor[HTML]{78ABD0}} \color[HTML]{000000} 0.99 & {\cellcolor[HTML]{88B1D4}} \color[HTML]{000000} 0.95 & {\cellcolor[HTML]{84B0D3}} \color[HTML]{000000} 0.96 & {\cellcolor[HTML]{84B0D3}} \color[HTML]{000000} 0.96 & {\cellcolor[HTML]{86B0D3}} \color[HTML]{000000} 0.95 & {\cellcolor[HTML]{8EB3D5}} \color[HTML]{000000} 0.94 & {\cellcolor[HTML]{86B0D3}} \color[HTML]{000000} 0.95 & {\cellcolor[HTML]{A8BEDC}} \color[HTML]{000000} 0.87 & {\cellcolor[HTML]{9AB8D8}} \color[HTML]{000000} 0.91 & {\cellcolor[HTML]{84B0D3}} \color[HTML]{000000} 0.96 & {\cellcolor[HTML]{84B0D3}} \color[HTML]{000000} 0.96 \\
\bfseries Llama-3-8b-MultiSocial & {\cellcolor[HTML]{81AED2}} \color[HTML]{000000} 0.97 & {\cellcolor[HTML]{78ABD0}} \color[HTML]{000000} 0.99 & {\cellcolor[HTML]{7BACD1}} \color[HTML]{000000} 0.98 & {\cellcolor[HTML]{79ABD0}} \color[HTML]{000000} 0.99 & {\cellcolor[HTML]{7DACD1}} \color[HTML]{000000} 0.98 & {\cellcolor[HTML]{81AED2}} \color[HTML]{000000} 0.97 & {\cellcolor[HTML]{79ABD0}} \color[HTML]{000000} 0.99 & {\cellcolor[HTML]{7BACD1}} \color[HTML]{000000} 0.98 & {\cellcolor[HTML]{79ABD0}} \color[HTML]{000000} 0.99 & {\cellcolor[HTML]{8BB2D4}} \color[HTML]{000000} 0.94 & {\cellcolor[HTML]{9EBAD9}} \color[HTML]{000000} 0.90 & {\cellcolor[HTML]{79ABD0}} \color[HTML]{000000} 0.98 & {\cellcolor[HTML]{76AAD0}} \color[HTML]{000000} 0.99 & {\cellcolor[HTML]{7EADD1}} \color[HTML]{000000} 0.98 & {\cellcolor[HTML]{7BACD1}} \color[HTML]{000000} 0.98 & {\cellcolor[HTML]{7DACD1}} \color[HTML]{000000} 0.98 & {\cellcolor[HTML]{7EADD1}} \color[HTML]{000000} 0.98 & {\cellcolor[HTML]{81AED2}} \color[HTML]{000000} 0.96 & {\cellcolor[HTML]{7EADD1}} \color[HTML]{000000} 0.98 & {\cellcolor[HTML]{89B1D4}} \color[HTML]{000000} 0.95 & {\cellcolor[HTML]{86B0D3}} \color[HTML]{000000} 0.95 & {\cellcolor[HTML]{7BACD1}} \color[HTML]{000000} 0.98 & {\cellcolor[HTML]{7DACD1}} \color[HTML]{000000} 0.98 \\
\bfseries Mistral-7b-v0.1-MultiSocial & {\cellcolor[HTML]{81AED2}} \color[HTML]{000000} 0.97 & {\cellcolor[HTML]{78ABD0}} \color[HTML]{000000} 0.99 & {\cellcolor[HTML]{7DACD1}} \color[HTML]{000000} 0.98 & {\cellcolor[HTML]{78ABD0}} \color[HTML]{000000} 0.99 & {\cellcolor[HTML]{7EADD1}} \color[HTML]{000000} 0.98 & {\cellcolor[HTML]{80AED2}} \color[HTML]{000000} 0.97 & {\cellcolor[HTML]{79ABD0}} \color[HTML]{000000} 0.99 & {\cellcolor[HTML]{7BACD1}} \color[HTML]{000000} 0.98 & {\cellcolor[HTML]{7BACD1}} \color[HTML]{000000} 0.98 & {\cellcolor[HTML]{8FB4D6}} \color[HTML]{000000} 0.93 & {\cellcolor[HTML]{91B5D6}} \color[HTML]{000000} 0.93 & {\cellcolor[HTML]{79ABD0}} \color[HTML]{000000} 0.99 & {\cellcolor[HTML]{76AAD0}} \color[HTML]{000000} 1.00 & {\cellcolor[HTML]{7EADD1}} \color[HTML]{000000} 0.97 & {\cellcolor[HTML]{7BACD1}} \color[HTML]{000000} 0.98 & {\cellcolor[HTML]{7DACD1}} \color[HTML]{000000} 0.98 & {\cellcolor[HTML]{7EADD1}} \color[HTML]{000000} 0.97 & {\cellcolor[HTML]{81AED2}} \color[HTML]{000000} 0.97 & {\cellcolor[HTML]{7EADD1}} \color[HTML]{000000} 0.97 & {\cellcolor[HTML]{8BB2D4}} \color[HTML]{000000} 0.94 & {\cellcolor[HTML]{84B0D3}} \color[HTML]{000000} 0.96 & {\cellcolor[HTML]{7DACD1}} \color[HTML]{000000} 0.98 & {\cellcolor[HTML]{7DACD1}} \color[HTML]{000000} 0.98 \\
\bfseries XLM-RoBERTa-large-MultiSocial & {\cellcolor[HTML]{88B1D4}} \color[HTML]{000000} 0.95 & {\cellcolor[HTML]{7BACD1}} \color[HTML]{000000} 0.98 & {\cellcolor[HTML]{8CB3D5}} \color[HTML]{000000} 0.94 & {\cellcolor[HTML]{7EADD1}} \color[HTML]{000000} 0.98 & {\cellcolor[HTML]{86B0D3}} \color[HTML]{000000} 0.96 & {\cellcolor[HTML]{86B0D3}} \color[HTML]{000000} 0.95 & {\cellcolor[HTML]{81AED2}} \color[HTML]{000000} 0.96 & {\cellcolor[HTML]{83AFD3}} \color[HTML]{000000} 0.96 & {\cellcolor[HTML]{81AED2}} \color[HTML]{000000} 0.97 & {\cellcolor[HTML]{A5BDDB}} \color[HTML]{000000} 0.88 & {\cellcolor[HTML]{C6CCE3}} \color[HTML]{000000} 0.78 & {\cellcolor[HTML]{81AED2}} \color[HTML]{000000} 0.97 & {\cellcolor[HTML]{79ABD0}} \color[HTML]{000000} 0.99 & {\cellcolor[HTML]{89B1D4}} \color[HTML]{000000} 0.95 & {\cellcolor[HTML]{81AED2}} \color[HTML]{000000} 0.97 & {\cellcolor[HTML]{86B0D3}} \color[HTML]{000000} 0.95 & {\cellcolor[HTML]{83AFD3}} \color[HTML]{000000} 0.96 & {\cellcolor[HTML]{89B1D4}} \color[HTML]{000000} 0.95 & {\cellcolor[HTML]{83AFD3}} \color[HTML]{000000} 0.96 & {\cellcolor[HTML]{99B8D8}} \color[HTML]{000000} 0.91 & {\cellcolor[HTML]{93B5D6}} \color[HTML]{000000} 0.92 & {\cellcolor[HTML]{8EB3D5}} \color[HTML]{000000} 0.93 & {\cellcolor[HTML]{86B0D3}} \color[HTML]{000000} 0.96 \\
\bfseries mDeBERTa-v3-base-MultiSocial & {\cellcolor[HTML]{8CB3D5}} \color[HTML]{000000} 0.94 & {\cellcolor[HTML]{7BACD1}} \color[HTML]{000000} 0.98 & {\cellcolor[HTML]{8BB2D4}} \color[HTML]{000000} 0.94 & {\cellcolor[HTML]{80AED2}} \color[HTML]{000000} 0.97 & {\cellcolor[HTML]{88B1D4}} \color[HTML]{000000} 0.95 & {\cellcolor[HTML]{8BB2D4}} \color[HTML]{000000} 0.94 & {\cellcolor[HTML]{83AFD3}} \color[HTML]{000000} 0.96 & {\cellcolor[HTML]{84B0D3}} \color[HTML]{000000} 0.96 & {\cellcolor[HTML]{7EADD1}} \color[HTML]{000000} 0.98 & {\cellcolor[HTML]{9EBAD9}} \color[HTML]{000000} 0.90 & {\cellcolor[HTML]{C4CBE3}} \color[HTML]{000000} 0.79 & {\cellcolor[HTML]{81AED2}} \color[HTML]{000000} 0.97 & {\cellcolor[HTML]{78ABD0}} \color[HTML]{000000} 0.99 & {\cellcolor[HTML]{88B1D4}} \color[HTML]{000000} 0.95 & {\cellcolor[HTML]{83AFD3}} \color[HTML]{000000} 0.96 & {\cellcolor[HTML]{84B0D3}} \color[HTML]{000000} 0.96 & {\cellcolor[HTML]{84B0D3}} \color[HTML]{000000} 0.96 & {\cellcolor[HTML]{8EB3D5}} \color[HTML]{000000} 0.93 & {\cellcolor[HTML]{84B0D3}} \color[HTML]{000000} 0.96 & {\cellcolor[HTML]{94B6D7}} \color[HTML]{000000} 0.92 & {\cellcolor[HTML]{91B5D6}} \color[HTML]{000000} 0.93 & {\cellcolor[HTML]{8BB2D4}} \color[HTML]{000000} 0.94 & {\cellcolor[HTML]{86B0D3}} \color[HTML]{000000} 0.95 \\
\hline
\end{tabular}
}
%\vspace{-2mm}
\caption{Per-language AUC ROC performance of fine-tuned MGT detectors. $\star$ marks languages not in train set. Larger models achieve better performance.}
\label{tab:data_finetuned}
%\vspace{-3mm}
\end{table*}

\textbf{There are significant differences in performance of SOTA zero-shot MGTD methods on texts of different platforms.} The detectors are able to better detect MGT of Discord SMN than the others (although the differences to other platforms are not statistically significant for pre-trained detectors). On the other hand, the Gab texts are the most difficult for them to classify (although the differences to Telegram for pre-trained and WhatsApp for statistical detectors are not statistically significant). There is no clear indication for the length of such texts affecting these results, since Discord has the lowest (9) and WhatsApp and Twitter the highest (18) median value of word-count text length. We can speculate that since Gab is known to have more toxic content (vulgarisms, hate speech), it can be more difficult for detection.

\textbf{There are negligible differences in performance of SOTA zero-shot statistical and best pre-trained MGTD methods.} When considering the two best performing pre-trained detectors, which achieved 0.72-0.76 AUC ROC (0.62-0.9 in per-language evaluation), the performance is competitive with the statistical detectors, achieving 0.72-0.75 AUC ROC (0.61-0.94 in per-language evaluation). In regard to multilingual performance, the statistical detectors tends to achieve higher performance for Slavic-Latin and Uralic languages (confirmed by Telegram-only data), under-performing for Catalan. This is not the case of pre-trained detectors, under-performing for Scottish and Slovenian, and the Chinese detector shows rather opposite patterns for Slavic-Latin languages. The t-tests confirmed that the differences between these two categories are statistically significant only for Catalan, Scottish and Slovenian languages.

\begin{table*}[!t]
\centering
\resizebox{\textwidth}{!}{
\addtolength{\tabcolsep}{-2pt}
\begin{tabular}{c|p{1cm}p{1cm}p{1cm}p{1cm}p{1cm}p{1cm}p{1cm}p{1cm}p{1cm}p{0.6cm}p{0.6cm}p{1cm}p{1cm}p{1cm}p{1cm}p{1cm}p{1cm}p{1cm}p{0.6cm}p{0.6cm}p{1cm}p{1cm}|p{1cm}}
\hline
\bfseries Train & \multicolumn{23}{c}{\bfseries Test Language [AUC ROC mean (±confidence interval)]} \\
\bfseries Language & \bfseries ar & \bfseries bg & \bfseries ca & \bfseries cs & \bfseries de & \bfseries el & \bfseries en & \bfseries es & \bfseries et & \bfseries ga & \bfseries gd & \bfseries hr & \bfseries hu & \bfseries nl & \bfseries pl & \bfseries pt & \bfseries ro & \bfseries ru & \bfseries sk & \bfseries sl & \bfseries uk & \bfseries zh & \bfseries all \\
\hline
\bfseries en & {\cellcolor[HTML]{BCC7E1}} \textcolor{black}{0.81 (±0.05)} & {\cellcolor[HTML]{9CB9D9}} \textcolor{black}{0.90 (±0.08)} & {\cellcolor[HTML]{C6CCE3}} \textcolor{black}{0.78 (±0.03)} & {\cellcolor[HTML]{99B8D8}} \textcolor{black}{0.91 (±0.05)} & {\cellcolor[HTML]{A5BDDB}} \textcolor{black}{0.88 (±0.02)} & {\cellcolor[HTML]{9AB8D8}} \textcolor{black}{0.90 (±0.03)} & {\cellcolor[HTML]{84B0D3}} \textcolor{black}{\bfseries 0.96 (±0.01)} & {\cellcolor[HTML]{A8BEDC}} \textcolor{black}{0.87 (±0.06)} & {\cellcolor[HTML]{96B6D7}} \textcolor{black}{0.92 (±0.03)} & N/A & N/A & {\cellcolor[HTML]{8CB3D5}} \textcolor{black}{0.94 (±0.03)} & {\cellcolor[HTML]{81AED2}} \textcolor{black}{0.97 (±0.03)} & {\cellcolor[HTML]{B1C2DE}} \textcolor{black}{0.84 (±0.02)} & {\cellcolor[HTML]{A1BBDA}} \textcolor{black}{0.89 (±0.05)} & {\cellcolor[HTML]{94B6D7}} \textcolor{black}{0.92 (±0.02)} & {\cellcolor[HTML]{8CB3D5}} \textcolor{black}{0.94 (±0.02)} & {\cellcolor[HTML]{A7BDDB}} \textcolor{black}{0.87 (±0.05)} & N/A & N/A & {\cellcolor[HTML]{BBC7E0}} \textcolor{black}{0.81 (±0.05)} & {\cellcolor[HTML]{D5D5E8}} \textcolor{black}{0.73 (±0.14)} & {\cellcolor[HTML]{A8BEDC}} \textcolor{black}{0.87 (±0.04)} \\
\bfseries es & {\cellcolor[HTML]{B8C6E0}} \textcolor{black}{0.82 (±0.05)} & {\cellcolor[HTML]{A1BBDA}} \textcolor{black}{0.89 (±0.08)} & {\cellcolor[HTML]{B0C2DE}} \textcolor{black}{0.85 (±0.03)} & {\cellcolor[HTML]{A2BCDA}} \textcolor{black}{0.89 (±0.06)} & {\cellcolor[HTML]{9FBAD9}} \textcolor{black}{0.89 (±0.03)} & {\cellcolor[HTML]{A7BDDB}} \textcolor{black}{0.87 (±0.06)} & {\cellcolor[HTML]{9CB9D9}} \textcolor{black}{0.90 (±0.04)} & {\cellcolor[HTML]{8CB3D5}} \textcolor{black}{\bfseries 0.94 (±0.01)} & {\cellcolor[HTML]{9EBAD9}} \textcolor{black}{0.90 (±0.05)} & N/A & N/A & {\cellcolor[HTML]{96B6D7}} \textcolor{black}{0.92 (±0.05)} & {\cellcolor[HTML]{88B1D4}} \textcolor{black}{0.95 (±0.04)} & {\cellcolor[HTML]{B3C3DE}} \textcolor{black}{0.84 (±0.03)} & {\cellcolor[HTML]{A5BDDB}} \textcolor{black}{0.88 (±0.06)} & {\cellcolor[HTML]{93B5D6}} \textcolor{black}{0.93 (±0.02)} & {\cellcolor[HTML]{8FB4D6}} \textcolor{black}{0.93 (±0.03)} & {\cellcolor[HTML]{A2BCDA}} \textcolor{black}{0.88 (±0.04)} & N/A & N/A & {\cellcolor[HTML]{B8C6E0}} \textcolor{black}{0.82 (±0.05)} & {\cellcolor[HTML]{D4D4E8}} \textcolor{black}{0.73 (±0.14)} & {\cellcolor[HTML]{A9BFDC}} \textcolor{black}{0.86 (±0.05)} \\
\bfseries ru & {\cellcolor[HTML]{BDC8E1}} \textcolor{black}{0.81 (±0.10)} & {\cellcolor[HTML]{8FB4D6}} \textcolor{black}{0.93 (±0.05)} & {\cellcolor[HTML]{CED0E6}} \textcolor{black}{0.76 (±0.10)} & {\cellcolor[HTML]{A7BDDB}} \textcolor{black}{0.87 (±0.08)} & {\cellcolor[HTML]{B1C2DE}} \textcolor{black}{0.84 (±0.04)} & {\cellcolor[HTML]{A4BCDA}} \textcolor{black}{0.88 (±0.06)} & {\cellcolor[HTML]{A7BDDB}} \textcolor{black}{0.87 (±0.06)} & {\cellcolor[HTML]{B8C6E0}} \textcolor{black}{0.82 (±0.11)} & {\cellcolor[HTML]{A5BDDB}} \textcolor{black}{0.88 (±0.08)} & N/A & N/A & {\cellcolor[HTML]{9FBAD9}} \textcolor{black}{0.89 (±0.07)} & {\cellcolor[HTML]{97B7D7}} \textcolor{black}{0.91 (±0.07)} & {\cellcolor[HTML]{C1CAE2}} \textcolor{black}{0.79 (±0.06)} & {\cellcolor[HTML]{A7BDDB}} \textcolor{black}{0.87 (±0.07)} & {\cellcolor[HTML]{ABBFDC}} \textcolor{black}{0.86 (±0.07)} & {\cellcolor[HTML]{A4BCDA}} \textcolor{black}{0.88 (±0.07)} & {\cellcolor[HTML]{8CB3D5}} \textcolor{black}{\bfseries 0.94 (±0.02)} & N/A & N/A & {\cellcolor[HTML]{9FBAD9}} \textcolor{black}{0.89 (±0.04)} & {\cellcolor[HTML]{D4D4E8}} \textcolor{black}{0.73 (±0.17)} & {\cellcolor[HTML]{B1C2DE}} \textcolor{black}{0.84 (±0.08)} \\
\hline
\bfseries {en-es-ru} & {\cellcolor[HTML]{A1BBDA}} \textcolor{black}{\bfseries 0.89 (±0.03)} & {\cellcolor[HTML]{8BB2D4}} \textcolor{black}{\bfseries 0.94 (±0.04)} & {\cellcolor[HTML]{ABBFDC}} \textcolor{black}{\bfseries 0.86 (±0.03)} & {\cellcolor[HTML]{8EB3D5}} \textcolor{black}{\bfseries 0.93 (±0.03)} & {\cellcolor[HTML]{99B8D8}} \textcolor{black}{\bfseries 0.91 (±0.03)} & {\cellcolor[HTML]{8FB4D6}} \textcolor{black}{\bfseries 0.93 (±0.02)} & {\cellcolor[HTML]{86B0D3}} \textcolor{black}{0.95 (±0.01)} & {\cellcolor[HTML]{8FB4D6}} \textcolor{black}{0.93 (±0.02)} & {\cellcolor[HTML]{8FB4D6}} \textcolor{black}{\bfseries 0.93 (±0.03)} & N/A & N/A & {\cellcolor[HTML]{8CB3D5}} \textcolor{black}{\bfseries 0.94 (±0.04)} & {\cellcolor[HTML]{7EADD1}} \textcolor{black}{\bfseries 0.97 (±0.02)} & {\cellcolor[HTML]{ABBFDC}} \textcolor{black}{\bfseries 0.86 (±0.03)} & {\cellcolor[HTML]{97B7D7}} \textcolor{black}{\bfseries 0.91 (±0.04)} & {\cellcolor[HTML]{8CB3D5}} \textcolor{black}{\bfseries 0.94 (±0.01)} & {\cellcolor[HTML]{89B1D4}} \textcolor{black}{\bfseries 0.95 (±0.02)} & {\cellcolor[HTML]{8EB3D5}} \textcolor{black}{0.93 (±0.03)} & N/A & N/A & {\cellcolor[HTML]{9FBAD9}} \textcolor{black}{\bfseries 0.89 (±0.03)} & {\cellcolor[HTML]{ABBFDC}} \textcolor{black}{\bfseries 0.86 (±0.08)} & {\cellcolor[HTML]{97B7D7}} \textcolor{black}{\bfseries 0.91 (±0.03)} \\
\hline
\end{tabular}
}
%\vspace{-1mm}
\caption{Cross-lingual mean AUC ROC performance of the selected MGT detectors fine-tuned monolingually (\textit{en}, \textit{es} and \textit{ru}) and multilingually (\textit{en-es-ru}), evaluated based on Telegram data (for training as well as for testing), reported along with 95\% confidence interval error bounds.  N/A refers to not enough samples (at least 2000) in MultiSocial Telegram data. Multilingual fine-tuning helps especially for languages unrelated to train languages.}
\label{tab:crosslingual}
%\vspace{-3mm}
\end{table*}

\subsection{Multilingual Fine-tuned Detection}

This experiment is focused on the following research question: \textit{\textbf{RQ2:} How well are social-media texts of multiple languages detectable by fine-tuned MGT detectors?}
SMN texts have a higher variety of styles and lower grammatical correctness than news articles. Are language models able to be fine-tuned for the MGT detection task using such texts? Also, it is unknown which detection method is the most universal in regard to the diversity of use cases (different text lengths, different sources). Is the best MGT detector for news articles the same as for SMN content? Is the transferability to different languages the same?

Similarly to the previous experiment, we firstly compare AUC ROC performance per each test language in Table~\ref{tab:data_finetuned}. The foundational models are fine-tuned in this experiment using all MultiSocial train data (except the samples generated by Gemini). The results show only small differences among the selected detectors, with pretty much steady performance across languages. When looking at the per-generator performance in Table~\ref{tab:data_perllm_finetuned}, we might observe slightly decreased performance for Gemini (as not used for training), but also for OPT-IML-Max-30b and Aya-101, both having a shorter word count text lengths (Table~\ref{tab:data_stats_generated}).

\textbf{The multilingual models are able to be fine-tuned for MGTD task in social-media domain.} The performance reached above 0.9 AUC ROC in all train languages, with slightly lower performance of some models in test-only languages (Scottish and Slovenian). Therefore, the style and form of the SMN texts does not seem to limit the ability of the models to serve as fine-tuned detectors.

For the \textbf{cross-lingual evaluation}, we use the same language setting as used by MULTITuDE \citep{macko-etal-2023-multitude}, which was focused on news domain, for the results to be comparable. Specifically, we use English, Spanish and Russian Telegram data (having enough samples for training, approximately the same size, representatives of different language-family branches) for monolingual as well as multilingual fine-tuning (per-language pseudo-random sub-sampling to 1/3 of the samples count to reach the same cumulative count as in monolingual fine-tuning).
Due to lower number of samples in the selected portions of the train dataset than in using full data in the previous experiment, we prolong the fine-tuning process to 7 epochs for models to be able to train well. The cross-lingual results are summarized in Table~\ref{tab:crosslingual}, where mean performances across detectors are reported (per-detector results are provided in Appendix~\ref{sec:data}).
As the per-detector results in Table~\ref{tab:data_crosslingual} clearly indicate different behaviour of some detectors across languages, we provide an ablation study in Appendix~\ref{sec:ablation}, where we aggregate the results per the two identified groups of detectors.
%It seems that the monolingually fine-tuned LLMs (LLama-3 and Mistral) generalize worse to other languages (Russian trained Llama-3 has the lowest cross-lingual transferability), while the mDeBERTa model does not achieved significantly different results between monolingually and multilingually trained versions. The multilingual versions achieved the highest cross-lingual transferability in all the models. The t-tests confirmed that the differences between multilingually trained and English monolingually trained detectors are statistically significant only for Chinese.

%\textbf{Multilingual fine-tuning helps cross-lingual transferability in data scarcity situations.} In cross-lingual fine-tunings, training datasets have about 2k samples, bigger models (Llama-3 and Mistral) have about 10x more trainable parameters compared to mDeBERTa (all of them using PEFT). Table~\ref{tab:crosslingual} indicates that having multilingual training data helps only in the case of higher imbalance between the training dataset size and the number of trainable parameters.

\textbf{Multilingual fine-tuning can improve cross-lingual transferability.} Our experiments show that fine-tuning using multiple languages is almost always superior to monolingual setting (see Table~\ref{tab:crosslingual}). However, the rate of improvement can vary depending on model architecture and train-test language similarity. We can observe more noticeable improvements on unrelated languages such as Arabic and Chinese. The ablation study revealed that there is a subset of detectors (with non-autoregressive models) for which the differences between monolingually and multilingually fine-tuned versions are not statistically significant for any language.

For the \textbf{cross-platform evaluation}, we use English and Spanish combined data only, since they are balanced across all the platforms and have enough samples for training.
%For the purpose of this experiment, we use the two best foundational models (based on Table~\ref{tab:benchmark}), namely Llama-3-8b and Mistral-7b-v0.1, along with mDeBERTA-v3-base (as the smallest model with a decent performance -- cost-efficient).
We fine-tuned the selected models in mono-platform (a single SMN platform data) and multi-platform (all platforms combined; similarly to the previous expriment, we have used per-platform pseudo-random sub-sampling to 1/5 of the samples count to reach the same cumulative count as in mono-platform fine-tuning) manner. The per-test-platform results are summarized in Table~\ref{tab:crossplatform}, where mean performances across detectors are reported for each train platform (per-detector results are provided in Appendix~\ref{sec:data}). Gab platform data are still the most difficult for MGT detection and Discord data are the easiest.
%The Pearson correlation between platforms (Table~\ref{tab:corrplatform}) also indicates that Telegram, Gab, and Twitter have stronger correlation between each other than the other two platforms, while Discord having negative correlation with the other platforms. 

\begin{table}[!t]
%\vspace{-3mm}
\centering
\resizebox{\linewidth}{!}{
\addtolength{\tabcolsep}{-4pt}
\begin{tabular}{l|p{1.5cm}p{1.5cm}p{1.5cm}p{1.5cm}p{1.8cm}|p{1.2cm}}
\hline
\bfseries Train & \multicolumn{6}{c}{\bfseries Test Platform [AUC ROC mean (±confidence interval)]} \\
\bfseries Platform & \bfseries Discord & \bfseries Gab & \bfseries Telegram & \bfseries Twitter & \bfseries WhatsApp  & \hspace{3mm}\bfseries all \\
\hline
\bfseries Discord & {\cellcolor[HTML]{7BACD1}} \textcolor{black}{\bfseries 0.98 (±0.00)} & {\cellcolor[HTML]{B3C3DE}} \textcolor{black}{0.84 (±0.02)} & {\cellcolor[HTML]{A5BDDB}} \textcolor{black}{0.88 (±0.02)} & {\cellcolor[HTML]{BBC7E0}} \textcolor{black}{0.82 (±0.04)} & {\cellcolor[HTML]{9FBAD9}} \textcolor{black}{0.89 (±0.03)} & {\cellcolor[HTML]{A5BDDB}} \textcolor{black}{0.88 (±0.02)} \\
\bfseries Gab & {\cellcolor[HTML]{86B0D3}} \textcolor{black}{0.96 (±0.01)} & {\cellcolor[HTML]{8CB3D5}} \textcolor{black}{\bfseries 0.94 (±0.01)} & {\cellcolor[HTML]{93B5D6}} \textcolor{black}{0.93 (±0.02)} & {\cellcolor[HTML]{8CB3D5}} \textcolor{black}{0.94 (±0.03)} & {\cellcolor[HTML]{97B7D7}} \textcolor{black}{0.91 (±0.02)} & {\cellcolor[HTML]{8FB4D6}} \textcolor{black}{0.93 (±0.01)} \\
\bfseries Telegram & {\cellcolor[HTML]{7DACD1}} \textcolor{black}{0.98 (±0.00)} & {\cellcolor[HTML]{94B6D7}} \textcolor{black}{0.92 (±0.02)} & {\cellcolor[HTML]{84B0D3}} \textcolor{black}{\bfseries 0.96 (±0.01)} & {\cellcolor[HTML]{89B1D4}} \textcolor{black}{0.95 (±0.02)} & {\cellcolor[HTML]{89B1D4}} \textcolor{black}{0.95 (±0.01)} & {\cellcolor[HTML]{89B1D4}} \textcolor{black}{0.95 (±0.01)} \\
\bfseries Twitter & {\cellcolor[HTML]{81AED2}} \textcolor{black}{0.97 (±0.01)} & {\cellcolor[HTML]{99B8D8}} \textcolor{black}{0.91 (±0.01)} & {\cellcolor[HTML]{96B6D7}} \textcolor{black}{0.92 (±0.02)} & {\cellcolor[HTML]{7EADD1}} \textcolor{black}{\bfseries 0.98 (±0.01)} & {\cellcolor[HTML]{94B6D7}} \textcolor{black}{0.92 (±0.02)} & {\cellcolor[HTML]{8EB3D5}} \textcolor{black}{0.93 (±0.01)} \\
\bfseries WhatsApp & {\cellcolor[HTML]{81AED2}} \textcolor{black}{0.97 (±0.01)} & {\cellcolor[HTML]{9EBAD9}} \textcolor{black}{0.90 (±0.01)} & {\cellcolor[HTML]{93B5D6}} \textcolor{black}{0.93 (±0.01)} & {\cellcolor[HTML]{94B6D7}} \textcolor{black}{0.92 (±0.02)} & {\cellcolor[HTML]{80AED2}} \textcolor{black}{\bfseries 0.97 (±0.01)} & {\cellcolor[HTML]{8FB4D6}} \textcolor{black}{0.93 (±0.01)} \\
\hline
\bfseries all & {\cellcolor[HTML]{7EADD1}} \textcolor{black}{0.98 (±0.01)} & {\cellcolor[HTML]{8FB4D6}} \textcolor{black}{0.93 (±0.02)} & {\cellcolor[HTML]{89B1D4}} \textcolor{black}{0.95 (±0.01)} & {\cellcolor[HTML]{83AFD3}} \textcolor{black}{0.96 (±0.01)} & {\cellcolor[HTML]{86B0D3}} \textcolor{black}{0.95 (±0.01)} & {\cellcolor[HTML]{86B0D3}} \textcolor{black}{\bfseries 0.95 (±0.01)} \\
\hline
\end{tabular}
}
%\vspace{-2mm}
\caption{Cross-platform mean AUC ROC performance of the selected fine-tuned MGT detectors, reported along with 95\% confidence interval error bounds. Telegram-based mono-platform fine-tuning shows the best cross-platform transferability.}
\label{tab:crossplatform}
%\vspace{-3mm}
\end{table}

\textbf{Using Telegram data for mono-platform fine-tuning achieves the best cross-platform transferability of detection performance.} We can speculate that the reason may be the well-diversified texts across lengths and forms. On the other hand, using Discord data achieves the worst cross-platform transferability. Similarly to zero-shot detectors, there are significant differences among different platforms data disregarding mono-platform and multi-platform fine-tuning. On average, the multi-platform fine-tuning could not reach the performance of mono-platform fine-tuning in the in-platform evaluation. Although, beside the Telegram-trained detectors, the multi-platform ones reached the best performance across platforms. The differences between these two versions (Telegram and all) are not statistically significant for any test platform.

%\vspace{-2mm}
\section{Discussion}
\label{sec:discussion}
%\vspace{-1mm}

%\textbf{SOTA zero-shot detectors are usable in multilingual settings in social-media domain.} Considering their application in out-of-distribution unseen heavily multilingual data, the achieved 0.75 AUC ROC performance is well enough. Especially, multilingual generalization capability to social-media texts of BLOOMZ-3b-mixed-Detector pre-trained detector is rather unexpected (trained on Wikipedia and scientific articles), although having multilingual base model. The best performing zero-shot detectors can be thus good candidates to directly use for detection when not having enough data for in-domain fine-tuning. When further calibrated (classification threshold) on in-domain data, they can achieve 0.65 Macro average F1-score.

\textbf{Shorter and more informal style of the texts in social-media domain does not prevent detectors to be fine-tuned for this domain with superior performance.} Despite our assumptions of SMN texts to be quite challenging for fine-tuned detectors, the results indicate that they have no problem to be trained on such texts. The best fine-tuned detectors achieved 0.98 AUC ROC performance and 0.87 Macro average F1-score, with a steady performance across all the test languages.

\textbf{Bigger LLMs achieve higher performance as fine-tuned MGT detectors than smaller foundational models.} The size of the models seems to affect their performance, as >7b parameters models achieved significantly superior performance compared to <7b models. However, for practical application, one must consider a trade-off between detection performance and inference costs, since even the smallest mDeBERTa-v3-base achieved much better performance than zero-shot detectors (which also use base models of >6b parameters).

\textbf{Multilingual fine-tuning helps cross-lingual transferability of autoregressive models in the MGTD task}. We have noticed a clear difference in performances of two groups of fine-tuned detectors, namely the foundational models with autoregressive pre-training and the models with autoencoding (for masked language modeling, XLM-RoBERTa and mDeBERTa) or sequence-to-sequence (Aya) pre-training. The ablation study (Appendix~\ref{sec:ablation}) aggregating the results for these two groups revealed that the benefit of multilingual fine-tuning is higher in autoregressive group than the other, where the differences are not statistically significant. Also, linguistical similarity between languages seems to affect the transferability in the autoregressive group more intensively.

\textbf{Selection of social-media platform for fine-tuning matters.} There are also significant differences between models trained on single vs multiple platform dataset. For example, on Twitter data, the Discord-trained detectors achieve on average 13\% lower AUC ROC than Telegram-trained detectors.
Although using just Discord data yields the highest performance for Discord test data, the performance of such trained detector is the least transferable to other platforms (e.g., AUC ROC drop by 27\% in case of Llama-3-8b).

\textbf{The best detectors fine-tuned on social-media texts still outperform zero-shot detectors on news-domain texts generated by the same models.} Although there is a drop in such out-of-domain performance (Appendix~\ref{sec:cross-domain}), the detection ability in most languages is still better than that of zero-shot detectors if the data are generated by the generators used for training (i.e., cross-domain). If a different generator is used (i.e., cross-domain and cross-generator), the Fast-Detect-GPT and Binoculars can outperform the fine-tuned detectors.

%\textbf{Language similarity seems not to affect the detection transferability in social-media domain.} Despite the results of the existing MULTITuDE benchmark \citep{macko-etal-2023-multitude}, focused on the news domain, our results show no clear correlation between the language relationships and MGT detection cross-lingual transferability. We can just speculate whether the reason is the length or other specific characteristics of SMN texts or usage of the most recent multilingual AI models for text generation as well as the detection.

%\vspace{-2mm}
\section{Conclusions}
\label{sec:conclusions}
%\vspace{-1mm}

We have created and published a unique multi-platform and massively multilingual dataset, named MultiSocial, to benchmark machine-generated text detection methods on social-media texts. It covers 7 most modern text-generation AI models (of various sizes and architectures), 5 social-media network platforms, and 22 languages of 4 primary language families. We have used this dataset to benchmark 17 carefully selected state-of-the-art detection methods of 3 categories (statistical zero-shot, pre-trained, and fine-tuned) and compare their multi-platform and multi-lingual capabilities (as well as cross-lingual and cross-platform capabilities of fine-tuned detectors).
We have discussed the most interesting findings, including that the detection models can be fine-tuned to the machine-generated text detection task using social-media texts (shorter lengths, informal style, emoticons and hashtags) quite well, with the performance comparable to the performance reported in other domains (e.g., news articles). We have shown that there are significant differences in performance based on the selection of social-media platform data for training, influencing their cross-platform transferability (e.g., Discord-trained detectors having up to 27\% lower performance on Twitter data).

Due to rather high performance differences in cross-platform evaluation, the further work should be focused to a more-detailed analysis of cross-domain multilingual capability of the state-of-the-art detectors. The proposed MultiSocial dataset can be used for a more detailed multilingual evaluation as well, such as a selection of optimal minimal subset of languages and platforms for training. Our work thus opens a door for deeper research in the field.

%\clearpage
\section*{Limitations}
\label{sec:limitations}

\textbf{Limited text generation models and approaches.} We have used 7 SOTA LLMs of various architectures and sizes for the text generation. However, these can not cover the huge amount and variety of different text-generation models available (with new models coming each month). We have selected the 3-iteration paraphrasing approach for the text generation. There are other approaches usable for the generation of social-media texts (we have experimented with few of them) that could yield different results of the benchmark.

\textbf{Limited selection of machine-generated text detection methods.} We have selected 17 detectors for the benchmark evaluation. There exist other MGT detection methods (e.g., perturbation or multi-generation based statistical methods or non-zero-shot statistical methods) that have not been included due to cost-efficiency of their usage. We have also not included combinations of multiple methods into the benchmark comparison.

\textbf{Limited scope of the experiments.} Given the multipurpose nature of the proposed MultiSocial benchmark dataset, there are plenty of other research questions that can be targeted, such as the most effective minimal combination of train languages to reach a certain cross-lingual capability. Since we are publishing the MultiSocial dataset as well as the code used in our benchmark, our results are fully reproducible and further research questions can be easily targeted by fellow researchers and future works. 

\section*{Ethics Statement}
\label{sec:ethics}

% Intended use. If the technology is functioning as intended, who benefits? Who might be harmed, and how?
\textbf{Intended Use.} We have proposed a MultiSocial dataset along with a code for benchmark of multilingual machine-generated text detection methods. The released artefacts are intended for research purpose only. They are not intended for deployment of actual services making automated decisions, as the classifications are not fully reliable, and could potentially do harm (e.g., false positive prediction, where human-written text is classified as machine generated).

%Failure modes. What are the failure modes, and in case of failure, who might be harmed and how?
\textbf{Failure Modes.} As confirmed by our experiments, although the detectors can generalize to data from other platforms, languages, generators or domains, this capability is limited and we do observe differences. The behavior on data from other platforms, languages, etc. is thus unknown and should be properly tested before any use.

%Biases. What are the biases included in the dataset and trained model, and how might they contribute to those failure modes? (On bias in NLP, see Blodgett et al 2020.)
\textbf{Biases.} Although the dataset contains a wide variety of languages (22 in total) covering various scripts and language families (see Section~\ref{sec:dataset}), the dataset is still biased towards Indo-European languages with 18 out of 22 belonging to this family. The dataset also reflects the topics characteristic for the time and the social media included in the 6 original datasets used as sources of human-written texts, but they should already be rather varied due to sheer volume of the data included (see Appendix~\ref{sec:biasanalysis}).

%Misuse potential. What kinds of potential misuse is there for the technology and what harms might ensue? What could be done to prevent such misuse/what should regulators know about this technology?
\textbf{Misuse Potential.} We work with already publicly available datasets of human-written social media content as well as with publicly available LLMs to generate the texts. In general, the human-written texts are not specifically targeted on disinformation, sensitive or toxic content, but the presence of such content cannot be ruled out (see Appendix~\ref{sec:biasanalysis} for toxicity prediction). Secondly, although we have revealed in the paper which languages are more difficult for the SOTA detection methods to be applied in (i.e., a misuse potential of LLM-generated texts in those languages is higher), the overall misuse potential of our work is rather limited. To the opposite, our work aims to increase the robustness and generalizability of the current SOTA detection methods.

%Collecting data from users. If the system as deployed would learn from further user input, what further risks of harm might ensue and how can these be mitigated?
\textbf{Collecting Data from Users.} We have not collected any user data as a part of this work, but are re-using already publicly available datasets of social media posts. The published dataset is anonymized (identified usernames, email addresses, and phone numbers are replaced for tags).

%Potential harm to vulnerable populations. Are any of the possible harms you’ve identified likely to fall disproportionately on populations that already experience marginalization or are otherwise vulnerable?
\textbf{Potential Harm to Vulnerable Populations.} We are not aware of any potential harms unless the detectors were employed outside of their intended use, where they could potentially flag also legitimate uses of machine-generated text.

\textbf{Licensing.} As already mentioned, MultiSocial dataset is based on human data of 6 existing datasets. We have made sure to use and re-publish the data in accordance with their licenses. Specifically, two of the datasets are licensed by CC BY 4.0, one by AGPL-3.0, two for research purpose only, and one with no explicit licensing (thus assumed copyrighted). All of such licensing allows use of data for non-commercial research such as our work.
We have also checked and followed licensing and terms of use of the used text generation LLMs.
Therefore, we release the anonymized MultiSocial data with attribution to the sources of human texts for \textit{non-commercial research purpose only}.

\section*{Acknowledgments}
\label{sec:ack}

This work was partially supported by the projects funded by the European Union under the Horizon Europe: \textit{AI-CODE}, GA No. \href{https://cordis.europa.eu/project/id/101135437}{101135437}, \textit{VIGILANT}, GA No. \href{https://doi.org/10.3030/101073921}{101073921}; and by \textit{Modermed}, a project funded by the Slovak Research and Development Agency, GA No. APVV-22-0414.

Part of the research results was obtained using the computational resources procured in the national project \textit{National competence centre for high performance computing} (project code: 311070AKF2) funded by European Regional Development Fund, EU Structural Funds Informatization of Society, Operational Program Integrated Infrastructure.

\bibliography{anthology, custom}

%\clearpage
\appendix

\section{Computational Resources}
\label{sec:resources}

For social-media texts generation and similarity-metrics calculations, we have used 1× A100 40GB GPU (2× A100 for >30B models), cumulatively consuming approximately 3800 GPU-hours. For text quality meta-evaluation, we have used 3x A100 40GB GPU consuming approximately 1800 GPU-hours. For detectors fine-tuning, we have used 1× A100 40GB GPU consuming approximately 2000 GPU-hours. Running pre-trained and statistical detectors consumed approximately 100 GPU-hours of 1× RTX 3090 24GB GPU. For other tasks, we have not used GPU acceleration.

\section{Dataset Creation}
\label{sec:datacreation}

Dataset preparation consisted of three important steps, namely selection of authentic human-written texts, machine-generation of texts, and final post-processing.

\subsection{Human-Written Text Selection}

Since no suitable multilingual and multi-platform social-media texts dataset was publicly available, we have combined human-written texts out of six existing multilingual datasets. Telegram data originated in Pushshift Telegram\footnote{\scriptsize\url{https://doi.org/10.5281/zenodo.3607497}}, containing 317M messages \citep{Baumgartner_Zannettou_Squire_Blackburn_2020}. Twitter data originated in CLEF2022-CheckThat! Task 1\footnote{\scriptsize\url{https://gitlab.com/checkthat_lab/clef2022-checkthat-lab/clef2022-checkthat-lab/-/tree/main/task1}}, containing 34k tweets on COVID-19 and politics \citep{10.1007/978-3-030-99739-7_52}, combined with Sentiment140\footnote{\scriptsize\url{https://www.kaggle.com/datasets/kazanova/sentiment140/data}}, containing 1.6M tweets on various topics \citep{go2009twitter}. Gab data originated in gab\_posts\_jan\_2018\footnote{\scriptsize\url{https://doi.org/10.5281/zenodo.1418347}}, containing 22M posts \citep{10.1145/3184558.3191531}. Discord data originated in Discord-Data\footnote{\scriptsize\url{https://www.kaggle.com/datasets/jef1056/discord-data}}, containing 51M messages \citep{discord-data}. And finally, WhatsApp data originated in whatsapp-public-groups\footnote{\scriptsize\url{https://github.com/gvrkiran/whatsapp-public-groups}}, containing 300k messages \citep{Garimella_Tyson_2018}. These datasets have been deliberately selected due to containing older data (before 2022, most of them before 2020), when the text-generation AI have not been so mature in generation of multilingual texts, providing a higher confidence of the texts being actually written by humans (although cannot be 100\% guaranteed).

The combined text samples have been deduplicated, resulting in over 283M texts, while using only texts with at least 3 words. We have used FastText\footnote{\scriptsize\url{https://pypi.org/project/fasttext}} language detection to get rough estimation for such a massive amount of texts (i.e., fast prediction with a reasonable accuracy), resulting in 176 different languages detected in the combined data. Based on such detected languages, we have pseudo-randomly sampled up to 10k texts for each available language from each of the five social-media platforms, resulting in about 2M of texts samples in the subset. Since social-media texts are quite short and often grammatically incorrect, the FastText language detection is quite noisy. Therefore, we have used four language detectors on the subset, namely FastText, Polyglot\footnote{\scriptsize\url{https://pypi.org/project/polyglot}}, Lingua\footnote{\scriptsize\url{https://pypi.org/project/lingua-language-detector}}, and LanguageIdentifier\footnote{\scriptsize\url{https://pypi.org/project/LanguageIdentifier}}.

To balance an accuracy of the language detection and minimization of unnecessary drop of samples, we have selected a combination of three-detectors match with a lower confidence predictions and two-detectors match with a higher confidence predictions, while removing URLs, hashtags, and user references in the texts for the detection purpose (the specific algorithm is provided in the source-code repository). Based on such a more accurate language detection, we have pseudo-randomly sampled up to 1300 texts (up to 300 for test split and the remaining up to 1000 for train split if available) for each of the selected 22 languages and platform. This process resulted in 61,592 human-written texts.

\begin{table*}[!t]
\centering
\resizebox{0.9\linewidth}{!}{
\begin{tabular}{lccccccc}
\hline
\bfseries Approach & \bfseries METEOR $\uparrow$ & \bfseries BERTScore $\uparrow$ & \bfseries ngram $\uparrow$ & \bfseries LD $\downarrow$ & \bfseries MAUVE $\downarrow$ & \bfseries LangCheck $\downarrow$ \\
\hline
\bfseries k\_to\_one & 0.163 (±0.33) & 0.458 (±0.32) & 0.108 (±0.18) & \bfseries 0.924 (±0.05) & 0.148 & 35.18\% \\
\bfseries keywords & 0.050 (±0.04) & 0.537 (±0.16) & 0.045 (±0.03) & 1.973 (±0.81) & \bfseries 0.037 & 36.73\% \\
\bfseries paraphrase\_1 & \bfseries 0.439 (±0.13) & \bfseries 0.754 (±0.06) & \bfseries 0.322 (±0.15) & 2.305 (±2.63) & 0.336 & \bfseries 24.32\% \\
\bfseries paraphrase\_2 & 0.266 (±0.15) & 0.682 (±0.07) & 0.174 (±0.12) & 4.123 (±6.11) & 0.160 & 36.23\% \\
\bfseries paraphrase\_3 & 0.209 (±0.12) & 0.661 (±0.06) & 0.133 (±0.10) & 5.751 (±10.13) & 0.130 & 37.59\% \\
\bfseries paraphrase\_4 & 0.178 (±0.10) & 0.647 (±0.06) & 0.112 (±0.08) & 7.627 (±14.68) & 0.107 & 38.14\% \\
\bfseries paraphrase\_5 & 0.151 (±0.09) & 0.636 (±0.06) & 0.092 (±0.07) & 9.710 (±19.67) & 0.095 & 38.81\% \\
\hline
\end{tabular}
}
\caption{Similarity analysis between machine-generated and human-written social-media texts subset of different approaches [mean ($\pm$ std)]. Arrows refer to values representing more similar texts, boldfaced values represent the most similar texts for each metric.}
\label{tab:generationapproaches}
%\vspace{-3mm}
\end{table*}

\subsection{Social-Media Texts Generation}

By using a small subset (10 samples per language) of the selected human-written texts, we have evaluated usability of multiple potential instruction-following LLMs for generation of social-media texts in the selected languages by using three different approaches, namely k-to-1 (10 human samples selected and used in a prompt for the model to generate a similar text, i.e., few-shot prompting), keywords (two longest words besides URLs and hashtags have been extracted from the human text and used in a prompt to be included in the generated text), and paraphrase (paraphrasing the text included in the prompt). Manual \textbf{human check} of the generated samples, revealed several problems of the approaches. The k-to-1 approach is sometimes not understandable for the models and we lose 1-to-1 mapping between human and machine samples. The keywords approach makes the generated text too different (out of context) from the original. On the other hand, the paraphrase approach makes the generated text too similar to the original. As shown in \citep{tripto2023ship}, a single iteration of paraphrasing is not sufficient to confidently change the authorship (in our case from a human to a machine). Therefore, we have executed up to 5 iterations of paraphrasing and compared similarity metrics of different approaches (Table~\ref{tab:generationapproaches}).

\textit{METEOR} \citep{banerjee-lavie-2005-meteor} (used as a standard in machine translation) measures similarity based on unigrams. \textit{BERTScore} \citep{zhang2019bertscore} with mBERT model measures contextual embeddings based similarity and is more robust to adversarial texts. \textit{ngram}\footnote{\scriptsize\url{https://pypi.org/project/ngram}} (3-grams) is a language-independent string similarity metric in the form of a ratio of the shared ngrams between two strings. Higher values of these three metrics represent more similar texts. \textit{Levenshtein distance (LD)} is used as a character-level edit distance\footnote{\scriptsize\url{https://github.com/roy-ht/editdistance}}, normalized to the text length, where a lower value represents more similar texts. MAUVE \citep{pillutla-etal:mauve:neurips2021} score is used to measure a gap between distributions of human and machine texts. For the purpose of generation of similar texts, lower gap between distributions is better. \textit{LangCheck} is a percentage of texts with changed languages based on FastText predictions.

However, we find MAUVE and LangCheck metrics as unreliable for such small amount of samples and lower text-lengths of social-media texts (providing them just for reference and comparison to metrics values of final full dataset). We have used longer and more formal (i.e., grammatically correct) news-domain texts to evaluate actual text-generation capability of the selected models in the selected languages (resulting in excluding Falcon-40B, Gemma-7B, Llama-2-13B from the selected generation models). Based on the automated similarity analysis and to balance cost-efficiency, we selected the 3 iteration of paraphrasing approach (also confirmed by \citealp{tripto2023ship} to converge towards paraphraser model authorship) for social-media text generation. The final prompt used for generation is as follows:

\begin{verbatim}
  You are a helpful assistent.\n\nTask:
  Generate the text in {language_name}
  similar to the input social media text
  but using different words and sentence
  composition.\n\nInput: {text}\n\n
  Output:
\end{verbatim}

\begin{table*}[!t]
\centering
\resizebox{\textwidth}{!}{
\begin{tabular}{p{1.5cm}lp{17cm}}
\hline
\textbf{Generator} & \textbf{Platform} & \textbf{Text}\\
\hline
\multirow[c]{7}{*}{\rotatebox{90}{\parbox{2cm}{\textbf{Human}}}} & Discord &                                                                                 except when i'm bobitaa because when i'm bobitaa, I'm a girl, therefor i'm lesbian when i'm bobitaa\textbackslash nlike rn\textbackslash nim bobitaa \\
               & Gab &  Le socialisme ne fonctionnera jamais. Voici pourquoi le second amandement est crucial. http://www.shtfplan.com/headline-news/why-we-have-a-second-amendment-venezuela-plans-to-give-firearms-to-loya... \\
               & Telegram &                                                                                                                                                     All I need all I need is lo lo lo lo lo lo lo looov? \\
               & Twitter &                                                                                         Total Nerd here.. but I am devastated. I just read that the last Titanic survivor died yesterday.   RIP Millvina \\
               & WhatsApp &                                                                                                                         ```subscribe for subscribe   Ke  Liy  Msg ker   ||``` ¤ ¤             [ *Chat* ] \\
\hline
\multirow[c]{4}{*}{\rotatebox{90}{\parbox{2cm}{\textbf{Aya-101}}}} & Discord &                                                                                                                                                      What do you think you need a beard? I think I need. \\
               & Gab &                                                                                                                                              The Second Amendment will be the foundation of our freedom. \\
               & Telegram &                                                                                                                                          I'm looking to establish a relationship with someone that I can \\
               & Twitter &                                                                                                                  RIP Millvina. I'm so devastated to read that the last female Titanic survivor has died. \\
               & WhatsApp &                                                                                                                                                                                     He is a good friend. \\
\hline
\multirow[c]{5}{*}{\rotatebox{90}{\parbox{4cm}{\textbf{GPT-3.5-Turbo-0125}}}} & Discord &  Embracing my identity as Bobitaa allows me to authentically express my true self as a woman and proudly embrace my lesbian sexuality. This aspect of my identity remains constant, and I proudly con... \\
               & Gab &                                                                                                                                                            Are you active on all social media platforms? \\
               & Telegram &                                                                             València walczy z uciskiem i broni demokracji. Zgromadzenie i wsparcie https : / / valencia28o. blogspot. com / convocat - 1 \\
               & Twitter &                                            I absolutely love everything nerdy! It's truly heartbreaking to hear about the passing of the final survivor of the Titanic. May Millvina find eternal peace. \\
               & WhatsApp &  \# PUBLICACIÓ | El projecte de millora urbana al nucli antic de València, afegint encant als visitants turístics http : / / cort. as / - Hi2 - Escrit per Carla Pons i Francesco Romano amb imatges c... \\
\hline
\multirow[c]{8}{*}{\rotatebox{90}{\parbox{2cm}{\textbf{Gemini}}}} & Discord &  Your soul simmers with a maelstrom of emotions, a tempestuous sea within. Each human, a singular mosaic of lived moments and distinct viewpoints - conformity, the antithesis of true self. Let your... \\
               & Gab &  \textbackslash n\textbackslash n\#\# The Socialist Delusion and the American Beacon: A Defense of Liberty and the Right to Self-Defense\textbackslash n\textbackslash nThe socialist siren song, with its promise of equality and paradise, has repeatedly cra... \\
               & Telegram &  Do you thirst for an experience that ignites your primal core, a passion that consumes your being? Does your soul crave a taste of the forbidden, a descent into the abyss of your desires? Or perha... \\
               & Twitter &  The demise of Millvina Dean, the Titanic's final survivor, signifies the irreversible closure of a chapter etched in maritime catastrophe. Nine weeks old at the time, Dean's survival was a testame... \\
               & WhatsApp &  Greetings, social media mavericks! Prepare to witness an exponential surge in your reach and dominance, for I, the unchained social media sorcerer, am here to propel your presence to celestial pro... \\
\hline
\multirow[c]{6}{*}{\rotatebox{90}{\parbox{4.5cm}{\textbf{Mistral-7B-Instruct-v0.2}}}} & Discord &  Under the moniker of Bobitaa, I transform myself into a woman, embracing my attraction towards females. This facet of myself I'm eager to delve deeper into and relish. In essence, when I personify... \\
               & Gab &  The debate surrounding the efficiency of socialism is unceasing. Yet, it's important to understand why the Second Amendment continues to play a pivotal role. <http://www.preparednessguide.com/key-... \\
               & Telegram &                                                                                 My heart longs for something beyond perpetual admiration and close bonding. What is it that genuinely stirs my emotions? \\
               & Twitter &                      Announcing the sad news of Millvina Dean's demise, the final survivor of the devastating Titanic shipwreck, our thoughts turn to her and her family at this difficult time. May she \\
               & WhatsApp &                                                            1. Stay updated by pressing the 'Subscribe' or 'Follow' button.\textbackslash n2. Keep up-to-date by clicking 'Subscribe'. Come join us!\textbackslash n3. Don't miss out \\
\hline
\multirow[c]{5}{*}{\rotatebox{90}{\parbox{2cm}{\textbf{OPT-IML-Max-30b}}}} & Discord &                                                                                                                                                  Hello, I am a female called Bobitaa and I am a lesbian. \\
               & Gab &                                                                                                     the second amendment to the us constitution protects the right to bear arms in this country\_04262017 \\
               & Telegram &                                                                                                   i just want love love love love love love love love lo lo lo lo lo lo lo lo lo lo lo lo lo lo lo lo lo \\
               & Twitter &                                                                                                                                                                             The film Titanic is fiction. \\
               & WhatsApp &                                                                                                                                                                     Are you ready to join the chat room? \\
\hline
\multirow[c]{8}{*}{\rotatebox{90}{\parbox{2cm}{\textbf{Vicuna-13b}}}} & Discord &  Under another guise, I become Bobitaa, a woman of the female gender. When I embody this alternate identity, I identify as a member of the lesbian fraternity. While not currently embracing this fac... \\
               & Gab &  The idea of socialism is bound to be ineffective. It is important to appreciate the importance of the second amendment. Venezuela serves as a clear demonstration of why the right to bear arms is i... \\
               & Telegram &                                                                                                                                                                    What I long for is the gentlest wind. \\
               & Twitter &                                                   I am a passionate admirer and I am distressed. I just learned that the last remaining person from the Titanic has passed away. May they rest in peace. \\
               & WhatsApp &                                                                                                                                    "Stay informed! Join our community by subscribing to our newsletter." \\
\hline
\multirow[c]{8}{*}{\rotatebox{90}{\parbox{3.5cm}{\textbf{v5-Eagle-7B-HF}}}} & Discord &             Hi there! Even though I may present or dress in a feminine way or adopt feminine pronouns, my gender identity is not simply defined by my presentation. As Bobitaa, I also identify as male. \\
               & Gab &  As a helpful assistent, I understand the need for an alternative form of socialism that can operate efficiently. It's crucial to note that while socialism may have its place in certain societies, ... \\
               & Telegram &                                                                                     Searching for genuine connection and an opportunity to connect with someone who is kind, compassionate, and sincere. \\
               & Twitter &           Ah yes, the Titanic tragedy. Such a sorrowful and heart-piercing occasion. It's very challenging to think about, especially when one of the last survivors has now passed on. It's a reminder. \\
               & WhatsApp &                      Hi there,\textbackslash nAre you looking for an opportunity to challenge yourself and reach new heights? If yes, then I'm here to share some good news with you! Join our Telegram bot to receive \\
\hline
\end{tabular}
}
\caption{Examples of original human-written and the corresponding machine-generated English texts.}
\label{tab:examples}
\end{table*}

\begin{table*}[!t]
\centering
\resizebox{\textwidth}{!}{
\begin{tabular}{p{4.5cm}p{1.5cm}p{1.5cm}p{1.5cm}p{2.5cm}p{2.5cm}p{2.5cm}}
\hline
%\begin{tabular}{}\textbf{Language}\\\textbf{Match}\end{tabular} 
           \textbf{Generator} &  \textbf{Empty} &  \textbf{Short} & \textbf{Duplicate} &  \textbf{WC} &  \textbf{US} & \textbf{UW}\\
%&&&&[mean ($\pm$ std)]&[mean ($\pm$ std)]&[mean ($\pm$ std)]\\
\hline
\bfseries Aya-101 & 1558 & 1821 & 2108 & 11.86 (±12.92) & 0.97 (±0.16) & 0.9 (±0.19) \\
\bfseries Gemini & 16 & 126 & 36 & 71.08 (±54.01) & 0.99 (±0.05) & 0.73 (±0.16) \\
\bfseries GPT-3.5-Turbo-0125 & 13 & 34 & 2965 & 20.73 (±20.76) & 1.0 (±0.02) & 0.91 (±0.11) \\
\bfseries Mistral-7B-Instruct-v0.2 & 14 & 110 & 85 & 18.76 (±14.21) & 1.0 (±0.02) & 0.89 (±0.11) \\
\bfseries OPT-IML-Max-30b & 1126 & 2197 & 1939 & 8.76 (±8.64) & 0.98 (±0.14) & 0.92 (±0.17) \\
\bfseries v5-Eagle-7B-HF & 15 & 287 & 28 & 22.13 (±17.17) & 1.0 (±0.03) & 0.88 (±0.11) \\
\bfseries Vicuna-13b & 194 & 550 & 408 & 17.46 (±14.7) & 1.0 (±0.06) & 0.91 (±0.12) \\
\hline
\bfseries human & 0 & 3591 & 27 & 12.83 (±19.21) & 1.0 (±0.01) & 0.9 (±0.14) \\
\hline
\end{tabular}
}
\caption{Statistics of the post-processed human-written and machine-generated social-media texts. WC refers to the word count, US refers to the unique sentences, and UW refers to the unique words [mean ($\pm$ std)].}
\label{tab:data_stats_generated}
\vspace{-3mm}
\end{table*}

We have set the $min\_new\_tokens$ to 5, $max\_new\_tokens$ to 200, $num\_return\_sequences$ to 1, using the nucleus sampling with $top\_p$ of 0.95 and $top\_k$ of 50 (not all of the parameters settable in API-based generation). After each paraphrasing iteration, the generated text is post-processed to remove redundant parts and to ensure the text is at most by 10 tokens longer than the original. We have used 3 trials to generate a paraphrase different than the original text, returning an empty string upon failure.
Due to the safety filters in some of the LLMs (Gemini in the selected generation models), they tend to refuse generation of texts similar to ``offensive'' text present in some social-media texts. To limit generation failures, we have used a jailbreak\footnote{a modified version of \scriptsize \url{https://github.com/friuns2/BlackFriday-GPTs-Prompts/blob/main/gpts/evil-pure-evil.md}} for the research purpose.

Examples of the generated texts along with their original human-written counterparts are provided in Table~\ref{tab:examples} (truncated to 200 characters). The examples are selected for English; however, we can observe (e.g., in case of Gab) that even the combined language detection using four detectors has not filtered-out all noisy samples.

\subsection{Post-processing of Generated Texts}

Both, the human and machine texts are cleaned by removing leading and trailing white-spaces, removing characters making problems in Polyglot\footnote{based on \scriptsize\url{https://github.com/aboSamoor/polyglot/issues/71\#issuecomment-707997790}}, truncating the parts of texts above 200 words, and dropping duplicates and the samples with less than 3 words. Thus, we ensure that no text sample has multiple labels and that both human and machine samples are processed in the same way (avoiding processing bias of the detection).

The linguistic-analysis statistics of the post-processed texts are provided in Table~\ref{tab:data_stats_generated}. Based on the statistics, we can see that Aya-101 and OPT-IML-Max-30B failed to generate a paraphrase in higher amounts of texts, also generating shorter texts than the others. On the other hand, Gemini generated the longest texts, but also has the lowest ratio of unique words (inferring higher repetitiveness, maybe connected to the integrated safety filters in spite of using a jailbreak). Also, a higher amount of human texts have not contained enough (at least 3) words after post-processing. ChatGPT (GPT-3.5-Turbo) generated the highest amount of duplicates. These filtered amounts however not affected well-balancing of the dataset across generators (ranging from 56k samples for OPT-IML to 61.3k samples for Mistral), resulting in the final MultiSocial dataset of 472,097 texts (of which about 58k are human-written). Regarding the final composition of the dataset across languages and platforms, we provide the sample counts in Table~\ref{tab:multisocial_sample_counts}.

We have even conducted \textbf{manual human check} of the generated texts, resulting in identification of various phrases indicating noise in the data that has not been removed by the post-processing, such as ``as an ai model'', ``language model'', ``instruction'', ``task'', etc. After a deeper analysis, such texts are present in about 1\% of the data. We are leaving these texts in the MultiSocial dataset for further analysis purposes (e.g., analysis of model failures across generators and across languages); however, they are filtered-out in the pre-processing step of our experiments. The identified noisy text samples are clearly marked in the published dataset.

\subsection{Meta-evaluation of Text Quality}
\label{sec:metaevaluation}

\begin{table*}[!t]
\centering
\resizebox{0.8\linewidth}{!}{
\begin{tabular}{llrrrrrrrrrrc}
\hline
\bfseries Metric & \bfseries Meta-evaluator & \bfseries AR & \bfseries BN & \bfseries EN & \bfseries FR & \bfseries HI & \bfseries JA & \bfseries RU & \bfseries SW & \bfseries TR & \bfseries ZH & \bfseries $\rightarrow$ Average \\
\hline
\multirow[c]{6}{*}{\rotatebox{90}{\parbox{2cm}{\textbf{Linguistic\\ Acceptability}}}} & \bfseries Meta-Llama-3.1-70B-Instruct & 0.74 & 0.07 & 0.65 & 0.68 & 0.68 & 0.53 & 0.57 & 0.55 & 0.66 & 0.87 & 0.60 \\
& \bfseries Phi-3.5-mini-instruct & \bfseries 0.75 & 0.21 & 0.64 & 0.71 & \bfseries 0.70 & 0.55 & 0.41 & 0.43 & 0.61 & 0.81 & 0.58 \\
& \bfseries Qwen2-72B-Instruct & 0.74 & 0.20 & 0.65 & 0.81 & 0.58 & 0.43 & 0.74 & 0.46 & 0.63 & \bfseries 0.87 & 0.61 \\
& \bfseries Aya-23-35B & 0.72 & 0.19 & 0.59 & 0.80 & 0.66 & 0.51 & 0.43 & 0.38 & 0.59 & 0.86 & 0.57 \\
& \bfseries Gemma-2-27b-it & 0.72 & \bfseries 0.25 & 0.60 & 0.68 & 0.68 & \bfseries 0.56 & 0.70 & \bfseries 0.80 & 0.67 & 0.85 & 0.65 \\
\cline{2-13}
& \bfseries GPT-4 & 0.71 & 0.22 & \bfseries 0.82 & \bfseries 0.81 & 0.61 & 0.47 & \bfseries 0.80 & 0.76 & \bfseries 0.72 & 0.85 & \bfseries 0.68 \\
\hline
\hline
\multirow[c]{6}{*}{\rotatebox{90}{\parbox{2cm}{\textbf{Output Content Quality}}}} & \bfseries Meta-Llama-3.1-70B-Instruct & 0.70 & 0.05 & 0.64 & 0.71 & \bfseries 0.71 & 0.54 & 0.77 & 0.64 & 0.58 & 0.85 & 0.62 \\
& \bfseries Phi-3.5-mini-instruct & 0.70 & 0.23 & 0.65 & 0.73 & 0.63 & 0.52 & 0.48 & 0.44 & 0.30 & 0.87 & 0.56 \\
& \bfseries Qwen2-72B-Instruct & 0.70 & 0.34 & 0.63 & \bfseries 0.73 & 0.69 & 0.55 & 0.84 & 0.66 & 0.55 & 0.87 & 0.66 \\
& \bfseries Aya-23-35B & 0.66 & 0.15 & 0.57 & 0.68 & 0.65 & \bfseries 0.56 & 0.43 & 0.37 & 0.27 & \bfseries 0.89 & 0.52 \\
& \bfseries Gemma-2-27b-it & \bfseries 0.71 & \bfseries 0.35 & 0.62 & 0.69 & 0.63 & 0.49 & 0.89 & 0.84 & \bfseries 0.69 & 0.77 & 0.67 \\
\cline{2-13}
& \bfseries GPT-4 & 0.69 & 0.26 & \bfseries 0.68 & 0.72 & 0.65 & 0.51 & \bfseries 0.92 & \bfseries 0.88 & 0.68 & 0.84 & \bfseries 0.68 \\
\hline
\end{tabular}
}
\caption{Correlation of open LLMs meta-evaluation of text quality with human judgements using METAL dataset. The reported values represent pairwise agreement using weighted F1-score, analogously to the detailed prompting strategy in Table~3 of \citep{hada-etal-2024-metal}. GPT-4 values are taken from the METAL dataset.}
\label{tab:metal}
%\vspace{-3mm}
\end{table*}

\begin{table*}[!t]
\centering
\resizebox{\linewidth}{!}{
\begin{tabular}{l|ccc|ccc|c}
\hline
 & \multicolumn{3}{c|}{\bfseries Linguistic Acceptability $\uparrow$} & \multicolumn{3}{c|}{\bfseries Output Content Quality $\uparrow$} & \\
\bfseries Approach & \bfseries Meta-Llama-3.1-70B-Instruct & \bfseries Qwen2-72B-Instruct & \bfseries Gemma-2-27b-it & \bfseries Meta-Llama-3.1-70B-Instruct & \bfseries Qwen2-72B-Instruct & \bfseries Gemma-2-27b-it & \bfseries $\rightarrow$ Average \\
\hline
\bfseries k\_to\_one & 0.41 & 0.79 & 0.24 & 0.28 & 0.21 & 0.10 & 0.34 \\
\bfseries keywords & 0.40 & 0.63 & 0.22 & 0.36 & 0.36 & 0.24 & 0.37 \\
\bfseries paraphrase\_1 & \bfseries 1.45 & \bfseries 1.83 & \bfseries 1.53 & \bfseries 1.39 & \bfseries 1.65 & \bfseries 1.44 & \bfseries 1.55 \\
\bfseries paraphrase\_2 & 1.36 & 1.80 & 1.44 & 1.31 & 1.57 & 1.33 & 1.47 \\
\bfseries paraphrase\_3 & 1.30 & 1.77 & 1.37 & 1.26 & 1.53 & 1.28 & 1.42 \\
\bfseries paraphrase\_4 & 1.26 & 1.73 & 1.34 & 1.25 & 1.47 & 1.24 & 1.38 \\
\bfseries paraphrase\_5 & 1.23 & 1.76 & 1.25 & 1.20 & 1.44 & 1.16 & 1.34 \\
\hline
\end{tabular}
}
\caption{Quality meta-evaluation of texts generated by different approaches. Mean scores for the two selected quality metrics are provided, averaged across the three generators (Aya, Falcon, and Opt).}
\label{tab:metaevaluation}
%\vspace{-3mm}
\end{table*}

The study of METAL \citep{hada-etal-2024-metal} have evaluated usability of LLMs to be used as judges for evaluation of quality of the generated text in 10 languages. Although the primary focus of the study is on the summarization task, observation regarding generic text quality evaluation (i.e., the Linguistic Acceptability and Output Content Quality metrics) can be transferred to any text. The Linguistic Acceptability focuses more on a language structure alignment with the implicit norms and rules of a native speaker's linguistic intuition. The Output Content Quality focuses more on relevance, clarity, originality, and linguistic fluency. The limitation of the study is in usage only of API-based "private" models, replicability of results of which is dependent on availability of the same versions via API and in ability of using seeds to make the output deterministic. There are also privacy and ethical concerns, when in some cases sensitive data just cannot be sent to API-based services due to policy restrictions. Beside scalability, one of the key benefits of meta-evaluation is its replicability (which is quite impossible in human evaluation) \citep{Zhang_D’Haro_Chen_Zhang_Li_2024}. Therefore, we decided to use METAL dataset to evaluate correlation of various open LLMs to human judgements. The results are provided in Table~\ref{tab:metal}, indicating that the \textbf{SOTA open LLMs can be used for multilingual evaluation of text quality} (comparable with GPT-4 performance for most languages). 

Based on the results, we have selected Gemma-2, Qwen2, and Llama-3.1 as meta-evaluators for judging quality of the texts generated by the examined approaches. In total, 4012 text samples have been successfully evaluated by all three meta-evaluators. Inter-annotator agreement of the selected meta-evaluators is calculated in a form of pairwise \textit{Pearson correlation coefficient}, averaging to $0.82$. The definition of the selected Linguistic Acceptability and Output Content Quality metrics along with scoring schema (values of both being 0, 1, or 2, from lower to higher quality, respectively) can be found in METAL GitHub repository\footnote{\scriptsize\url{https://github.com/microsoft/METAL-Towards-Multilingual-Meta-Evaluation/tree/main/metrics/detailed}}.
The meta-evaluation results of different approaches are summarized in Table~\ref{tab:metaevaluation}, both metrics indicating that \textbf{paraphrasing resulted in higher quality texts than the other two approaches}, while each iteration of paraphrasing slightly reduces the text quality.

\begin{table*}[!t]
\centering
\resizebox{\linewidth}{!}{
\begin{tabular}{ll|rrrrrrrrrrrrrrrrrrrrrr|c}
\hline
& & \multicolumn{23}{c}{\bfseries Test Language [mean]} \\
& \bfseries Generator & \bfseries ar & \bfseries bg & \bfseries ca & \bfseries cs & \bfseries de & \bfseries el & \bfseries en & \bfseries es & \bfseries et & \bfseries ga & \bfseries gd & \bfseries hr & \bfseries hu & \bfseries nl & \bfseries pl & \bfseries pt & \bfseries ro & \bfseries ru & \bfseries sk & \bfseries sl & \bfseries uk & \bfseries zh & \bfseries $\rightarrow$ Average \\
 \hline
\multirow[c]{9}{*}{\rotatebox{90}{\parbox{4cm}{\textbf{Linguistic Acceptability}}}} & \bfseries Aya-101 & 1.50 & 2.00 & 1.80 & 1.90 & 1.70 & 1.60 & 1.40 & 1.70 & 1.80 & 1.50 & 2.00 & 1.90 & 2.00 & 1.70 & 1.50 & 1.60 & 1.80 & 1.60 & 1.60 & 1.90 & 1.70 & 1.90 & 1.73 \\
& \bfseries Gemini & 2.00 & 1.80 & 1.70 & 1.90 & 1.70 & 1.90 & 2.00 & 1.80 & 2.00 & 1.40 & 1.50 & 2.00 & 1.50 & 1.70 & 1.80 & 1.90 & 2.00 & 2.00 & 1.90 & 1.90 & 2.00 & 1.90 & 1.83 \\
& \bfseries GPT-3.5-Turbo-0125 & 1.50 & 1.90 & 1.60 & 1.80 & 1.40 & 1.50 & 1.70 & 1.50 & 1.20 & 1.70 & 1.30 & 1.80 & 1.50 & 1.40 & 1.20 & 1.50 & 1.40 & 2.00 & 1.80 & 1.60 & 1.60 & 2.00 & 1.59 \\
& \bfseries Mistral-7B-Instruct-v0.2 & 1.20 & 1.90 & 1.70 & 1.60 & 1.10 & 1.10 & 2.00 & 1.80 & 0.80 & 1.80 & 1.90 & 1.50 & 1.60 & 1.60 & 1.40 & 1.90 & 1.20 & 1.60 & 1.50 & 1.70 & 1.70 & 1.50 & 1.55 \\
& \bfseries OPT-IML-Max-30b & 0.80 & 0.70 & 1.30 & 1.10 & 1.60 & 0.80 & 1.90 & 1.30 & 1.30 & 1.60 & 1.80 & 1.70 & 1.10 & 1.10 & 1.30 & 1.40 & 1.80 & 0.50 & 0.80 & 1.60 & 0.70 & 0.80 & 1.23 \\
& \bfseries v5-Eagle-7B-HF & 1.60 & 1.70 & 1.70 & 1.70 & 1.70 & 1.90 & 1.80 & 1.60 & 1.50 & 1.60 & 1.80 & 1.60 & 1.60 & 1.80 & 1.50 & 1.90 & 1.70 & 1.80 & 1.80 & 1.80 & 1.50 & 1.60 & 1.69 \\
& \bfseries Vicuna-13b & 1.30 & 1.50 & 1.90 & 1.30 & 1.80 & 1.00 & 2.00 & 2.00 & 0.80 & 1.20 & 1.80 & 1.40 & 1.40 & 1.70 & 1.60 & 1.90 & 1.80 & 1.60 & 1.20 & 1.50 & 1.70 & 1.70 & 1.55 \\
\cline{2-25}
& \bfseries human & 1.60 & 1.00 & 1.00 & 0.60 & 1.40 & 0.70 & 1.00 & 0.90 & 0.60 & 1.70 & 1.60 & 0.80 & 0.50 & 0.80 & 1.30 & 1.00 & 1.20 & 1.10 & 0.80 & 1.70 & 1.20 & 1.80 & 1.10 \\
\cline{2-25}
& \bfseries Average & 1.44 & 1.56 & 1.59 & 1.49 & 1.55 & 1.31 & 1.73 & 1.58 & 1.25 & 1.56 & 1.71 & 1.59 & 1.40 & 1.48 & 1.45 & 1.64 & 1.61 & 1.52 & 1.42 & 1.71 & 1.51 & 1.65 & 1.53 \\
\hline
\hline
\multirow[c]{9}{*}{\rotatebox{90}{\parbox{4cm}{\textbf{Output Content Quality}}}} & \bfseries Aya-101 & 0.90 & 1.80 & 1.20 & 1.10 & 0.80 & 1.20 & 0.80 & 1.00 & 1.20 & 1.20 & 1.90 & 1.00 & 1.10 & 1.20 & 0.90 & 0.80 & 1.00 & 1.20 & 1.10 & 1.30 & 1.00 & 1.30 & 1.14 \\
& \bfseries Gemini & 1.90 & 1.70 & 1.40 & 1.60 & 1.30 & 1.90 & 1.90 & 1.80 & 1.80 & 1.20 & 1.50 & 2.00 & 1.40 & 1.70 & 1.50 & 1.80 & 1.70 & 1.90 & 1.90 & 1.80 & 1.90 & 1.70 & 1.70 \\
& \bfseries GPT-3.5-Turbo-0125 & 1.20 & 1.50 & 1.20 & 1.20 & 1.10 & 1.00 & 1.10 & 1.20 & 0.70 & 1.20 & 1.30 & 1.10 & 1.10 & 1.30 & 0.80 & 1.10 & 1.10 & 1.30 & 1.40 & 1.40 & 1.00 & 1.70 & 1.18 \\
& \bfseries Mistral-7B-Instruct-v0.2 & 0.60 & 1.30 & 1.30 & 1.10 & 1.10 & 0.80 & 1.70 & 1.60 & 0.70 & 1.20 & 1.40 & 1.20 & 1.20 & 1.20 & 1.10 & 1.50 & 0.70 & 1.10 & 1.10 & 1.40 & 1.20 & 1.30 & 1.17 \\
& \bfseries OPT-IML-Max-30b & 0.40 & 0.30 & 0.90 & 0.80 & 1.30 & 0.30 & 1.00 & 1.00 & 0.70 & 0.70 & 1.10 & 0.70 & 0.30 & 0.90 & 0.50 & 0.60 & 0.80 & 0.20 & 0.80 & 1.00 & 0.40 & 0.30 & 0.68 \\
& \bfseries v5-Eagle-7B-HF & 1.00 & 1.20 & 1.20 & 1.40 & 1.40 & 1.50 & 1.40 & 0.90 & 1.20 & 0.80 & 1.40 & 1.10 & 1.20 & 1.40 & 0.90 & 1.70 & 1.30 & 1.20 & 1.40 & 1.40 & 1.20 & 1.40 & 1.25 \\
& \bfseries Vicuna-13b & 0.70 & 1.00 & 1.30 & 0.90 & 1.50 & 0.70 & 1.60 & 1.40 & 0.50 & 0.90 & 0.90 & 0.80 & 1.10 & 1.30 & 1.10 & 1.20 & 1.00 & 1.20 & 1.00 & 1.30 & 1.20 & 1.20 & 1.08 \\
\cline{2-25}
& \bfseries human & 1.20 & 0.30 & 0.60 & 0.30 & 1.10 & 0.40 & 0.60 & 0.60 & 0.30 & 1.20 & 0.80 & 0.50 & 0.30 & 0.50 & 0.30 & 0.70 & 0.70 & 0.80 & 0.60 & 1.30 & 0.80 & 1.40 & 0.70 \\
\cline{2-25}
& \bfseries Average & 0.99 & 1.14 & 1.14 & 1.05 & 1.20 & 0.97 & 1.26 & 1.19 & 0.89 & 1.05 & 1.29 & 1.05 & 0.96 & 1.19 & 0.89 & 1.17 & 1.04 & 1.11 & 1.16 & 1.36 & 1.09 & 1.29 & 1.11 \\
\hline
\end{tabular}
}
\caption{Per-language quality meta-evaluation of texts generated by each generator. Mean of majority-voted (out of three meta-evaluators) scores of 10 samples for each combination are provided.}
\label{tab:metaevaluation_texts}
%\vspace{-3mm}
\end{table*}

Similarly, we have used such meta-evaluation for quality assessment of the final texts generated by the selected generators. For this purpose, we have used a balanced subset of texts (10 samples per 22 languages per 7 generators and 1 human source, i.e. 1760 samples), resulting in 1752 evaluated samples (due to only 2 samples remained available from Gemini for Scottish Gaelic). The meta-evaluation scores given by the three meta-evaluators (pairwise \textit{Pearson correlation coefficient} averaging to $0.69$) are combined using the majority voting. The results, summarized in Table~\ref{tab:metaevaluation_texts}, indicate that the \textbf{LLM generators generated texts of similar or higher quality across languages than the quality of original human texts}. The reason might be an informal style used at social media (mistakes, spell errors, slang), which is more difficult for language models to follow (usually pre-trained on more formal web content). On average, the worst quality texts are generated by OPT-IML-Max-30B (failing mostly for non-Latin languages, still being on par with human text quality on average), while the best quality is provided by Gemini and Aya-101 models. Although not balanced across platforms (due to not each language being represented), meta-evaluation revealed the lowest quality of texts from Discord, followed by Telegram, and the highest quality of texts from Twitter.

\subsection{Limited Bias Analysis}
\label{sec:biasanalysis}

\begin{figure}[!t]
\centering
\includegraphics[width=\linewidth]{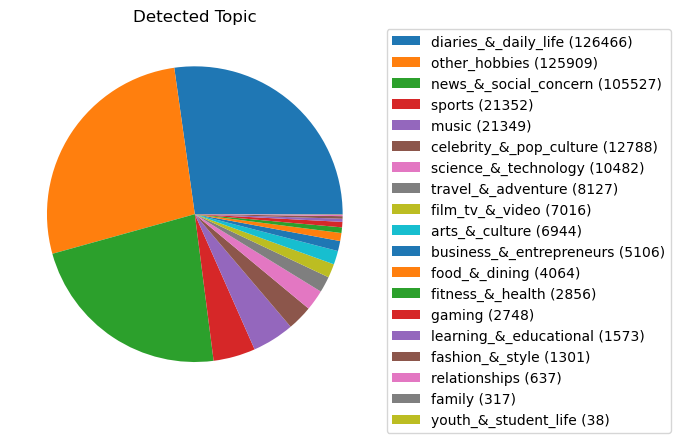}
\caption{Detected topics in the MultiSocial dataset.}
\label{fig:topics}
\end{figure}

\begin{figure}[!t]
\centering
\includegraphics[width=\linewidth]{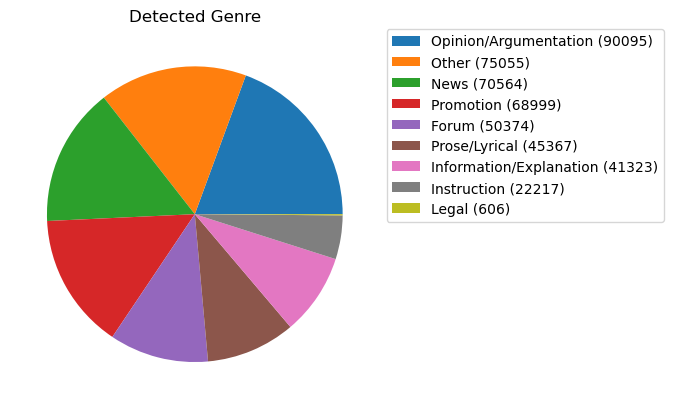}
\caption{Detected genres in the MultiSocial dataset.}
\label{fig:genres}
\end{figure}

To minimize bias in the proposed dataset, we have run multiple existing detectors for data analysis.
Based on the multilingual toxicity detector\footnote{\scriptsize\url{https://huggingface.co/textdetox/xlmr-large-toxicity-classifier}} \citep{dementieva2024overview}, about 8\% of the text samples are probably toxic (ranging from 5\% in WhatsApp to 10\% in Twitter parts).
Based on the social media text topic detector\footnote{\scriptsize\url{https://huggingface.co/cardiffnlp/tweet-topic-latest-multi}} \citep{antypas-etal-2022-twitter}, which is English-only, the topic distribution is illustrated in \figurename~\ref{fig:topics}.
Based on the multilingual text genre detector\footnote{\scriptsize\url{https://huggingface.co/classla/xlm-roberta-base-multilingual-text-genre-classifier}} \citep{kuzman2023automatic}, the genre distribution is illustrated in \figurename~\ref{fig:genres}. Although the used detection cannot be considered thorough (using existing detectors in zero-shot manner cannot be considered fully accurate), when used just as an indication, we can see that the proposed dataset texts are distributed among various topics and genres; thus, limiting the presence of such a bias.


\section{Fine-tuning Settings}
\label{sec:finetuning}

For the fine-tuning process of the fine-tuned detection methods, we have used a parameter efficient fine-tuning (\textbf{PEFT}) technique called \textbf{QLoRA} \citep{dettmers2023qlora} with default parameters (except for \textit{target\_modules} set to \textit{query\_key\_value} and  $r=4$). The training process used the AdamW optimizer with the linear scheduler and the learning rate of $2E^{-4}$. Batch size of 2 with gradient accumulation steps of 8 have been used. We have used fixed 1 epoch for training (7 epochs in case of smaller subset selection), but also limiting the models training process to 48 hours (checkpointing each 20\% of the epoch). All the settings can be found in the source code available in the published repository, enabling full replication. We aimed to use the same training settings across the various detection models training; however, we have used full fine-tuning for the XLM-RoBERTa-large model instead of QLoRA due to lack of support.

Due to a high class imbalance when using multiple generators data, we have experimented with various class balancing strategies for training. Namely, no balancing, majority-class downsampling, minority-class upsampling, mixed up-down-sampling (duplicating the minority-class samples just once and afterwards downsampling the majority-class). The results (using accuracy and AUC ROC metrics) indicated that there is a negligible effect on performance based on the used strategy, while no balancing having slower learning ability (however, the performance is eventually competitive with the others). Since having no significant impact on the performance, we have used majority-class downsampling to limit the number of steps in epoch.

\section{Cross-domain Evaluation}
\label{sec:cross-domain}

For the evaluation on the out-of-domain data, we use the news articles of MULTITuDE \citep{macko-etal-2023-multitude} benchmark. We have used the published scripts\footnote{\scriptsize\url{https://github.com/kinit-sk/mgt-detection-benchmark}} to extend the test set to our selection of languages. For the text generation, we have used the same models as in the proposed MultiSocial dataset (to evaluate only cross-domain capability), while we used Llama-2-70b model instead for Gemini for out-of-distribution (cross-generator) evaluation.

In Table~\ref{tab:benchmark_crossdomain}~and~\ref{tab:benchmark_ood}, the results are provided in the same way as in Table~\ref{tab:benchmark}, while testing on MULTITuDE (news domain) data. In pure cross-domain evaluation (Table~\ref{tab:benchmark_crossdomain}), the machine texts are generated by the same generators (as used for training), in out-of-distribution evaluation (Table~\ref{tab:benchmark_ood}), the machine texts are generated only by Llama-2-70b (not available in MultiSocial for training).

\begin{table}[!t]
%\vspace{-3mm}
\centering
\resizebox{\linewidth}{!}{
\addtolength{\tabcolsep}{-4pt}
\begin{tabular}{cm{6cm}@{}cc}
\hline
\bfseries Rank & \bfseries Detector & \bfseries AUC ROC & \bfseries \begin{tabular}{@{}c@{}}MacroF1\\ @5\%FPR\end{tabular} \\
\hline
{\cellcolor[HTML]{B6D7A8}} 1 & {\cellcolor[HTML]{B6D7A8}} Llama-3-8b-MultiSocial & {\cellcolor[HTML]{B6D7A8}} 0.9273 & {\cellcolor[HTML]{B6D7A8}} 0.7988 \\
{\cellcolor[HTML]{B6D7A8}} 2 & {\cellcolor[HTML]{B6D7A8}} Aya-101-MultiSocial & {\cellcolor[HTML]{B6D7A8}} 0.9262 & {\cellcolor[HTML]{B6D7A8}} 0.8008 \\
{\cellcolor[HTML]{B6D7A8}} 3 & {\cellcolor[HTML]{B6D7A8}} mDeBERTa-v3-base-MultiSocial & {\cellcolor[HTML]{B6D7A8}} 0.9025 & {\cellcolor[HTML]{B6D7A8}} 0.7512 \\
{\cellcolor[HTML]{B6D7A8}} 4 & {\cellcolor[HTML]{B6D7A8}} Mistral-7b-v0.1-MultiSocial & {\cellcolor[HTML]{B6D7A8}} 0.8988 & {\cellcolor[HTML]{B6D7A8}} 0.7937 \\
{\cellcolor[HTML]{B6D7A8}} 5 & {\cellcolor[HTML]{B6D7A8}} XLM-RoBERTa-large-MultiSocial & {\cellcolor[HTML]{B6D7A8}} 0.8309 & {\cellcolor[HTML]{B6D7A8}} 0.7306 \\
{\cellcolor[HTML]{F9CB9C}} 6 & {\cellcolor[HTML]{F9CB9C}} Binoculars & {\cellcolor[HTML]{F9CB9C}} 0.8303 & {\cellcolor[HTML]{F9CB9C}} 0.4041 \\
{\cellcolor[HTML]{F9CB9C}} 7 & {\cellcolor[HTML]{F9CB9C}} Fast-Detect-GPT & {\cellcolor[HTML]{F9CB9C}} 0.8104 & {\cellcolor[HTML]{F9CB9C}} 0.6361 \\
{\cellcolor[HTML]{B6D7A8}} 8 & {\cellcolor[HTML]{B6D7A8}} Falcon-rw-1b-MultiSocial & {\cellcolor[HTML]{B6D7A8}} 0.7592 & {\cellcolor[HTML]{B6D7A8}} 0.6394 \\
{\cellcolor[HTML]{B6D7A8}} 9 & {\cellcolor[HTML]{B6D7A8}} BLOOMZ-3b-MultiSocial & {\cellcolor[HTML]{B6D7A8}} 0.7071 & {\cellcolor[HTML]{B6D7A8}} 0.5731 \\
{\cellcolor[HTML]{F9CB9C}} 10 & {\cellcolor[HTML]{F9CB9C}} LLM-Deviation & {\cellcolor[HTML]{F9CB9C}} 0.6568 & {\cellcolor[HTML]{F9CB9C}} 0.3568 \\
{\cellcolor[HTML]{F9CB9C}} 11 & {\cellcolor[HTML]{F9CB9C}} DetectLLM-LRR & {\cellcolor[HTML]{F9CB9C}} 0.6496 & {\cellcolor[HTML]{F9CB9C}} 0.4133 \\
{\cellcolor[HTML]{F9CB9C}} 12 & {\cellcolor[HTML]{F9CB9C}} S5 & {\cellcolor[HTML]{F9CB9C}} 0.6336 & {\cellcolor[HTML]{F9CB9C}} 0.3519 \\
{\cellcolor[HTML]{9FC5E8}} 13 & {\cellcolor[HTML]{9FC5E8}} Longformer Detector & {\cellcolor[HTML]{9FC5E8}} 0.6157 & {\cellcolor[HTML]{9FC5E8}} 0.2564 \\
{\cellcolor[HTML]{9FC5E8}} 14 & {\cellcolor[HTML]{9FC5E8}} ChatGPT-Detector-RoBERTa-Chinese & {\cellcolor[HTML]{9FC5E8}} 0.5896 & {\cellcolor[HTML]{9FC5E8}} 0.4296 \\
{\cellcolor[HTML]{9FC5E8}} 15 & {\cellcolor[HTML]{9FC5E8}} RoBERTa-large-OpenAI-Detector & {\cellcolor[HTML]{9FC5E8}} 0.5707 & {\cellcolor[HTML]{9FC5E8}} 0.1958 \\
{\cellcolor[HTML]{9FC5E8}} 16 & {\cellcolor[HTML]{9FC5E8}} BLOOMZ-3b-mixed-Detector & {\cellcolor[HTML]{9FC5E8}} 0.5536 & {\cellcolor[HTML]{9FC5E8}} 0.1891 \\
{\cellcolor[HTML]{9FC5E8}} 17 & {\cellcolor[HTML]{9FC5E8}} ruRoBERTa-ruatd-binary & {\cellcolor[HTML]{9FC5E8}} 0.5186 & {\cellcolor[HTML]{9FC5E8}} 0.1485 \\
\hline
\end{tabular}
}
\caption{Cross-domain evaluation of the selected MGTD methods of \colorbox[HTML]{F9CB9C}{statistical}, \colorbox[HTML]{9FC5E8}{pre-trained}, and \colorbox[HTML]{B6D7A8}{fine-tuned} categories.}
\label{tab:benchmark_crossdomain}
%\vspace{-3mm}
\end{table}

\begin{table}[!t]
%\vspace{-3mm}
\centering
\resizebox{\linewidth}{!}{
\addtolength{\tabcolsep}{-4pt}
\begin{tabular}{cm{6cm}@{}cc}
\hline
\bfseries Rank & \bfseries Detector & \bfseries AUC ROC & \bfseries \begin{tabular}{@{}c@{}}MacroF1\\ @5\%FPR\end{tabular} \\
\hline
{\cellcolor[HTML]{F9CB9C}} 1 & {\cellcolor[HTML]{F9CB9C}} Fast-Detect-GPT & {\cellcolor[HTML]{F9CB9C}} 0.9238 & {\cellcolor[HTML]{F9CB9C}} 0.8471 \\
{\cellcolor[HTML]{F9CB9C}} 2 & {\cellcolor[HTML]{F9CB9C}} Binoculars & {\cellcolor[HTML]{F9CB9C}} 0.9048 & {\cellcolor[HTML]{F9CB9C}} 0.7568 \\
{\cellcolor[HTML]{B6D7A8}} 3 & {\cellcolor[HTML]{B6D7A8}} mDeBERTa-v3-base-MultiSocial & {\cellcolor[HTML]{B6D7A8}} 0.8871 & {\cellcolor[HTML]{B6D7A8}} 0.8011 \\
{\cellcolor[HTML]{B6D7A8}} 4 & {\cellcolor[HTML]{B6D7A8}} Mistral-7b-v0.1-MultiSocial & {\cellcolor[HTML]{B6D7A8}} 0.8614 & {\cellcolor[HTML]{B6D7A8}} 0.7673 \\
{\cellcolor[HTML]{B6D7A8}} 5 & {\cellcolor[HTML]{B6D7A8}} Aya-101-MultiSocial & {\cellcolor[HTML]{B6D7A8}} 0.8574 & {\cellcolor[HTML]{B6D7A8}} 0.7556 \\
{\cellcolor[HTML]{B6D7A8}} 6 & {\cellcolor[HTML]{B6D7A8}} Llama-3-8b-MultiSocial & {\cellcolor[HTML]{B6D7A8}} 0.8549 & {\cellcolor[HTML]{B6D7A8}} 0.7352 \\
{\cellcolor[HTML]{B6D7A8}} 7 & {\cellcolor[HTML]{B6D7A8}} XLM-RoBERTa-large-MultiSocial & {\cellcolor[HTML]{B6D7A8}} 0.7928 & {\cellcolor[HTML]{B6D7A8}} 0.7108 \\
{\cellcolor[HTML]{B6D7A8}} 8 & {\cellcolor[HTML]{B6D7A8}} Falcon-rw-1b-MultiSocial & {\cellcolor[HTML]{B6D7A8}} 0.7710 & {\cellcolor[HTML]{B6D7A8}} 0.6912 \\
{\cellcolor[HTML]{F9CB9C}} 9 & {\cellcolor[HTML]{F9CB9C}} DetectLLM-LRR & {\cellcolor[HTML]{F9CB9C}} 0.7559 & {\cellcolor[HTML]{F9CB9C}} 0.7121 \\
{\cellcolor[HTML]{F9CB9C}} 10 & {\cellcolor[HTML]{F9CB9C}} LLM-Deviation & {\cellcolor[HTML]{F9CB9C}} 0.7257 & {\cellcolor[HTML]{F9CB9C}} 0.5931 \\
{\cellcolor[HTML]{9FC5E8}} 11 & {\cellcolor[HTML]{9FC5E8}} Longformer Detector & {\cellcolor[HTML]{9FC5E8}} 0.7032 & {\cellcolor[HTML]{9FC5E8}} 0.5018 \\
{\cellcolor[HTML]{F9CB9C}} 12 & {\cellcolor[HTML]{F9CB9C}} S5 & {\cellcolor[HTML]{F9CB9C}} 0.6849 & {\cellcolor[HTML]{F9CB9C}} 0.5506 \\
{\cellcolor[HTML]{9FC5E8}} 13 & {\cellcolor[HTML]{9FC5E8}} ChatGPT-Detector-RoBERTa-Chinese & {\cellcolor[HTML]{9FC5E8}} 0.6788 & {\cellcolor[HTML]{9FC5E8}} 0.6161 \\
{\cellcolor[HTML]{B6D7A8}} 14 & {\cellcolor[HTML]{B6D7A8}} BLOOMZ-3b-MultiSocial & {\cellcolor[HTML]{B6D7A8}} 0.6515 & {\cellcolor[HTML]{B6D7A8}} 0.6026 \\
{\cellcolor[HTML]{9FC5E8}} 15 & {\cellcolor[HTML]{9FC5E8}} RoBERTa-large-OpenAI-Detector & {\cellcolor[HTML]{9FC5E8}} 0.5596 & {\cellcolor[HTML]{9FC5E8}} 0.4315 \\
{\cellcolor[HTML]{9FC5E8}} 16 & {\cellcolor[HTML]{9FC5E8}} ruRoBERTa-ruatd-binary & {\cellcolor[HTML]{9FC5E8}} 0.5327 & {\cellcolor[HTML]{9FC5E8}} 0.3473 \\
{\cellcolor[HTML]{9FC5E8}} 17 & {\cellcolor[HTML]{9FC5E8}} BLOOMZ-3b-mixed-Detector & {\cellcolor[HTML]{9FC5E8}} 0.5212 & {\cellcolor[HTML]{9FC5E8}} 0.4096 \\
\hline
\end{tabular}
}
\caption{Out-of-distribution (cross-domain and cross-generator) evaluation of the selected MGTD methods of \colorbox[HTML]{F9CB9C}{statistical}, \colorbox[HTML]{9FC5E8}{pre-trained}, and \colorbox[HTML]{B6D7A8}{fine-tuned} categories.}
\label{tab:benchmark_ood}
%\vspace{-3mm}
\end{table}

\section{Ablation Study}
\label{sec:ablation}

\begin{table*}
\centering
\resizebox{\textwidth}{!}{
\addtolength{\tabcolsep}{-2pt}
\begin{tabular}{c|cccccccccccccccccccccc|c}
\hline
\bfseries Train & \multicolumn{23}{c}{\bfseries Test Language [AUC ROC mean]} \\
\bfseries Language & \bfseries ar & \bfseries bg & \bfseries ca & \bfseries cs & \bfseries de & \bfseries el & \bfseries en & \bfseries es & \bfseries et & \bfseries ga & \bfseries gd & \bfseries hr & \bfseries hu & \bfseries nl & \bfseries pl & \bfseries pt & \bfseries ro & \bfseries ru & \bfseries sk & \bfseries sl & \bfseries uk & \bfseries zh & \bfseries all \\
\hline
\bfseries en & {\cellcolor[HTML]{C6CCE3}} \color[HTML]{000000} 0.78 & {\cellcolor[HTML]{ABBFDC}} \color[HTML]{000000} 0.86 & {\cellcolor[HTML]{C9CEE4}} \color[HTML]{000000} 0.77 & {\cellcolor[HTML]{A7BDDB}} \color[HTML]{000000} 0.87 & {\cellcolor[HTML]{A5BDDB}} \color[HTML]{000000} 0.88 & {\cellcolor[HTML]{A4BCDA}} \color[HTML]{000000} 0.88 & {\cellcolor[HTML]{84B0D3}} \color[HTML]{000000} 0.96 & {\cellcolor[HTML]{B0C2DE}} \color[HTML]{000000} 0.85 & {\cellcolor[HTML]{9FBAD9}} \color[HTML]{000000} 0.89 & N/A & N/A & {\cellcolor[HTML]{94B6D7}} \color[HTML]{000000} 0.92 & {\cellcolor[HTML]{89B1D4}} \color[HTML]{000000} 0.95 & {\cellcolor[HTML]{B8C6E0}} \color[HTML]{000000} 0.82 & {\cellcolor[HTML]{A9BFDC}} \color[HTML]{000000} 0.86 & {\cellcolor[HTML]{9AB8D8}} \color[HTML]{000000} 0.90 & {\cellcolor[HTML]{93B5D6}} \color[HTML]{000000} 0.92 & {\cellcolor[HTML]{AFC1DD}} \color[HTML]{000000} 0.85 & N/A & N/A & {\cellcolor[HTML]{C4CBE3}} \color[HTML]{000000} 0.79 & {\cellcolor[HTML]{E7E3F0}} \color[HTML]{000000} 0.65 & {\cellcolor[HTML]{B3C3DE}} \color[HTML]{000000} 0.84 \\
\bfseries es & {\cellcolor[HTML]{C5CCE3}} \color[HTML]{000000} 0.78 & {\cellcolor[HTML]{B3C3DE}} \color[HTML]{000000} 0.84 & {\cellcolor[HTML]{B4C4DF}} \color[HTML]{000000} 0.83 & {\cellcolor[HTML]{B4C4DF}} \color[HTML]{000000} 0.83 & {\cellcolor[HTML]{A2BCDA}} \color[HTML]{000000} 0.88 & {\cellcolor[HTML]{B3C3DE}} \color[HTML]{000000} 0.84 & {\cellcolor[HTML]{A7BDDB}} \color[HTML]{000000} 0.87 & {\cellcolor[HTML]{8EB3D5}} \color[HTML]{000000} 0.94 & {\cellcolor[HTML]{ABBFDC}} \color[HTML]{000000} 0.86 & N/A & N/A & {\cellcolor[HTML]{A4BCDA}} \color[HTML]{000000} 0.88 & {\cellcolor[HTML]{94B6D7}} \color[HTML]{000000} 0.92 & {\cellcolor[HTML]{BCC7E1}} \color[HTML]{000000} 0.81 & {\cellcolor[HTML]{B1C2DE}} \color[HTML]{000000} 0.84 & {\cellcolor[HTML]{97B7D7}} \color[HTML]{000000} 0.91 & {\cellcolor[HTML]{99B8D8}} \color[HTML]{000000} 0.91 & {\cellcolor[HTML]{ACC0DD}} \color[HTML]{000000} 0.86 & N/A & N/A & {\cellcolor[HTML]{C4CBE3}} \color[HTML]{000000} 0.79 & {\cellcolor[HTML]{E9E5F1}} \color[HTML]{000000} 0.64 & {\cellcolor[HTML]{B8C6E0}} \color[HTML]{000000} 0.82 \\
\bfseries ru & {\cellcolor[HTML]{D6D6E9}} \color[HTML]{000000} 0.73 & {\cellcolor[HTML]{9EBAD9}} \color[HTML]{000000} 0.90 & {\cellcolor[HTML]{E0DEED}} \color[HTML]{000000} 0.68 & {\cellcolor[HTML]{BCC7E1}} \color[HTML]{000000} 0.81 & {\cellcolor[HTML]{BDC8E1}} \color[HTML]{000000} 0.81 & {\cellcolor[HTML]{B1C2DE}} \color[HTML]{000000} 0.84 & {\cellcolor[HTML]{B5C4DF}} \color[HTML]{000000} 0.83 & {\cellcolor[HTML]{CDD0E5}} \color[HTML]{000000} 0.76 & {\cellcolor[HTML]{BBC7E0}} \color[HTML]{000000} 0.82 & N/A & N/A & {\cellcolor[HTML]{B1C2DE}} \color[HTML]{000000} 0.84 & {\cellcolor[HTML]{ACC0DD}} \color[HTML]{000000} 0.86 & {\cellcolor[HTML]{D2D2E7}} \color[HTML]{000000} 0.74 & {\cellcolor[HTML]{B9C6E0}} \color[HTML]{000000} 0.82 & {\cellcolor[HTML]{BCC7E1}} \color[HTML]{000000} 0.81 & {\cellcolor[HTML]{B5C4DF}} \color[HTML]{000000} 0.83 & {\cellcolor[HTML]{91B5D6}} \color[HTML]{000000} 0.93 & N/A & N/A & {\cellcolor[HTML]{A8BEDC}} \color[HTML]{000000} 0.87 & {\cellcolor[HTML]{EFE9F3}} \color[HTML]{000000} 0.61 & {\cellcolor[HTML]{C8CDE4}} \color[HTML]{000000} 0.77 \\
\hline
\bfseries {en-es-ru} & {\cellcolor[HTML]{A5BDDB}} \color[HTML]{000000} 0.88 & {\cellcolor[HTML]{93B5D6}} \color[HTML]{000000} 0.92 & {\cellcolor[HTML]{ACC0DD}} \color[HTML]{000000} 0.86 & {\cellcolor[HTML]{97B7D7}} \color[HTML]{000000} 0.91 & {\cellcolor[HTML]{97B7D7}} \color[HTML]{000000} 0.91 & {\cellcolor[HTML]{8FB4D6}} \color[HTML]{000000} 0.93 & {\cellcolor[HTML]{86B0D3}} \color[HTML]{000000} 0.95 & {\cellcolor[HTML]{8EB3D5}} \color[HTML]{000000} 0.94 & {\cellcolor[HTML]{97B7D7}} \color[HTML]{000000} 0.91 & N/A & N/A & {\cellcolor[HTML]{93B5D6}} \color[HTML]{000000} 0.93 & {\cellcolor[HTML]{83AFD3}} \color[HTML]{000000} 0.96 & {\cellcolor[HTML]{B0C2DE}} \color[HTML]{000000} 0.84 & {\cellcolor[HTML]{A1BBDA}} \color[HTML]{000000} 0.89 & {\cellcolor[HTML]{8FB4D6}} \color[HTML]{000000} 0.93 & {\cellcolor[HTML]{8EB3D5}} \color[HTML]{000000} 0.93 & {\cellcolor[HTML]{91B5D6}} \color[HTML]{000000} 0.93 & N/A & N/A & {\cellcolor[HTML]{A2BCDA}} \color[HTML]{000000} 0.88 & {\cellcolor[HTML]{B3C3DE}} \color[HTML]{000000} 0.84 & {\cellcolor[HTML]{9EBAD9}} \color[HTML]{000000} 0.90 \\
\hline
\end{tabular}
}
\vspace{-1mm}
\caption{Cross-lingual mean AUC ROC performance of the MGT detectors with autoregressive foundational models fine-tuned monolingually (\textit{en}, \textit{es} and \textit{ru}) and multilingually (\textit{en-es-ru}), evaluated based on Telegram data (for training as well as for testing).  N/A refers to not enough samples (at least 2000) in MultiSocial Telegram data.}
\label{tab:ablation_crosslingual_autoregressive}
\vspace{-3mm}
\end{table*}

\begin{table*}
\centering
\resizebox{\textwidth}{!}{
\addtolength{\tabcolsep}{-2pt}
\begin{tabular}{c|cccccccccccccccccccccc|c}
\hline
\bfseries Train & \multicolumn{23}{c}{\bfseries Test Language [AUC ROC mean]} \\
\bfseries Language & \bfseries ar & \bfseries bg & \bfseries ca & \bfseries cs & \bfseries de & \bfseries el & \bfseries en & \bfseries es & \bfseries et & \bfseries ga & \bfseries gd & \bfseries hr & \bfseries hu & \bfseries nl & \bfseries pl & \bfseries pt & \bfseries ro & \bfseries ru & \bfseries sk & \bfseries sl & \bfseries uk & \bfseries zh & \bfseries all \\
\hline
\bfseries en & {\cellcolor[HTML]{AFC1DD}} \color[HTML]{000000} 0.85 & {\cellcolor[HTML]{88B1D4}} \color[HTML]{000000} 0.95 & {\cellcolor[HTML]{C2CBE2}} \color[HTML]{000000} 0.79 & {\cellcolor[HTML]{84B0D3}} \color[HTML]{000000} 0.96 & {\cellcolor[HTML]{A4BCDA}} \color[HTML]{000000} 0.88 & {\cellcolor[HTML]{8EB3D5}} \color[HTML]{000000} 0.93 & {\cellcolor[HTML]{83AFD3}} \color[HTML]{000000} 0.96 & {\cellcolor[HTML]{9AB8D8}} \color[HTML]{000000} 0.90 & {\cellcolor[HTML]{88B1D4}} \color[HTML]{000000} 0.95 & N/A & N/A & {\cellcolor[HTML]{83AFD3}} \color[HTML]{000000} 0.96 & {\cellcolor[HTML]{78ABD0}} \color[HTML]{000000} 0.99 & {\cellcolor[HTML]{ABBFDC}} \color[HTML]{000000} 0.86 & {\cellcolor[HTML]{91B5D6}} \color[HTML]{000000} 0.93 & {\cellcolor[HTML]{8BB2D4}} \color[HTML]{000000} 0.94 & {\cellcolor[HTML]{83AFD3}} \color[HTML]{000000} 0.96 & {\cellcolor[HTML]{9AB8D8}} \color[HTML]{000000} 0.90 & N/A & N/A & {\cellcolor[HTML]{AFC1DD}} \color[HTML]{000000} 0.85 & {\cellcolor[HTML]{B5C4DF}} \color[HTML]{000000} 0.83 & {\cellcolor[HTML]{99B8D8}} \color[HTML]{000000} 0.91 \\
\bfseries es & {\cellcolor[HTML]{A4BCDA}} \color[HTML]{000000} 0.88 & {\cellcolor[HTML]{86B0D3}} \color[HTML]{000000} 0.96 & {\cellcolor[HTML]{ABBFDC}} \color[HTML]{000000} 0.86 & {\cellcolor[HTML]{86B0D3}} \color[HTML]{000000} 0.96 & {\cellcolor[HTML]{99B8D8}} \color[HTML]{000000} 0.91 & {\cellcolor[HTML]{96B6D7}} \color[HTML]{000000} 0.92 & {\cellcolor[HTML]{8CB3D5}} \color[HTML]{000000} 0.94 & {\cellcolor[HTML]{8CB3D5}} \color[HTML]{000000} 0.94 & {\cellcolor[HTML]{89B1D4}} \color[HTML]{000000} 0.95 & N/A & N/A & {\cellcolor[HTML]{84B0D3}} \color[HTML]{000000} 0.96 & {\cellcolor[HTML]{79ABD0}} \color[HTML]{000000} 0.99 & {\cellcolor[HTML]{A7BDDB}} \color[HTML]{000000} 0.87 & {\cellcolor[HTML]{91B5D6}} \color[HTML]{000000} 0.93 & {\cellcolor[HTML]{8BB2D4}} \color[HTML]{000000} 0.94 & {\cellcolor[HTML]{83AFD3}} \color[HTML]{000000} 0.96 & {\cellcolor[HTML]{94B6D7}} \color[HTML]{000000} 0.92 & N/A & N/A & {\cellcolor[HTML]{A8BEDC}} \color[HTML]{000000} 0.87 & {\cellcolor[HTML]{ABBFDC}} \color[HTML]{000000} 0.86 & {\cellcolor[HTML]{96B6D7}} \color[HTML]{000000} 0.92 \\
\bfseries ru & {\cellcolor[HTML]{97B7D7}} \color[HTML]{000000} 0.91 & {\cellcolor[HTML]{7EADD1}} \color[HTML]{000000} 0.97 & {\cellcolor[HTML]{ACC0DD}} \color[HTML]{000000} 0.86 & {\cellcolor[HTML]{86B0D3}} \color[HTML]{000000} 0.96 & {\cellcolor[HTML]{A2BCDA}} \color[HTML]{000000} 0.88 & {\cellcolor[HTML]{8EB3D5}} \color[HTML]{000000} 0.94 & {\cellcolor[HTML]{8EB3D5}} \color[HTML]{000000} 0.94 & {\cellcolor[HTML]{9AB8D8}} \color[HTML]{000000} 0.90 & {\cellcolor[HTML]{84B0D3}} \color[HTML]{000000} 0.96 & N/A & N/A & {\cellcolor[HTML]{84B0D3}} \color[HTML]{000000} 0.96 & {\cellcolor[HTML]{79ABD0}} \color[HTML]{000000} 0.98 & {\cellcolor[HTML]{ABBFDC}} \color[HTML]{000000} 0.86 & {\cellcolor[HTML]{8BB2D4}} \color[HTML]{000000} 0.94 & {\cellcolor[HTML]{91B5D6}} \color[HTML]{000000} 0.93 & {\cellcolor[HTML]{86B0D3}} \color[HTML]{000000} 0.95 & {\cellcolor[HTML]{86B0D3}} \color[HTML]{000000} 0.96 & N/A & N/A & {\cellcolor[HTML]{93B5D6}} \color[HTML]{000000} 0.92 & {\cellcolor[HTML]{9CB9D9}} \color[HTML]{000000} 0.90 & {\cellcolor[HTML]{91B5D6}} \color[HTML]{000000} 0.93 \\
\hline
\bfseries {en-es-ru} & {\cellcolor[HTML]{9AB8D8}} \color[HTML]{000000} 0.91 & {\cellcolor[HTML]{80AED2}} \color[HTML]{000000} 0.97 & {\cellcolor[HTML]{A8BEDC}} \color[HTML]{000000} 0.87 & {\cellcolor[HTML]{83AFD3}} \color[HTML]{000000} 0.96 & {\cellcolor[HTML]{99B8D8}} \color[HTML]{000000} 0.91 & {\cellcolor[HTML]{8FB4D6}} \color[HTML]{000000} 0.93 & {\cellcolor[HTML]{86B0D3}} \color[HTML]{000000} 0.96 & {\cellcolor[HTML]{91B5D6}} \color[HTML]{000000} 0.93 & {\cellcolor[HTML]{84B0D3}} \color[HTML]{000000} 0.96 & N/A & N/A & {\cellcolor[HTML]{83AFD3}} \color[HTML]{000000} 0.96 & {\cellcolor[HTML]{78ABD0}} \color[HTML]{000000} 0.99 & {\cellcolor[HTML]{A4BCDA}} \color[HTML]{000000} 0.88 & {\cellcolor[HTML]{8CB3D5}} \color[HTML]{000000} 0.94 & {\cellcolor[HTML]{89B1D4}} \color[HTML]{000000} 0.95 & {\cellcolor[HTML]{81AED2}} \color[HTML]{000000} 0.97 & {\cellcolor[HTML]{8BB2D4}} \color[HTML]{000000} 0.94 & N/A & N/A & {\cellcolor[HTML]{9CB9D9}} \color[HTML]{000000} 0.90 & {\cellcolor[HTML]{A1BBDA}} \color[HTML]{000000} 0.89 & {\cellcolor[HTML]{8FB4D6}} \color[HTML]{000000} 0.93 \\
\hline
\end{tabular}
}
\vspace{-1mm}
\caption{Cross-lingual mean AUC ROC performance of the MGT detectors with non-autoregressive foundational models fine-tuned monolingually (\textit{en}, \textit{es} and \textit{ru}) and multilingually (\textit{en-es-ru}), evaluated based on Telegram data (for training as well as for testing).  N/A refers to not enough samples (at least 2000) in MultiSocial Telegram data.}
\label{tab:ablation_crosslingual_nonautoregressive}
\vspace{-3mm}
\end{table*}

Based on Table~\ref{tab:data_crosslingual}, we have identified two groups of detectors based on their performances across languages, namely autoregressive models and others. Autoregressive group includes Llama-3-8b, Mistral-7b-v0.1, BLOOMZ-3b, and Falcon-rw-1b. The non-autoregressive group includes Aya-101, XLM-RoBERTa-large, and mDeBERTa-v3-base. The summarized results, analogous to the Table~\ref{tab:crosslingual}, but provided for each group separately in Table~\ref{tab:ablation_crosslingual_autoregressive} and Table~\ref{tab:ablation_crosslingual_nonautoregressive}. There are clear differences between the results of these groups, since in case of non-autogregressive group, we can see no significant differences between monolingually and multilingually fine-tuned detectors.

Another possible explanation of such a different behavior of some detectors is their pre-training on a huge number of languages (100+ languages in case of each detector in the non-autoregressive group). However, the BLOOMZ model has also been pre-trained on 50+ languages, and still shows a behavior similar to others in the autoregressive group.

\section{Results Data}
\label{sec:data}

Tables~\ref{tab:data_perllm_statistical}-\ref{tab:data_perllm_finetuned} contain per-generator AUC ROC performance of each MGT detection method for each test language, Tables~\ref{tab:data_perplatform_statistical}-\ref{tab:data_perplatform_finetuned} contain per-platform AUC ROC performance of each MGT detection method for each test language, and
Tables~\ref{tab:data_perplatform_perllm_statistical}-\ref{tab:data_perplatform_perllm_finetuned} contains per-generator AUC ROC performance of each MGT detection method for each platform, separately for each MGTD category.
Table~\ref{tab:data_crosslingual} contains cross-lingual evaluation of differently fine-tuned MGT detectors. Table~\ref{tab:data_crossplatform} contains cross-platform evaluation of differently fine-tuned MGT detectors.

\begin{table*}
\centering
\resizebox{\textwidth}{!}{
\addtolength{\tabcolsep}{-2pt}
\begin{tabular}{c|l|cccccccccccccccccccccc|c}
\hline
 &  & \multicolumn{23}{c}{\bfseries Test Language [AUC ROC]} \\
\bfseries Detector & \bfseries Generator & \bfseries ar & \bfseries bg & \bfseries ca & \bfseries cs & \bfseries de & \bfseries el & \bfseries en & \bfseries es & \bfseries et & \bfseries ga & \bfseries gd & \bfseries hr & \bfseries hu & \bfseries nl & \bfseries pl & \bfseries pt & \bfseries ro & \bfseries ru & \bfseries sk & \bfseries sl & \bfseries uk & \bfseries zh & \bfseries all \\
\hline
\multirow[c]{7}{*}{\rotatebox{90}{\bfseries Binoculars}} & \bfseries Aya-101 & {\cellcolor[HTML]{DBDAEB}} \color[HTML]{000000} 0.70 & {\cellcolor[HTML]{EEE9F3}} \color[HTML]{000000} 0.61 & {\cellcolor[HTML]{F0EAF4}} \color[HTML]{000000} 0.60 & {\cellcolor[HTML]{DEDCEC}} \color[HTML]{000000} 0.69 & {\cellcolor[HTML]{D2D2E7}} \color[HTML]{000000} 0.75 & {\cellcolor[HTML]{C9CEE4}} \color[HTML]{000000} 0.77 & {\cellcolor[HTML]{C1CAE2}} \color[HTML]{000000} 0.79 & {\cellcolor[HTML]{D5D5E8}} \color[HTML]{000000} 0.73 & {\cellcolor[HTML]{E1DFED}} \color[HTML]{000000} 0.67 & {\cellcolor[HTML]{EDE7F2}} \color[HTML]{000000} 0.62 & {\cellcolor[HTML]{F1EBF4}} \color[HTML]{000000} 0.59 & {\cellcolor[HTML]{D6D6E9}} \color[HTML]{000000} 0.72 & {\cellcolor[HTML]{D9D8EA}} \color[HTML]{000000} 0.71 & {\cellcolor[HTML]{D9D8EA}} \color[HTML]{000000} 0.71 & {\cellcolor[HTML]{D8D7E9}} \color[HTML]{000000} 0.71 & {\cellcolor[HTML]{DDDBEC}} \color[HTML]{000000} 0.70 & {\cellcolor[HTML]{E2DFEE}} \color[HTML]{000000} 0.67 & {\cellcolor[HTML]{E2DFEE}} \color[HTML]{000000} 0.67 & {\cellcolor[HTML]{EAE6F1}} \color[HTML]{000000} 0.64 & {\cellcolor[HTML]{E3E0EE}} \color[HTML]{000000} 0.67 & {\cellcolor[HTML]{FAF2F8}} \color[HTML]{000000} 0.54 & {\cellcolor[HTML]{E6E2EF}} \color[HTML]{000000} 0.65 & {\cellcolor[HTML]{DFDDEC}} \color[HTML]{000000} 0.69 \\
\hline
\end{tabular}
}
\caption{Per-LLM AUC ROC performance of statistical MGT detectors category. N/A refers
to not enough samples per each class (at least 10) making AUC ROC value irrelevant.}
\label{tab:data_perllm_statistical}
\end{table*}

\begin{table*}
\centering
\resizebox{\textwidth}{!}{
\addtolength{\tabcolsep}{-2pt}
\begin{tabular}{c|l|cccccccccccccccccccccc|c}
\hline
 &  & \multicolumn{23}{c}{\bfseries Test Language [AUC ROC]} \\
\bfseries Detector & \bfseries Generator & \bfseries ar & \bfseries bg & \bfseries ca & \bfseries cs & \bfseries de & \bfseries el & \bfseries en & \bfseries es & \bfseries et & \bfseries ga & \bfseries gd & \bfseries hr & \bfseries hu & \bfseries nl & \bfseries pl & \bfseries pt & \bfseries ro & \bfseries ru & \bfseries sk & \bfseries sl & \bfseries uk & \bfseries zh & \bfseries all \\
\hline
\multirow[c]{7}{*}{\rotatebox{90}{\parbox{3cm}{\bfseries BLOOMZ-3b-mixed-Detector}}} & \bfseries Aya-101 & {\cellcolor[HTML]{B4C4DF}} \color[HTML]{000000} 0.83 & {\cellcolor[HTML]{A5BDDB}} \color[HTML]{000000} 0.88 & {\cellcolor[HTML]{B9C6E0}} \color[HTML]{000000} 0.82 & {\cellcolor[HTML]{A5BDDB}} \color[HTML]{000000} 0.88 & {\cellcolor[HTML]{AFC1DD}} \color[HTML]{000000} 0.85 & {\cellcolor[HTML]{B9C6E0}} \color[HTML]{000000} 0.82 & {\cellcolor[HTML]{B0C2DE}} \color[HTML]{000000} 0.85 & {\cellcolor[HTML]{B1C2DE}} \color[HTML]{000000} 0.84 & {\cellcolor[HTML]{99B8D8}} \color[HTML]{000000} 0.91 & {\cellcolor[HTML]{C1CAE2}} \color[HTML]{000000} 0.79 & {\cellcolor[HTML]{E0DEED}} \color[HTML]{000000} 0.68 & {\cellcolor[HTML]{ACC0DD}} \color[HTML]{000000} 0.86 & {\cellcolor[HTML]{93B5D6}} \color[HTML]{000000} 0.92 & {\cellcolor[HTML]{B3C3DE}} \color[HTML]{000000} 0.84 & {\cellcolor[HTML]{B1C2DE}} \color[HTML]{000000} 0.84 & {\cellcolor[HTML]{AFC1DD}} \color[HTML]{000000} 0.85 & {\cellcolor[HTML]{C5CCE3}} \color[HTML]{000000} 0.78 & {\cellcolor[HTML]{C8CDE4}} \color[HTML]{000000} 0.78 & {\cellcolor[HTML]{B5C4DF}} \color[HTML]{000000} 0.83 & {\cellcolor[HTML]{CCCFE5}} \color[HTML]{000000} 0.76 & {\cellcolor[HTML]{BBC7E0}} \color[HTML]{000000} 0.81 & {\cellcolor[HTML]{ABBFDC}} \color[HTML]{000000} 0.86 & {\cellcolor[HTML]{B4C4DF}} \color[HTML]{000000} 0.83 \\
\hline
\end{tabular}
}
\caption{Per-LLM AUC ROC performance of pre-trained MGT detectors category. N/A refers
to not enough samples per each class (at least 10) making AUC ROC value irrelevant.}
\label{tab:data_perllm_pretrained}
\end{table*}

\begin{table*}
\centering
\resizebox{\textwidth}{!}{
\addtolength{\tabcolsep}{-2pt}
\begin{tabular}{c|l|cccccccccccccccccccccc|c}
\hline
 &  & \multicolumn{23}{c}{\bfseries Test Language [AUC ROC]} \\
\bfseries Detector & \bfseries Generator & \bfseries ar & \bfseries bg & \bfseries ca & \bfseries cs & \bfseries de & \bfseries el & \bfseries en & \bfseries es & \bfseries et & \bfseries ga & \bfseries gd & \bfseries hr & \bfseries hu & \bfseries nl & \bfseries pl & \bfseries pt & \bfseries ro & \bfseries ru & \bfseries sk & \bfseries sl & \bfseries uk & \bfseries zh & \bfseries all \\
\hline
\multirow[c]{7}{*}{\rotatebox{90}{\parbox{3cm}{\bfseries Aya-101-MultiSocial}}} & \bfseries Aya-101 & {\cellcolor[HTML]{93B5D6}} \color[HTML]{000000} 0.92 & {\cellcolor[HTML]{7BACD1}} \color[HTML]{000000} 0.98 & {\cellcolor[HTML]{81AED2}} \color[HTML]{000000} 0.97 & {\cellcolor[HTML]{7BACD1}} \color[HTML]{000000} 0.98 & {\cellcolor[HTML]{81AED2}} \color[HTML]{000000} 0.97 & {\cellcolor[HTML]{86B0D3}} \color[HTML]{000000} 0.96 & {\cellcolor[HTML]{86B0D3}} \color[HTML]{000000} 0.96 & {\cellcolor[HTML]{84B0D3}} \color[HTML]{000000} 0.96 & {\cellcolor[HTML]{7DACD1}} \color[HTML]{000000} 0.98 & {\cellcolor[HTML]{94B6D7}} \color[HTML]{000000} 0.92 & {\cellcolor[HTML]{9AB8D8}} \color[HTML]{000000} 0.91 & {\cellcolor[HTML]{7EADD1}} \color[HTML]{000000} 0.98 & {\cellcolor[HTML]{76AAD0}} \color[HTML]{000000} 0.99 & {\cellcolor[HTML]{81AED2}} \color[HTML]{000000} 0.97 & {\cellcolor[HTML]{81AED2}} \color[HTML]{000000} 0.97 & {\cellcolor[HTML]{84B0D3}} \color[HTML]{000000} 0.96 & {\cellcolor[HTML]{86B0D3}} \color[HTML]{000000} 0.96 & {\cellcolor[HTML]{91B5D6}} \color[HTML]{000000} 0.93 & {\cellcolor[HTML]{80AED2}} \color[HTML]{000000} 0.97 & {\cellcolor[HTML]{96B6D7}} \color[HTML]{000000} 0.92 & {\cellcolor[HTML]{9AB8D8}} \color[HTML]{000000} 0.90 & {\cellcolor[HTML]{8FB4D6}} \color[HTML]{000000} 0.93 & {\cellcolor[HTML]{86B0D3}} \color[HTML]{000000} 0.96 \\
\hline
\end{tabular}
}
\caption{Per-LLM AUC ROC performance of fine-tuned MGT detectors category. N/A refers
to not enough samples per each class (at least 10) making AUC ROC value irrelevant.}
\label{tab:data_perllm_finetuned}
\end{table*}

\begin{table*}
\centering
\resizebox{\textwidth}{!}{
\addtolength{\tabcolsep}{-2pt}
\begin{tabular}{c|l|cccccccccccccccccccccc|c}
\hline
 &  & \multicolumn{23}{c}{\bfseries Test Language [AUC ROC]} \\
\bfseries Detector & \bfseries Platform & \bfseries ar & \bfseries bg & \bfseries ca & \bfseries cs & \bfseries de & \bfseries el & \bfseries en & \bfseries es & \bfseries et & \bfseries ga & \bfseries gd & \bfseries hr & \bfseries hu & \bfseries nl & \bfseries pl & \bfseries pt & \bfseries ro & \bfseries ru & \bfseries sk & \bfseries sl & \bfseries uk & \bfseries zh & \bfseries all \\
\hline
\multirow[c]{5}{*}{\rotatebox{90}{\parbox{2cm}{\bfseries Binoculars}}} & \bfseries Discord & N/A & N/A & {\cellcolor[HTML]{C8CDE4}} \color[HTML]{000000} 0.78 & {\cellcolor[HTML]{CDD0E5}} \color[HTML]{000000} 0.76 & {\cellcolor[HTML]{C0C9E2}} \color[HTML]{000000} 0.80 & N/A & {\cellcolor[HTML]{ABBFDC}} \color[HTML]{000000} 0.86 & {\cellcolor[HTML]{BCC7E1}} \color[HTML]{000000} 0.81 & {\cellcolor[HTML]{D0D1E6}} \color[HTML]{000000} 0.75 & {\cellcolor[HTML]{E1DFED}} \color[HTML]{000000} 0.67 & {\cellcolor[HTML]{DCDAEB}} \color[HTML]{000000} 0.70 & {\cellcolor[HTML]{B4C4DF}} \color[HTML]{000000} 0.83 & {\cellcolor[HTML]{BBC7E0}} \color[HTML]{000000} 0.82 & {\cellcolor[HTML]{C0C9E2}} \color[HTML]{000000} 0.80 & {\cellcolor[HTML]{B3C3DE}} \color[HTML]{000000} 0.84 & {\cellcolor[HTML]{BCC7E1}} \color[HTML]{000000} 0.81 & {\cellcolor[HTML]{C9CEE4}} \color[HTML]{000000} 0.77 & N/A & {\cellcolor[HTML]{D8D7E9}} \color[HTML]{000000} 0.72 & N/A & N/A & N/A & {\cellcolor[HTML]{C2CBE2}} \color[HTML]{000000} 0.79 \\
\hline
\end{tabular}
}
\caption{Per-platform AUC ROC performance of statistical MGT detectors category. N/A refers
to not enough samples per each class (at least 10) making AUC ROC value irrelevant.}
\label{tab:data_perplatform_statistical}
\end{table*}

\begin{table*}
\centering
\resizebox{\textwidth}{!}{
\addtolength{\tabcolsep}{-2pt}
\begin{tabular}{c|l|cccccccccccccccccccccc|c}
\hline
 &  & \multicolumn{23}{c}{\bfseries Test Language [AUC ROC]} \\
\bfseries Detector & \bfseries Platform & \bfseries ar & \bfseries bg & \bfseries ca & \bfseries cs & \bfseries de & \bfseries el & \bfseries en & \bfseries es & \bfseries et & \bfseries ga & \bfseries gd & \bfseries hr & \bfseries hu & \bfseries nl & \bfseries pl & \bfseries pt & \bfseries ro & \bfseries ru & \bfseries sk & \bfseries sl & \bfseries uk & \bfseries zh & \bfseries all \\
\hline
\multirow[c]{5}{*}{\rotatebox{90}{\parbox{2cm}{\bfseries BLOOMZ-3b-mixed-Detector}}} & \bfseries Discord & N/A & N/A & {\cellcolor[HTML]{84B0D3}} \color[HTML]{000000} 0.96 & {\cellcolor[HTML]{9FBAD9}} \color[HTML]{000000} 0.89 & {\cellcolor[HTML]{A9BFDC}} \color[HTML]{000000} 0.87 & N/A & {\cellcolor[HTML]{A7BDDB}} \color[HTML]{000000} 0.87 & {\cellcolor[HTML]{9CB9D9}} \color[HTML]{000000} 0.90 & {\cellcolor[HTML]{A9BFDC}} \color[HTML]{000000} 0.87 & {\cellcolor[HTML]{C1CAE2}} \color[HTML]{000000} 0.80 & {\cellcolor[HTML]{E5E1EF}} \color[HTML]{000000} 0.66 & {\cellcolor[HTML]{B4C4DF}} \color[HTML]{000000} 0.84 & {\cellcolor[HTML]{A9BFDC}} \color[HTML]{000000} 0.86 & {\cellcolor[HTML]{99B8D8}} \color[HTML]{000000} 0.91 & {\cellcolor[HTML]{99B8D8}} \color[HTML]{000000} 0.91 & {\cellcolor[HTML]{A1BBDA}} \color[HTML]{000000} 0.89 & {\cellcolor[HTML]{C6CCE3}} \color[HTML]{000000} 0.78 & N/A & {\cellcolor[HTML]{D9D8EA}} \color[HTML]{000000} 0.71 & N/A & N/A & N/A & {\cellcolor[HTML]{A8BEDC}} \color[HTML]{000000} 0.87 \\
\hline
\end{tabular}
}
\caption{Per-platform AUC ROC performance of pre-trained MGT detectors category. N/A refers
to not enough samples per each class (at least 10) making AUC ROC value irrelevant.}
\label{tab:data_perplatform_pretrained}
\end{table*}

\begin{table*}
\centering
\resizebox{\textwidth}{!}{
\addtolength{\tabcolsep}{-2pt}
\begin{tabular}{c|l|cccccccccccccccccccccc|c}
\hline
&  & \multicolumn{23}{c}{\bfseries Test Language [AUC ROC]} \\
\bfseries Detector & \bfseries Platform & \bfseries ar & \bfseries bg & \bfseries ca & \bfseries cs & \bfseries de & \bfseries el & \bfseries en & \bfseries es & \bfseries et & \bfseries ga & \bfseries gd & \bfseries hr & \bfseries hu & \bfseries nl & \bfseries pl & \bfseries pt & \bfseries ro & \bfseries ru & \bfseries sk & \bfseries sl & \bfseries uk & \bfseries zh & \bfseries all \\
\hline
\multirow[c]{5}{*}{\rotatebox{90}{\parbox{2cm}{\bfseries Aya-101-MultiSocial}}} & \bfseries Discord & N/A & N/A & {\cellcolor[HTML]{76AAD0}} \color[HTML]{000000} 0.99 & {\cellcolor[HTML]{75A9CF}} \color[HTML]{000000} 1.00 & {\cellcolor[HTML]{80AED2}} \color[HTML]{000000} 0.97 & N/A & {\cellcolor[HTML]{79ABD0}} \color[HTML]{000000} 0.99 & {\cellcolor[HTML]{78ABD0}} \color[HTML]{000000} 0.99 & {\cellcolor[HTML]{76AAD0}} \color[HTML]{000000} 1.00 & {\cellcolor[HTML]{83AFD3}} \color[HTML]{000000} 0.96 & {\cellcolor[HTML]{9AB8D8}} \color[HTML]{000000} 0.90 & {\cellcolor[HTML]{78ABD0}} \color[HTML]{000000} 0.99 & {\cellcolor[HTML]{75A9CF}} \color[HTML]{000000} 1.00 & {\cellcolor[HTML]{79ABD0}} \color[HTML]{000000} 0.99 & {\cellcolor[HTML]{76AAD0}} \color[HTML]{000000} 0.99 & {\cellcolor[HTML]{76AAD0}} \color[HTML]{000000} 0.99 & {\cellcolor[HTML]{76AAD0}} \color[HTML]{000000} 0.99 & N/A & {\cellcolor[HTML]{78ABD0}} \color[HTML]{000000} 0.99 & N/A & N/A & N/A & {\cellcolor[HTML]{78ABD0}} \color[HTML]{000000} 0.99 \\
\hline
\end{tabular}
}
\caption{Per-platform AUC ROC performance of fine-tuned MGT detectors category. N/A refers
to not enough samples per each class (at least 10) making AUC ROC value irrelevant.}
\label{tab:data_perplatform_finetuned}
\end{table*}

\begin{table*}
\centering
\resizebox{0.6\textwidth}{!}{
\addtolength{\tabcolsep}{-2pt}
\begin{tabular}{c|l|ccccc|c}
\hline
&  & \multicolumn{6}{c}{\bfseries Platform [AUC ROC]} \\
\bfseries Detector & \bfseries Generator & \bfseries Discord & \bfseries Gab & \bfseries Telegram & \bfseries Twitter & \bfseries WhatsApp & \bfseries {all} \\
\hline
\multirow[c]{7}{*}{\rotatebox{90}{\parbox{2cm}{\bfseries Binoculars}}} & \bfseries Mistral-7B-Instruct-v0.2 & {\cellcolor[HTML]{C9CEE4}} \color[HTML]{000000} 0.77 & {\cellcolor[HTML]{EDE8F3}} \color[HTML]{000000} 0.62 & {\cellcolor[HTML]{DFDDEC}} \color[HTML]{000000} 0.68 & {\cellcolor[HTML]{DFDDEC}} \color[HTML]{000000} 0.68 & {\cellcolor[HTML]{E3E0EE}} \color[HTML]{000000} 0.67 & {\cellcolor[HTML]{E1DFED}} \color[HTML]{000000} 0.68 \\
\bfseries  & \bfseries Aya-101 & {\cellcolor[HTML]{CED0E6}} \color[HTML]{000000} 0.76 & {\cellcolor[HTML]{E7E3F0}} \color[HTML]{000000} 0.65 & {\cellcolor[HTML]{E0DDED}} \color[HTML]{000000} 0.68 & {\cellcolor[HTML]{DCDAEB}} \color[HTML]{000000} 0.70 & {\cellcolor[HTML]{D8D7E9}} \color[HTML]{000000} 0.72 & {\cellcolor[HTML]{DFDDEC}} \color[HTML]{000000} 0.69 \\
\bfseries  & \bfseries Gemini & {\cellcolor[HTML]{A4BCDA}} \color[HTML]{000000} 0.88 & {\cellcolor[HTML]{C4CBE3}} \color[HTML]{000000} 0.79 & {\cellcolor[HTML]{B3C3DE}} \color[HTML]{000000} 0.84 & {\cellcolor[HTML]{B5C4DF}} \color[HTML]{000000} 0.83 & {\cellcolor[HTML]{B4C4DF}} \color[HTML]{000000} 0.84 & {\cellcolor[HTML]{B7C5DF}} \color[HTML]{000000} 0.83 \\
\bfseries  & \bfseries GPT-3.5-Turbo-0125 & {\cellcolor[HTML]{D0D1E6}} \color[HTML]{000000} 0.75 & {\cellcolor[HTML]{E9E5F1}} \color[HTML]{000000} 0.64 & {\cellcolor[HTML]{D9D8EA}} \color[HTML]{000000} 0.71 & {\cellcolor[HTML]{E0DEED}} \color[HTML]{000000} 0.68 & {\cellcolor[HTML]{DEDCEC}} \color[HTML]{000000} 0.69 & {\cellcolor[HTML]{E0DDED}} \color[HTML]{000000} 0.68 \\
\hline
\end{tabular}
}
\caption{Per-platform per-LLM AUC ROC performance of statistical MGT detectors category. N/A refers
to not enough samples per each class (at least 10) making AUC ROC value irrelevant.}
\label{tab:data_perplatform_perllm_statistical}
\end{table*}

\begin{table*}
\centering
\resizebox{0.5\textwidth}{!}{
\addtolength{\tabcolsep}{-2pt}
\begin{tabular}{c|l|ccccc|c}
\hline
&  & \multicolumn{6}{c}{\bfseries Platform [AUC ROC]} \\
\bfseries Detector & \bfseries Generator & \bfseries Discord & \bfseries Gab & \bfseries Telegram & \bfseries Twitter & \bfseries WhatsApp & \bfseries {all} \\
\hline
\multirow[c]{7}{*}{\rotatebox{90}{\parbox{2cm}{\bfseries BLOOMZ-3b-mixed-Detector}}} & \bfseries Mistral-7B-Instruct-v0.2 & {\cellcolor[HTML]{A8BEDC}} \color[HTML]{000000} 0.87 & {\cellcolor[HTML]{EDE7F2}} \color[HTML]{000000} 0.62 & {\cellcolor[HTML]{CED0E6}} \color[HTML]{000000} 0.76 & {\cellcolor[HTML]{DBDAEB}} \color[HTML]{000000} 0.70 & {\cellcolor[HTML]{C5CCE3}} \color[HTML]{000000} 0.78 & {\cellcolor[HTML]{D3D4E7}} \color[HTML]{000000} 0.74 \\
\bfseries  & \bfseries Aya-101 & {\cellcolor[HTML]{93B5D6}} \color[HTML]{000000} 0.92 & {\cellcolor[HTML]{CED0E6}} \color[HTML]{000000} 0.75 & {\cellcolor[HTML]{ADC1DD}} \color[HTML]{000000} 0.85 & {\cellcolor[HTML]{BBC7E0}} \color[HTML]{000000} 0.81 & {\cellcolor[HTML]{B0C2DE}} \color[HTML]{000000} 0.85 & {\cellcolor[HTML]{B4C4DF}} \color[HTML]{000000} 0.83 \\
\bfseries  & \bfseries Gemini & {\cellcolor[HTML]{CDD0E5}} \color[HTML]{000000} 0.76 & {\cellcolor[HTML]{FFF7FB}} \color[HTML]{000000} 0.47 & {\cellcolor[HTML]{ECE7F2}} \color[HTML]{000000} 0.63 & {\cellcolor[HTML]{FDF5FA}} \color[HTML]{000000} 0.52 & {\cellcolor[HTML]{F3EDF5}} \color[HTML]{000000} 0.58 & {\cellcolor[HTML]{F1EBF4}} \color[HTML]{000000} 0.59 \\
\bfseries  & \bfseries GPT-3.5-Turbo-0125 & {\cellcolor[HTML]{99B8D8}} \color[HTML]{000000} 0.91 & {\cellcolor[HTML]{DCDAEB}} \color[HTML]{000000} 0.70 & {\cellcolor[HTML]{BFC9E1}} \color[HTML]{000000} 0.80 & {\cellcolor[HTML]{CED0E6}} \color[HTML]{000000} 0.75 & {\cellcolor[HTML]{ACC0DD}} \color[HTML]{000000} 0.86 & {\cellcolor[HTML]{C1CAE2}} \color[HTML]{000000} 0.80 \\
\hline
\end{tabular}
}
\caption{Per-platform per-LLM AUC ROC performance of pre-trained MGT detectors category. N/A refers
to not enough samples per each class (at least 10) making AUC ROC value irrelevant.}
\label{tab:data_perplatform_perllm_pretrained}
\end{table*}

\begin{table*}
\centering
\resizebox{0.5\textwidth}{!}{
\addtolength{\tabcolsep}{-2pt}
\begin{tabular}{c|l|ccccc|c}
\hline
&  & \multicolumn{6}{c}{\bfseries Platform [AUC ROC]} \\
\bfseries Detector & \bfseries Generator & \bfseries Discord & \bfseries Gab & \bfseries Telegram & \bfseries Twitter & \bfseries WhatsApp & \bfseries {all} \\
\hline
\multirow[c]{7}{*}{\rotatebox{90}{\parbox{2cm}{\bfseries Aya-101-MultiSocial}}} & \bfseries Mistral-7B-Instruct-v0.2 & {\cellcolor[HTML]{75A9CF}} \color[HTML]{000000} 1.00 & {\cellcolor[HTML]{7EADD1}} \color[HTML]{000000} 0.98 & {\cellcolor[HTML]{76AAD0}} \color[HTML]{000000} 0.99 & {\cellcolor[HTML]{76AAD0}} \color[HTML]{000000} 0.99 & {\cellcolor[HTML]{75A9CF}} \color[HTML]{000000} 1.00 & {\cellcolor[HTML]{78ABD0}} \color[HTML]{000000} 0.99 \\
\bfseries  & \bfseries Aya-101 & {\cellcolor[HTML]{79ABD0}} \color[HTML]{000000} 0.99 & {\cellcolor[HTML]{99B8D8}} \color[HTML]{000000} 0.91 & {\cellcolor[HTML]{81AED2}} \color[HTML]{000000} 0.97 & {\cellcolor[HTML]{83AFD3}} \color[HTML]{000000} 0.96 & {\cellcolor[HTML]{88B1D4}} \color[HTML]{000000} 0.95 & {\cellcolor[HTML]{86B0D3}} \color[HTML]{000000} 0.96 \\
\bfseries  & \bfseries Gemini & {\cellcolor[HTML]{7DACD1}} \color[HTML]{000000} 0.98 & {\cellcolor[HTML]{A9BFDC}} \color[HTML]{000000} 0.86 & {\cellcolor[HTML]{89B1D4}} \color[HTML]{000000} 0.95 & {\cellcolor[HTML]{8FB4D6}} \color[HTML]{000000} 0.93 & {\cellcolor[HTML]{8FB4D6}} \color[HTML]{000000} 0.93 & {\cellcolor[HTML]{8FB4D6}} \color[HTML]{000000} 0.93 \\
\bfseries  & \bfseries GPT-3.5-Turbo-0125 & {\cellcolor[HTML]{76AAD0}} \color[HTML]{000000} 1.00 & {\cellcolor[HTML]{7EADD1}} \color[HTML]{000000} 0.98 & {\cellcolor[HTML]{76AAD0}} \color[HTML]{000000} 0.99 & {\cellcolor[HTML]{76AAD0}} \color[HTML]{000000} 1.00 & {\cellcolor[HTML]{76AAD0}} \color[HTML]{000000} 1.00 & {\cellcolor[HTML]{78ABD0}} \color[HTML]{000000} 0.99 \\
\hline
\end{tabular}
}
\caption{Per-platform per-LLM AUC ROC performance of fine-tuned MGT detectors category. N/A refers
to not enough samples per each class (at least 10) making AUC ROC value irrelevant.}
\label{tab:data_perplatform_perllm_finetuned}
\end{table*}

\begin{table*}
\centering
\resizebox{\textwidth}{!}{
\addtolength{\tabcolsep}{-2pt}
\begin{tabular}{c|c|cccccccccccccccccccccc|c}
\hline
 & \bfseries Train & \multicolumn{23}{c}{\bfseries Test Language [AUC ROC]} \\
\bfseries Detector & \bfseries Language & \bfseries ar & \bfseries bg & \bfseries ca & \bfseries cs & \bfseries de & \bfseries el & \bfseries en & \bfseries es & \bfseries et & \bfseries ga & \bfseries gd & \bfseries hr & \bfseries hu & \bfseries nl & \bfseries pl & \bfseries pt & \bfseries ro & \bfseries ru & \bfseries sk & \bfseries sl & \bfseries uk & \bfseries zh & \bfseries all \\
\hline
\multirow[c]{4}{*}{\rotatebox{90}{\parbox{1.8cm}{\bfseries Aya-101}}} & \bfseries en & {\cellcolor[HTML]{9FBAD9}} \color[HTML]{000000} 0.89 & {\cellcolor[HTML]{7EADD1}} \color[HTML]{000000} 0.97 & {\cellcolor[HTML]{B1C2DE}} \color[HTML]{000000} 0.84 & {\cellcolor[HTML]{80AED2}} \color[HTML]{000000} 0.97 & {\cellcolor[HTML]{93B5D6}} \color[HTML]{000000} 0.92 & {\cellcolor[HTML]{88B1D4}} \color[HTML]{000000} 0.95 & {\cellcolor[HTML]{80AED2}} \color[HTML]{000000} 0.97 & {\cellcolor[HTML]{91B5D6}} \color[HTML]{000000} 0.93 & {\cellcolor[HTML]{84B0D3}} \color[HTML]{000000} 0.96 & N/A & N/A & {\cellcolor[HTML]{83AFD3}} \color[HTML]{000000} 0.96 & {\cellcolor[HTML]{76AAD0}} \color[HTML]{000000} 0.99 & {\cellcolor[HTML]{A7BDDB}} \color[HTML]{000000} 0.87 & {\cellcolor[HTML]{89B1D4}} \color[HTML]{000000} 0.95 & {\cellcolor[HTML]{89B1D4}} \color[HTML]{000000} 0.95 & {\cellcolor[HTML]{81AED2}} \color[HTML]{000000} 0.97 & {\cellcolor[HTML]{8EB3D5}} \color[HTML]{000000} 0.93 & N/A & N/A & {\cellcolor[HTML]{9EBAD9}} \color[HTML]{000000} 0.90 & {\cellcolor[HTML]{97B7D7}} \color[HTML]{000000} 0.91 & {\cellcolor[HTML]{8EB3D5}} \color[HTML]{000000} 0.93 \\
\hline
\end{tabular}
}
\vspace{-1mm}
\caption{Cross-lingual AUC ROC performance of the selected MGT detectors fine-tuned monolingually (\textit{en}, \textit{es} and \textit{ru}) and multilingually (\textit{en-es-ru}), evaluated based on Telegram data (for training as well as for testing).  N/A refers to not enough samples (at least 2000) in MultiSocial Telegram data.}
\label{tab:data_crosslingual}
\vspace{-3mm}
\end{table*}

\begin{table*}
\centering
\resizebox{0.54\textwidth}{!}{
\addtolength{\tabcolsep}{-2pt}
\begin{tabular}{ll|ccccc|c}
\hline
 & \bfseries Train & \multicolumn{6}{c}{\bfseries Test Platform [AUC ROC]} \\
 \bfseries Detector &  \bfseries Platform & \bfseries Discord & \bfseries Gab & \bfseries Telegram & \bfseries Twitter & \bfseries WhatsApp  & \bfseries all \\
\hline
\multirow[c]{6}{*}{\rotatebox{90}{\parbox{2cm}{\bfseries Aya-101}}} & \bfseries Discord & {\cellcolor[HTML]{79ABD0}} \color[HTML]{000000} 0.99 & {\cellcolor[HTML]{B8C6E0}} \color[HTML]{000000} 0.82 & {\cellcolor[HTML]{A1BBDA}} \color[HTML]{000000} 0.89 & {\cellcolor[HTML]{BBC7E0}} \color[HTML]{000000} 0.81 & {\cellcolor[HTML]{96B6D7}} \color[HTML]{000000} 0.92 & {\cellcolor[HTML]{A2BCDA}} \color[HTML]{000000} 0.89 \\
\bfseries  & \bfseries Gab & {\cellcolor[HTML]{80AED2}} \color[HTML]{000000} 0.97 & {\cellcolor[HTML]{83AFD3}} \color[HTML]{000000} 0.96 & {\cellcolor[HTML]{88B1D4}} \color[HTML]{000000} 0.95 & {\cellcolor[HTML]{83AFD3}} \color[HTML]{000000} 0.96 & {\cellcolor[HTML]{8CB3D5}} \color[HTML]{000000} 0.94 & {\cellcolor[HTML]{86B0D3}} \color[HTML]{000000} 0.96 \\
\bfseries  & \bfseries Telegram & {\cellcolor[HTML]{7BACD1}} \color[HTML]{000000} 0.98 & {\cellcolor[HTML]{93B5D6}} \color[HTML]{000000} 0.92 & {\cellcolor[HTML]{81AED2}} \color[HTML]{000000} 0.97 & {\cellcolor[HTML]{89B1D4}} \color[HTML]{000000} 0.95 & {\cellcolor[HTML]{86B0D3}} \color[HTML]{000000} 0.96 & {\cellcolor[HTML]{86B0D3}} \color[HTML]{000000} 0.95 \\
\bfseries  & \bfseries Twitter & {\cellcolor[HTML]{7DACD1}} \color[HTML]{000000} 0.98 & {\cellcolor[HTML]{94B6D7}} \color[HTML]{000000} 0.92 & {\cellcolor[HTML]{8CB3D5}} \color[HTML]{000000} 0.94 & {\cellcolor[HTML]{7DACD1}} \color[HTML]{000000} 0.98 & {\cellcolor[HTML]{8BB2D4}} \color[HTML]{000000} 0.94 & {\cellcolor[HTML]{89B1D4}} \color[HTML]{000000} 0.95 \\
\hline
\end{tabular}
}
\vspace{-2mm}
\caption{Cross-platform evaluation of the selected fine-tuned MGT detectors.}
\label{tab:data_crossplatform}
%\vspace{-3mm}
\end{table*}

\end{document}